\documentclass{article} 
\usepackage{iclr2027_conference,times}

\usepackage{graphicx}
\usepackage{subcaption}
\usepackage{pifont}
\usepackage{booktabs}
\newcommand{\xmark}{\ding{55}}

\usepackage{algorithm}
\usepackage{algpseudocode}
\usepackage[normalem]{ulem}
\usepackage{arydshln}
\usepackage{amsmath}
\usepackage{amssymb}

\usepackage{amsthm}
\newtheorem{proposition}{Proposition}

\newtheorem{lemma}{Lemma}
\theoremstyle{remark}
\newtheorem{remark}{Remark}

\newcommand{\best}[1]{{\boldmath$#1$}}
\newcommand{\second}[1]{{\boldmath\dashuline{$#1$}}}
\usepackage{amsmath,amsfonts,bm}

\def\eqref#1{equation~\ref{#1}}

\def\1{\bm{1}}

\DeclareMathAlphabet{\mathsfit}{\encodingdefault}{\sfdefault}{m}{sl}
\SetMathAlphabet{\mathsfit}{bold}{\encodingdefault}{\sfdefault}{bx}{n}

\usepackage{hyperref}
\usepackage{url}

\title{ReDiF: Resource-Efficient Few-Step Diffusion Distillation via Reinforcement Learning}

\author{Amirhossein Tighkhorshid, Zahra Dehghanian \& Hamid R. Rabiee \\
Department of Computer Engineering \\
Sharif University of Technology \\
Tehran, Iran \\
\texttt{\{amir.tighkhorshid78, zahra.dehghanian97, rabiee\}@sharif.edu}
}

\iclrfinalcopy 
\begin{document}

\maketitle
\lhead{Under review as a conference paper at ICLR 2027}

\begin{abstract}
Step distillation accelerates diffusion sampling by training a few-step student to imitate a many-step teacher, but distillation itself remains expensive. Typically, this requires thousands of GPU-hours and a large pre-generated trajectory dataset. We introduce ReDiF, which casts step distillation as terminal-reward policy optimization rather than step-wise regression. The student is optimized against a reward computed on the terminal sample, measuring alignment with the teacher's output, instead of matching the teacher's intermediate trajectory under a reconstruction or consistency loss. Because the reward need not be differentiable or trajectory-aligned, ReDiF admits non-differentiable objectives, multi-objective combinations, and preferences the teacher does not express, while exploration lets the student find sampling paths matched to its own step schedule. ReDiF converges in 400 policy updates with 3200 rollouts on a single A100 GPU with 1,000 noise-class pairs and no paired dataset: about 4 GPU-hours, against roughly 336 A100-hours reported for DMD2 on the same EDM teacher. At 8 steps on ImageNet-64, it achieves an FID 3.64 points better than the strongest retrained distillation baseline in the low-training regime under the same single-GPU budget. The formulation is also orthogonal to existing distillation objectives: added to the DMD2 loss, it further improves DMD2's FID by 4.4.
\end{abstract}

\section{Introduction}
\label{sec:intro}

Sampling from diffusion models can be slow and computationally expensive because high-quality generation typically requires iterative denoising over many steps, while naively reducing the step count can degrade sample quality~\citep{ho2020denoising,song2021scorebased,karras2022elucidating}. Training-free approaches try to reduce the step count with better numerical solvers; they are cheap to apply but degrade sharply at very low step counts, where numerical approximation error leads to reduced fidelity and loss of fine-grained detail~\citep{song2020ddim,song2021scorebased,lu2022dpmsolver,zheng2023fast,shih2023parallel}. Distillation instead adapts the model itself: a teacher running many denoising steps transfers its behavior to a cheaper student, trained to operate effectively at very low step counts where conventional solvers can suffer substantial quality degradation~\citep{salimans2022progressive,song2023consistency,zheng2023truncated}. Distillation can reduce the number of denoising steps, the model size, or both; we focus on step distillation, although the formulation is not tied to a particular student architecture.
 
Existing distillation methods differ primarily in how they supervise the student. Trajectory-based approaches learn from the teacher's sampling trajectory: Progressive Distillation halves the number of steps at each training stage, so that each stage only has to absorb a small reduction in step count~\citep{salimans2022progressive}, and Consistency Distillation trains the student to map any point on the teacher's trajectory directly to its endpoint~\citep{song2023consistency,luo2023lcm,kim2024ctm}. Distribution-matching methods such as DMD, DMD2, SiD, and SiDA instead optimize objectives that approximate the teacher's output distribution~\citep{yin2024DMD,yin2024IDMD,zhou2024sid,zhou2024sida}. 

These approaches have substantially improved few-step generation, to the point where one-step students match or exceed their teachers~\citep{yin2024IDMD,zhou2024sid,zhou2024sida}, but at a high training cost: several thousand A100-hours, at effective batch sizes in the thousands~\citep{Zhou2024SFD,liu2024instaflow,yin2024IDMD}. For many of these methods, the effective batch size is a particularly restrictive constraint. While computation can be traded against wall-clock time, a large effective batch imposes a hardware requirement that cannot be removed simply by training longer on a single GPU, which can make the distillation of large models difficult without access to multi-GPU infrastructure. This motivates a different question: can diffusion distillation be carried out with far less compute, training time, and batch size, within the limited resource budget of a single GPU?
 
 
We introduce ReDiF, which formulates step distillation as terminal-reward policy optimization. Starting from the teacher's parameters, the student generates a few-step sample and is optimized against a scalar reward measuring its alignment with the teacher's output from the same conditioning. Because the student is initialized with the teacher's weights, optimization focuses on correcting the discrepancy introduced by the coarser sampling schedule rather than learning the generative process from scratch. Unlike conventional distillation objectives, ReDiF requires neither step-wise supervision, auxiliary score networks or discriminators, nor paired teacher--student trajectories, and the reward may be non-differentiable or multi-objective.

ReDiF distills a 256-step EDM teacher to 8 steps on a single A100 using 1,000 noise-class pairs in approximately 4 GPU-hours, compared with the ${\sim}336$ A100-hours reported for DMD2 on the same teacher. Under the same single-GPU budget, ReDiF achieves an FID 3.64 points lower than the strongest retrained distillation baseline. Moreover, adding ReDiF to the DMD2 objective further improves DMD2's FID by 4.4 points, demonstrating that the proposed objective is complementary to existing distillation approaches.
 
\begin{figure*}[t]
    \centering
    \setlength{\tabcolsep}{2pt}
    \renewcommand{\arraystretch}{1.2}
    \begin{tabular}{c|ccccc|c}
        & \textbf{Epoch 0} & \textbf{Epoch 4} & \textbf{Epoch 8} & \textbf{Epoch 12} & \textbf{Epoch 16} & \textbf{Teacher Ref.} \\
        \midrule
        \rotatebox[origin=c]{90}{\textbf{ }} &
        \includegraphics[width=0.145\textwidth]{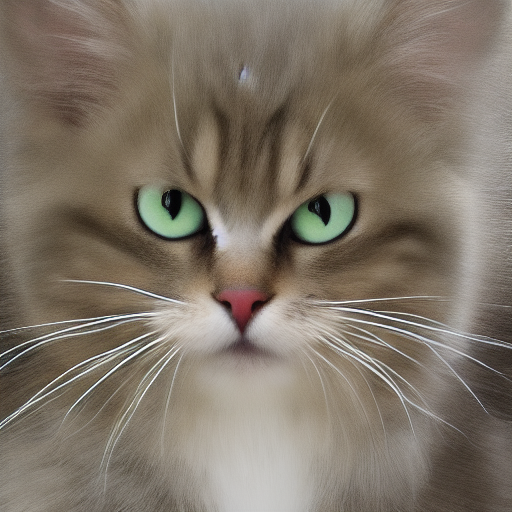} &
        \includegraphics[width=0.145\textwidth]{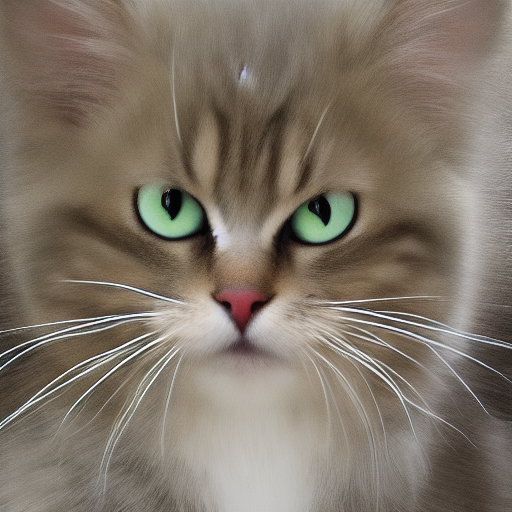} &
        \includegraphics[width=0.145\textwidth]{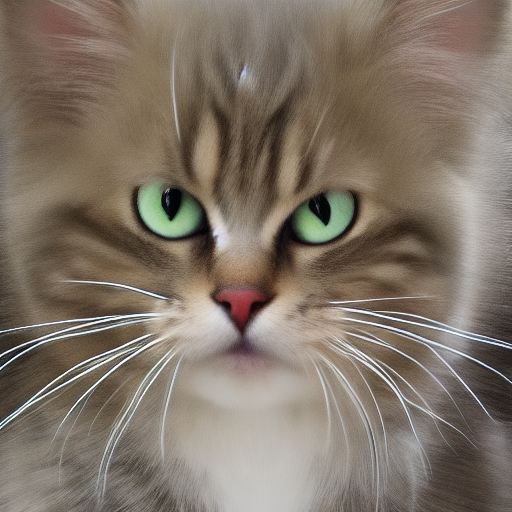} &
        \includegraphics[width=0.145\textwidth]{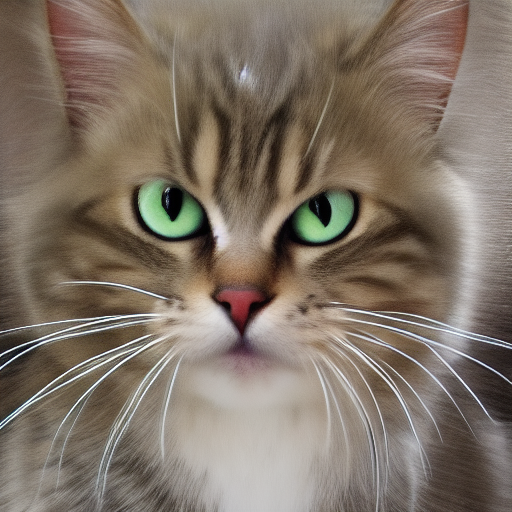} &
        \includegraphics[width=0.145\textwidth]{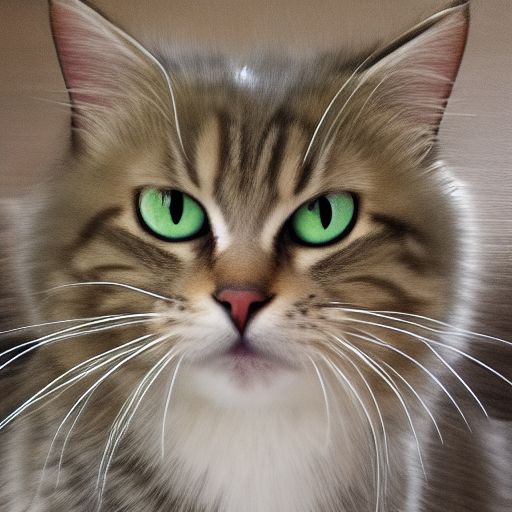} &
        \includegraphics[width=0.145\textwidth]{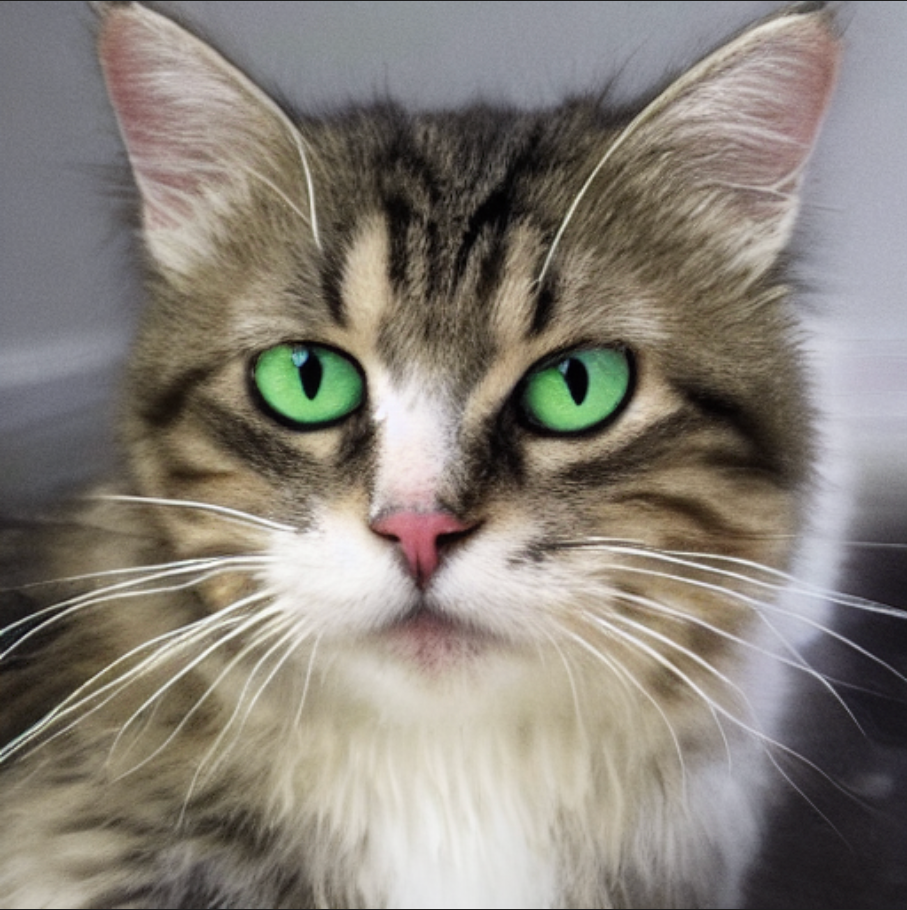} \\
    \end{tabular}
    \caption{
    ReDiF training on the Stable Diffusion v1.5 branch ($50\rightarrow5$ steps). Within a few epochs a
    student sampling at a tenth of the teacher's step count reaches a quality close to the teacher's
    reference (rightmost column), although the only supervision is a scalar computed on the final
    image. Columns are successive epochs of one run at fixed conditioning.
    Figure~\ref{fig:training_progress_oop} continues this run past the budget.
    }
    \label{fig:training_progress}
\end{figure*}

\section{Related Work}
\label{sec:related_work}

\textbf{Trajectory-based distillation.}
KDIGM~\citep{luhman2021knowledge} first explored compressing a multi-step diffusion model into a one-step student, but at a substantial cost in sample quality. Progressive Distillation~\citep{salimans2022progressive} halves the teacher's sampling steps over multiple training stages, while Guided Distillation~\citep{meng2023distillation} extends this approach to classifier-free guided samplers. Consistency Distillation~\citep{song2023consistency} replaces the recursive procedure with a self-consistency objective along the probability-flow ODE, training the student to map any point on the trajectory to its endpoint. RDD~\citep{Feng2024RDD} and SFD~\citep{Zhou2024SFD} refine this objective, LCM and LCM-LoRA~\citep{luo2023lcm,luo2023lcmlora} extend it to latent diffusion, and CTM~\citep{kim2024ctm} generalizes it to arbitrary trajectory jumps. InstaFlow~\citep{liu2024instaflow} instead straightens the sampling trajectory using rectified flow. These methods derive supervision from teacher trajectories or intermediate states, which can constrain the student's sampling behavior under aggressive step reduction.

\textbf{Distribution matching.}
An alternative line of work matches the student's output distribution to the teacher's rather than explicitly imitating its trajectory. DMD~\citep{yin2024DMD} uses an approximate reverse KL estimated from score differences, while SiD~\citep{zhou2024sid} relies on score identities for distribution matching. DMD2~\citep{yin2024IDMD}, SiDA~\citep{zhou2024sida}, and ADD~\citep{sauer2023adversarialdiffusiondistillation} incorporate adversarial components. These methods achieve highly competitive few-step generation quality, but introduce additional machinery, such as auxiliary networks and, for the adversarial variants, real training images. TFD~\citep{zhang2026tfd} removes the auxiliary network through a hidden-state-based objective, but still relies on reference data in the settings it considers.

\textbf{Data-free distillation.}
Data-free methods remove the dependence on real training images by using synthetic teacher-generated supervision. BOOT~\citep{gu2023boot} trains on consecutive teacher samples, DKDM~\citep{xiang2025dkdm} manages the required synthetic data, and PMDD~\citep{tran2025progressive} extends the setting to progressive multi-step distillation. Guided SiD~\citep{zhouguided}, DiffDFKD~\citep{qi2025data}, and Random Conditioning Distillation~\citep{kim2025random} address conditioning-space coverage without real images. Although these methods eliminate real data, they generally retain expensive supervised or distribution-matching objectives and rely on large synthetic datasets or a large number of conditioning signals.

\textbf{Reward-based optimization.}
Reward-based optimization has primarily been used to align diffusion models with external objectives. DDPO~\citep{Black2023DDPO} formulates diffusion sampling as an MDP and optimizes terminal rewards using policy gradients, while DPOK~\citep{Fan2023DPOK}, Diffusion-DPO~\citep{Wallace2023DiffusionDPO}, LOOP~\citep{Gupta2025LOOP}, and RLCM~\citep{oertell2024rlconsistencymodelsfaster} explore related alignment and optimization settings. DRaFT~\citep{clark2024draft} backpropagates rewards through the sampling chain, and LaSRO~\citep{Jia2024LaSRO} learns a latent surrogate reward. Closest to our setting, RG-LCD~\citep{li2024reward} directly optimizes a reward to compensate for the quality degradation of LCM, while retaining the consistency loss as its primary objective. Unlike ReDiF, it does not use reinforcement learning and backpropagates the reward directly. Its main contribution is preventing reward hacking, while using rewards for alignment and quality improvement rather than as the sole objective for distillation.

\textbf{Gap and our approach.}
Existing distillation methods primarily rely on trajectory-level supervision or distribution-matching objectives, while reward-based approaches have mainly been used for alignment or as auxiliary terms in distillation. ReDiF addresses this gap by making terminal-reward RL-based optimization the sole training objective for step distillation. The student is initialized with the teacher's weights and optimized using a reward computed on its final sample, without step-wise reconstruction, auxiliary networks, real training images, or paired teacher--student data. Appendix~\ref{app:comparison} compares ReDiF with existing methods along these axes.

\section{Method}
\label{sec:method}

\begin{figure*}[htbp]
    \centering
    \includegraphics[width=\linewidth]{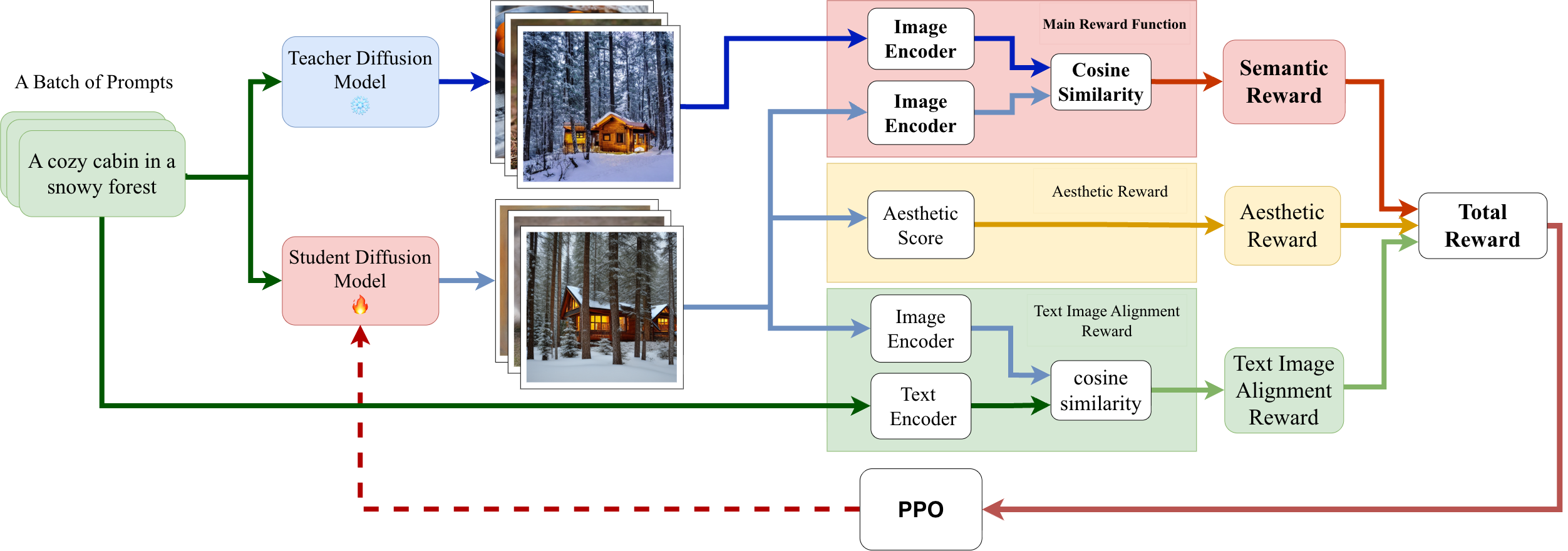}
    \caption{Overview of ReDiF. The same noise and conditioning are provided to both the frozen teacher and the trainable student. Their final images are embedded using frozen, pretrained encoders, and the similarity between the embeddings serves as the terminal reward, which PPO maximizes by updating only the student. Everything to the right of the two generators is frozen and operates in a forward-only manner, while the teacher reference is generated by the teacher's own sampler within the training loop. Aesthetic and Text-Image Alignment are two auxiliary preference rewards.}
    \label{fig:teacher-student-ppo}
\end{figure*}

ReDiF formulates diffusion step distillation as RL-based terminal-reward policy optimization. The student rolls out a complete denoising trajectory with its current weights, its final image is compared with the teacher's final image from the same noise and conditioning, and the resulting scalar is the only supervision it receives; no intermediate state is ever given a target. Figure~\ref{fig:teacher-student-ppo} shows the loop.

\subsection{Why Per-Step Supervision Is Costly}
\label{subsec:cost}

Let $\Phi^{*}$ be the teacher's $N$-step sampler and $\Phi_\theta$ the student's $M$-step sampler, $M\ll N$. Trajectory-matching methods give each supervised transition a target $y_k$ and minimize $\mathcal{L}_{\mathrm{step}}(\theta)=\sum_{k}\mathbb{E}\,\ell(\Phi_\theta(x_{t_k},t_k),y_k)$, with $y_k$ a teacher prediction for reconstruction and progressive losses and an EMA copy of the student for consistency losses; distribution-matching methods match the two output distributions through an
auxiliary score network. Three sources of cost motivate our design; they are arguments, not lower bounds, and Section~\ref{sec:experiments} tests the design empirically.

\textbf{Per-step errors compound.} A deviation at one transition is propagated by all later ones. If
the teacher's maps between consecutive student grid points are $L_k$-Lipschitz, the terminal error is
at most $\Gamma\max_k\varepsilon_k$ with $\Gamma=\sum_k\prod_{j<k}L_j$, which is at least $M$ whenever
$L_j\ge1$ and exponential in $M$ when $L_j\ge L>1$ (Lemma~\ref{lem:amplification},
Appendix~\ref{app:amplification}). A terminal tolerance $\delta$ is therefore guaranteed by a per-step
tolerance of $\delta/\Gamma$: stricter than the accuracy that is eventually measured, which is taken
on the final image alone.

\textbf{Moving targets.} When $y_k$ comes from a network that itself changes during training, as the
EMA target of consistency distillation or the fake-score network of distribution matching, its error
has a systematic component that more samples of the same target do not average away
(Appendix~\ref{app:moving_targets}). Such networks must also be re-fitted to the student throughout
training, at a cost of the same order as the student's own update
(Appendix~\ref{app:cost_model}), and published distribution-matching methods train at effective batch
sizes from $128$ to $8{,}192$ (Table~\ref{tab:published}). We treat this as an empirical constraint of
single-device training rather than as a property we prove.

\textbf{The problem is a fine-tuning problem.} The student starts at the teacher's weights and needs
to compensate only for a coarser schedule; it does not learn the generative process from scratch.
Policy optimization is known to adapt pretrained diffusion models to a new reward in comparatively few
updates~\citep{Black2023DDPO,oertell2024rlconsistencymodelsfaster}, and step distillation satisfies
that premise by construction. This is a working hypothesis, not a theorem: we do not bound how far the
optimal student lies from the teacher, and check it on the curves of Appendix~\ref{app:reward_curves}.

Together these suggest measuring agreement only where it is evaluated, at the terminal sample, estimating no intermediate target, and training nothing besides the student: scoring the finished image with frozen encoders satisfies all three.

\subsection{Step Distillation as a Policy Optimization Problem}
\label{subsec:mdp}

Each student transition conditions only on the current state $s_t=(x_t,t,c)$, so the sampler is
Markov by construction and defines a finite-horizon MDP with horizon $M$, pure noise as the initial
state and the generated image as the terminal state. PPO needs a likelihood ratio, so the action must
have a density. Following DDPO~\citep{Black2023DDPO}, the action is the sampled next latent,
\[
a_t=x_{t-1}\sim\pi_\theta(x_{t-1}\mid s_t)=\mathcal{N}\!\big(\mu_\theta(x_t,t,c),\,\sigma_t^{2}I\big),
\]
where $\mu_\theta$ is the teacher's sampler update on the student's $M$-step schedule, applied to the
student's denoiser output, and $\sigma_t^{2}$ is the variance of the stochastic form of that update,
fixed by the scheduler. Only $\mu_\theta$ is learned, and only through the LoRA
adapter~\citep{hu2021lora}.

Supervision exists only once a trajectory has produced an image, so the reward is sparse and
terminal: $r_t=0$ for $t>0$, $r_0=R(x_0)$ and $J(\theta)=\mathbb{E}_\tau[R(x_0)]$. The score-function
identity~\citep{williams1992simple,sutton1999policy} spreads the scalar over all transitions,
\[
\nabla_\theta J
=\mathbb{E}\Big[(R-b)\textstyle\sum_{k=1}^{M}\nabla_\theta\log\pi_\theta(x_{t_{k-1}}\mid x_{t_k},t_k,c)\Big],
\]
for any baseline $b$ independent of the actions (Proposition~\ref{prop:credit},
Appendix~\ref{app:credit}). No intermediate target appears, and the reward enters only through its
value, so it may be non-differentiable. The price is variance: one scalar multiplies a sum of $M$
score terms, and the variance grows with the horizon. This is one of the two pressures that fix the
student step count (Appendix~\ref{app:step_count}).

The ratio requires $\sigma_t>0$, so rollouts inject noise at every transition, whereas evaluation uses
the deterministic sampler ($\eta=0$). Training changes only $\mu_\theta$, so the deployed sampler
applies the function that was optimized, but along noise-free rather than perturbed states. We do not
claim this mismatch is harmless; every reported number is measured in the deployed configuration, so
it includes whatever the mismatch costs.

\subsection{The Semantic Terminal Reward}
\label{subsec:reward}

We compute the reward in representation space,
\[
R(I_T,I_S)=\sum_{i=1}^{K}\lambda_i\,\mathcal{S}_i\!\left(f_i(I_T),f_i(I_S)\right),
\qquad\sum_i\lambda_i=1,
\]
where $I_T$ and $I_S$ are the teacher and student images from the same $(x_T,c)$, $f_i$ are frozen
encoders (CLIP~\citep{radford2021clip}, DINOv3~\citep{simeoni2025dinov3}, the Perception Encoder
(PE)~\citep{bolya2025perception} and LPIPS~\citep{zhang2018lpips}), and $\mathcal{S}_i$ is cosine
similarity between normalized embeddings, or the negative distance for LPIPS. The teacher side
$f_i(I_T)$ does not depend on $\theta$: $I_T$ comes from the teacher's own sampler inside the
training loop, on the conditioning signal the rollout already uses. Drawing a reference anew per
rollout or reusing it across rollouts that share a conditioning signal leaves the objective
unchanged; neither assembles a dataset of teacher trajectories.

Because teacher and student start from the same noise and conditioning, the reward asks the student
to land where the teacher lands on that input, not only to match its output distribution. The
teacher's sampler is stochastic, so the reference is a sample from its conditional distribution
rather than a fixed target: the student is scored per sample without score estimation, on a scale
comparable across conditionings, so a high reward means agreement rather than an easy conditioning. Section~\ref{subsec:ppo} relies on
this when it standardizes rewards over the batch.

A pixel-space reward penalizes every deviation from the teacher's image, including texture and
sampling noise that do not affect perceived agreement. A representation-space reward is, to first
order, invariant to every perturbation in the null space of the encoder's Jacobian
(Proposition~\ref{prop:semantic}, Appendix~\ref{app:semantic}), so it constrains fewer directions.
This cuts both ways: the student may reach any terminal state the encoder cannot distinguish from the
teacher's, by whichever path suits its schedule, but those same directions are where the policy
drifts once the reward stops improving elsewhere (Appendix~\ref{app:overopt}). Summing encoders
shrinks the invariant set to the intersection of the null spaces; whether that helps depends on the
encoder and branch (Appendix~\ref{app:reward_ablation}).

\subsection{Optimization}
\label{subsec:ppo}

We optimize the clipped PPO surrogate~\citep{schulman2017ppo},
\[
\mathcal{L}_{\mathrm{PPO}}(\theta)
=\mathbb{E}_t\!\left[\min\!\left(\rho_t(\theta)\hat A_t,\;
\operatorname{clip}(\rho_t(\theta),1-\epsilon,1+\epsilon)\hat A_t\right)\right],
\qquad
\rho_t(\theta)=\frac{\pi_\theta(a_t\mid s_t)}{\pi_{\theta_{\mathrm{old}}}(a_t\mid s_t)},
\]
with $\pi_\theta$ the policy of Section~\ref{subsec:mdp} under the guidance convention set out at
the end of this section. The horizon is short and the reward terminal, so we use no value function
and no learned baseline. All transitions of a trajectory share one advantage, the trajectory's reward
standardized over the rollout batch $\mathcal{B}$ of that iteration, as in DDPO~\citep{Black2023DDPO}:
$\hat A=\operatorname{clip}\big((R-\mu_{\mathcal{B}})/(\sigma_{\mathcal{B}}+\varepsilon),
-A_{\max},A_{\max}\big)$, with $\varepsilon$ keeping it defined when all rewards agree and
$A_{\max}=5$ bounding an outlying rollout. We do not group rollouts by conditioning and use no
group-relative advantage: since $\mathcal{B}$ mixes conditionings, this removes the difficulty of the
batch rather than of one conditioning, and a shared baseline is reasonable because each rollout is
scored against its own teacher reference (Section~\ref{subsec:reward}). Each iteration collects
rollouts with $\theta_{\mathrm{old}}$ frozen, stores the sampled latents and their log-densities, then
runs $E$ inner epochs of minibatch updates. Following DDPO the clip ranges are small
(Table~\ref{tab:implementation}), because one update of shared weights moves the log-density of every
transition at once.

Clipping bounds each update but not the cumulative displacement over many, so we add a regularizer
toward the frozen teacher policy,
$\mathcal{L}=\mathcal{L}_{\mathrm{PPO}}-\beta D_{\mathrm{KL}}(\pi_\theta\|\pi_T)$, which for Gaussian
policies with shared variance reduces to $\|\mu_\theta-\mu_T\|^{2}/(2\sigma_t^{2})$ per transition. As
a loss term rather than a reward term, its strength is set independently of the reward scale and does
not pass through standardization. Since the student is the teacher plus a LoRA adapter, $\mu_T$ comes
from disabling the adapter, at one teacher evaluation per transition when $\beta>0$
(Appendix~\ref{app:cost_model}). We use $\beta=0.02$ on ImageNet-64 and $\beta=0$ on COCO. Because the
reward keeps rising after quality saturates, the training length is fixed from held-out FID rather
than from the reward (Appendix~\ref{app:overopt}). Algorithm~\ref{alg:redif}
(Appendix~\ref{app:algorithm}) summarizes the procedure.

\emph{LoRA as an implicit constraint.} Training the student as a low-rank adapter also restricts how
far the policy can move, a role the clip range and the KL term play on other quantities. It keeps FID
and seed spread stable across clip ranges, whereas a full-parameter student depends strongly on both
(Appendix~\ref{app:lora}).

\textbf{Guidance-consistent policies.}
\label{subsec:cfg}
The PPO ratio assumes that actions are drawn from the distribution whose log-density the update
evaluates. Classifier-free guidance changes that distribution: with scale $w$ the transition mean is
$\mu^{(w)}_\theta=\mu_\theta(x_t,t,\varnothing)+w(\mu_\theta(x_t,t,c)-\mu_\theta(x_t,t,\varnothing))$,
so trajectories are generated by $\pi^{(w)}_\theta=\mathcal{N}(\mu^{(w)}_\theta,\sigma_t^{2}I)$, not by
the conditional branch alone. Sampling from $\pi^{(w)}_{\theta_{\mathrm{old}}}$ while evaluating
log-densities under the unguided $\pi_\theta$ compares two distributions rather than successive
iterates of one policy: the ratio differs from $1$ already at $\theta=\theta_{\mathrm{old}}$, so the
first inner epoch is biased and the clip range no longer bounds the policy change it is meant to
bound. Consistency requires the same $w$ in rollout and update.

The distinction is vacuous for the class-conditional EDM teacher, which we run without guidance. For
Stable Diffusion the guided composition with $w=5$ is the policy in both rollout and update, doubling
the network evaluations per transition; the reward, a function of the terminal image only, is
unaffected.

\subsection{RL versus Direct Backpropagation}
\label{subsec:why_rl}
\label{subsec:preference}

The rewards of Section~\ref{subsec:reward} are computed by differentiable encoders, so they could be
backpropagated through the unrolled sampler as in DRaFT~\citep{clark2024draft}. We treat them as
black boxes for four reasons. \emph{Credit assignment:} policy optimization draws several
trajectories per iteration, forms a baseline from their rewards, and reinforces the transitions of
trajectories that scored above it while suppressing those that scored below. Every transition,
including an early one whose only merit is setting up a good endpoint, is therefore credited by
comparing sampled alternatives, without differentiating the sampler. Direct backpropagation also
credits early transitions, but through the local derivative of a single deterministic trajectory,
carried back through the Jacobians of every later step, and a truncated backward pass gives the
earliest transitions no credit at all. In expectation over the same stochastic chain the two estimate
the same gradient~\citep{mohamed2020monte}; they differ in variance and in what each sample requires.
\emph{Memory:} a pathwise gradient must retain, or recompute, the graph of every sampler step it
differentiates through, whereas the score-function estimator needs one transition at a time.
\emph{Generality:} policy optimization uses only the value of the reward, so non-differentiable
rewards need no change; the preference terms below are exactly that. \emph{Reward exploitation:}
pathwise ascent follows the encoder's input gradient exactly in every sample, including directions
that raise the score without improving the image~\citep{li2024reward}, whereas policy optimization
never evaluates that gradient, follows it only in expectation and smoothed by the policy noise, and
clips each update; it is not immune (Appendix~\ref{app:overopt}).
Section~\ref{subsec:matched} compares the two under an identical reward, initialization and budget;
the direct-backpropagation rows also use a deterministic chain and a truncated backward pass, so the
comparison measures these differences together.

\paragraph{Adding preferences to the distillation reward.} Because distillation is already cast as
reward maximization, a preference is added by adding a term. With $R_{\mathrm{pref}}$ a score on the
student's image alone,
\begin{equation}
  R_{\mathrm{tot}}(I_T, I_S) \;=\; R(I_T, I_S) \;+\; \gamma\, R_{\mathrm{pref}}(I_S) .
  \label{eq:r-combined}
\end{equation}
Nothing else in Sections~\ref{subsec:mdp}--\ref{subsec:ppo} changes: the rollouts, the teacher
reference, the advantage and the PPO update are identical, and $R_{\mathrm{pref}}$ is evaluated on
the same decoded image at one extra function evaluation per rollout, with no reference image and no
gradient. We use the JPEG-encoded size and a counting-based sharpness score built from the width of
luminance transitions, both piecewise constant, so a pathwise method receives zero gradient
(Appendix~\ref{app:preference}).

This is not a new capability: the score-function estimator has always admitted such rewards, and
DDPO~\citep{Black2023DDPO} already optimizes JPEG compressibility. What the terminal formulation
contributes is that distillation and preference are the same kind of objective, so combining them
needs no second loss, optimizer or surrogate.

Two practical points govern \eqref{eq:r-combined}. Advantages are standardized after the two terms
are summed, so the preference's influence is set by $\gamma$ times the spread of $R_{\mathrm{pref}}$
across the rollout batch relative to that of $R$, not by their absolute scales. And the two conflict
by construction, since the teacher reference does not have the preferred property, so the optimum is
a compromise in which the distillation term anchors the student; preference rewards are proxies whose
cheapest optimum is often degenerate (Appendix~\ref{app:overopt}). That no change to the procedure is
needed we check on the COCO student, trained under the settings and budget of
Table~\ref{tab:matched} with the added term as the only difference:
Figure~\ref{fig:preference-progression} follows two prompts and seeds through training under each
preference, which pull in opposite directions on the same image statistic. It shows ease of use, not
how much fidelity a preference costs.

\section{Experiments}
\label{sec:experiments}

\subsection{Setup}
\label{subsec:setup}

\textbf{Teachers.} We distill a class-conditional EDM teacher~\citep{karras2022elucidating} on
ImageNet-64 in pixel space from $256$ to $8$ steps, and Stable Diffusion
v1.5~\citep{rombach2022ldm} in latent space from $50$ to $5$ steps, evaluated on
COCO~\citep{lin2015microsoftcoco}; SD v1.5 rather than
SDXL~\citep{podell2023sdxlimprovinglatentdiffusion} because every experiment in this paper,
including every retrained baseline, runs on one NVIDIA A100 (80\,GB).

\textbf{Metrics and protocol.} We report FID~\citep{heusel2018ganstrainedtimescaleupdate},
PRDC~\citep{naeem2020reliablefidelitydiversitymetrics} and, on COCO,
CLIPScore~\citep{hessel2022clipscorereferencefreeevaluationmetric}, treating Recall and Coverage as
primary alongside FID because reward-driven objectives admit reward-maximizing solutions with
degenerate diversity. FID uses $10{,}000$ generated samples against $10{,}000$ real validation images
on ImageNet-64 and $5{,}000$ validation prompts on COCO. Absolute values under this protocol are
higher than published ones (the $256$-step teacher reaches $13.74$ against a published $1.36$), so
they are comparable only within a table and published numbers are kept separate
(Appendix~\ref{app:published}). All reported checkpoints are the last checkpoint of the budget, ours
and the baselines' alike. Appendix~\ref{app:protocol} gives the full protocol.

\textbf{Student step count.} $M$ is bounded below by the need for a horizon over which the policy can
explore, and above by estimator variance and per-iteration cost. We use $M=8$ on ImageNet-64 and
$M=5$ on COCO; Appendix~\ref{app:step_count} reports the sweep over $M\in\{1,2,4,8,16\}$ behind the
ImageNet-64 choice.

\textbf{Training and cost accounting.} ReDiF trains a LoRA adapter of rank~4, with
effective batch $32$ on ImageNet-64 and $8$ on COCO, from conditioning pools of $1{,}000$
(noise, class-label) pairs and $160$ prompts. No real images enter training and no auxiliary network
is trained: each rollout is scored against a reference produced by the teacher's own sampler during
training, so no paired dataset exists at any point. Those teacher evaluations are the largest single
term in the training cost and are included in the PFLOPs column (Appendix~\ref{app:cost_model}). Two reward configurations are carried forward from
Appendix~\ref{app:reward_ablation}: \textbf{R\textsubscript{1}} is CLIP alone, the cheapest instance
of Section~\ref{subsec:reward}, and \textbf{R\textsubscript{2}} sums CLIP, DINOv3, PE and LPIPS with
equal weights. R\textsubscript{1} is the core reward function and works well on its own, which is why
we carry it throughout as the single-encoder reference. R\textsubscript{2} is our main reward: it
tracks R\textsubscript{1} closely, is ahead of it in places, and because a sample must satisfy four
encoders at once rather than one, it is the more robust of the two against the single-encoder
shortcuts behind reward hacking (Appendix~\ref{app:overopt}).
Hyperparameters are in Appendix~\ref{app:implementation}.

\subsection{Comparison in the Low-Training Regime}
\label{subsec:matched}

Our claim is about the low-training regime: what an objective delivers on one GPU in a few hours, not
what it reaches at convergence. Table~\ref{tab:matched} compares methods inside it. Every row is
trained by us from the same teacher initialization, on the same A100, at the same student step count,
from the authors' official implementation adapted to a LoRA student, at the hyperparameters their own
papers report (Appendix~\ref{app:baseline_lr}). CTM and TFD are absent: no setting we tried put them at a
reasonable operating point under this budget, so we quote them only at their authors'
(Appendix~\ref{app:protocol}). Every row trains for the same number of epochs over the same
conditioning pool and reads the same pool of teacher samples, synthesized once before training; the
PFLOPs column is the analytical cost of reproducing a row from scratch, pool included, and is
defined in Appendix~\ref{app:cost_model}. Table~\ref{tab:published} is the
complementary view, in which each method is quoted at the budget its authors judged it to need.

\begin{table}[t]
\centering
\caption{\textbf{The low-training regime.} All rows are trained by us on one A100, from the same teacher initialization and for the same number of epochs, under an otherwise identical protocol (Appendix~\ref{app:protocol}). PFLOPs is the analytical cost of reproducing a row from scratch, including the teacher-sample pool that is synthesized once and read by every row and that dominates it; Appendix~\ref{app:cost_model} defines it. GPU-h is wall-clock on the same device; cost columns are not ranked. Pub.\ FID is the authors' own number under their protocol and step count and is \emph{not} comparable to the FID column. Mean $\pm$ sample sd, last checkpoint; per branch, best in bold, second best bold and dash-underlined, teacher rows unranked.}
\label{tab:matched}
\resizebox{\linewidth}{!}{
\begin{tabular}{lcccccccccc}
\toprule
\textbf{Method} & \textbf{Steps} & \textbf{Batch} &
\textbf{GPU-h} & \textbf{PFLOPs} & \textbf{P} $\uparrow$ & \textbf{R} $\uparrow$ & \textbf{D} $\uparrow$ & \textbf{C} $\uparrow$ & \textbf{FID} $\downarrow$ &
\textbf{Pub.\ FID} \\
\midrule
\multicolumn{11}{l}{\emph{(a) EDM teacher, ImageNet-64 ($256\rightarrow8$)}} \\
Teacher (256 steps, reference) & 256 & -- & -- & -- & 0.8310 & 0.7103 & 0.8930 & 0.7831 & 13.74 & 1.36 \\
Teacher (8 steps)              & 8 & -- & -- & -- & 0.5308 & 0.4780 & 0.3910 & 0.5326 & 35.04 & -- \\
\hdashline
Progressive Distill.~\citep{salimans2022progressive} & $256\!\rightarrow\!8$ & 32 & $3.39$ & $1137.6$ & $0.5995 \pm 0.0017$ & $0.5591 \pm 0.0073$ & $0.6171 \pm 0.0010$ & $0.6770 \pm 0.0001$ & $38.02 \pm 0.19$ & 15.39 \\
Consistency Distill.~\citep{song2023consistency} & $256\!\rightarrow\!8$ & 32 & $3.30$ & $1138.5$ & $0.7110 \pm 0.0028$ & $0.5387 \pm 0.0190$ & $0.4004 \pm 0.0043$ & $0.6219 \pm 0.0051$ & $22.24 \pm 0.68$ & 6.20 \\
DMD~\citep{yin2024DMD} & $256\!\rightarrow\!8$ & 128 & $3.25$ & $1191.4$ & $0.7327 \pm 0.0043$ & $0.4596 \pm 0.0072$ & \second{0.7751 \pm 0.0031} & $0.7602 \pm 0.0043$ & $32.83 \pm 0.66$ & 2.62 \\
DMD2~\citep{yin2024IDMD} & $256\!\rightarrow\!8$ & 128 & $4.07$ & $1365.9$ & \best{0.7791 \pm 0.0039} & $0.6162 \pm 0.0058$ & $0.5349 \pm 0.0016$ & $0.7881 \pm 0.0034$ & $25.67 \pm 0.45$ & 1.28 \\
SiD~\citep{zhou2024sid} & $256\!\rightarrow\!8$ & 32 & $3.93$ & $1153.5$ & $0.7054 \pm 0.0037$ & $0.5453 \pm 0.0001$ & $0.6978 \pm 0.0006$ & $0.7621 \pm 0.0032$ & $40.31 \pm 0.06$ & 1.524 \\
SiDA~\citep{zhou2024sida} & $256\!\rightarrow\!8$ & 32 & $3.03$ & $1157.6$ & $0.7057 \pm 0.0057$ & $0.5281 \pm 0.0006$ & $0.7304 \pm 0.0023$ & $0.7313 \pm 0.0042$ & $35.36 \pm 0.13$ & 1.110 \\
\hdashline
Direct backprop.\ (R\textsubscript{1}: CLIP) & $256\!\rightarrow\!8$ & 32 & $3.36$ & $1156.0$ & $0.6747 \pm 0.0600$ & $0.5842 \pm 0.0037$ & $0.6273 \pm 0.0081$ & $0.7980 \pm 0.0320$ & $21.72 \pm 1.33$ & -- \\
Direct backprop.\ (R\textsubscript{2}: CLIP+DINOv3+PE+LPIPS) & $256\!\rightarrow\!8$ & 32 & $3.94$ & $1157.8$ & $0.6631 \pm 0.0024$ & $0.5783 \pm 0.0134$ & $0.4716 \pm 0.0072$ & $0.6823 \pm 0.0193$ & $19.76 \pm 0.24$ & -- \\
\hdashline
ReDiF (R\textsubscript{1}: CLIP) & $256\!\rightarrow\!8$ & 32 & $3.67$ & $1212.1$ & \second{0.7529 \pm 0.0180} & \best{0.6265 \pm 0.0213} & $0.6318 \pm 0.0291$ & \best{0.8223 \pm 0.0169} & \second{18.84 \pm 2.01} & -- \\
ReDiF (R\textsubscript{2}: CLIP+DINOv3+PE+LPIPS) & $256\!\rightarrow\!8$ & 32 & $3.84$ & $1213.5$ & $0.7432 \pm 0.0084$ & \second{0.6221 \pm 0.0158} & \best{0.7991 \pm 0.0101} & \second{0.8022 \pm 0.0129} & \best{18.60 \pm 1.63} & -- \\
\midrule
\multicolumn{11}{l}{\emph{(b) Stable Diffusion v1.5 teacher, COCO ($50\rightarrow5$)}} \\
Teacher (50 steps, reference) & 50 & -- & -- & -- & 0.6103 & 0.2029 & 0.8878 & 0.6832 & 24.59 & 8.74 \\
Teacher (5 steps) & 5 & -- & -- & -- & 0.4420 & 0.1762 & 0.6291 & 0.5813 & 31.45 & -- \\
\hdashline
Progressive Distill.~\citep{salimans2022progressive} & $50\!\rightarrow\!5$ & 8 & $0.93$ & $18.16$ & $0.4395 \pm 0.0012$ & $0.1653 \pm 0.0010$ & $0.6420 \pm 0.0011$ & $0.6032 \pm 0.0002$ & $31.60 \pm 2.38$ & 26.00 \\
Consistency Distill.~\citep{song2023consistency} & $50\!\rightarrow\!5$ & 8 & $0.68$ & $17.90$ & $0.4414 \pm 0.0021$ & $0.1742 \pm 0.0004$ & $0.6443 \pm 0.0071$ & $0.6023 \pm 0.0009$ & $31.79 \pm 0.22$ & 23.62 \\
DMD~\citep{yin2024DMD} & $50\!\rightarrow\!5$ & 8 & $0.58$ & $16.32$ & \best{0.5446 \pm 0.0002} & $0.1781 \pm 0.0009$ & $0.6720 \pm 0.0011$ & $0.6033 \pm 0.0020$ & $32.04 \pm 0.70$ & 11.49 \\
DMD2~\citep{yin2024IDMD} & $50\!\rightarrow\!5$ & 8 & $1.12$ & $19.03$ & $0.4655 \pm 0.0003$ & $0.1771 \pm 0.0013$ & $0.6420 \pm 0.0029$ & $0.6052 \pm 0.0017$ & $31.60 \pm 0.02$ & 8.35 \\
\hdashline
Direct backprop.\ (R\textsubscript{1}: CLIP) & $50\!\rightarrow\!5$ & 8 & $0.68$ & $16.48$ & $0.4474 \pm 0.0051$ & $0.1679 \pm 0.0046$ & $0.6427 \pm 0.0074$ & $0.6001 \pm 0.0061$ & $31.82 \pm 0.24$ & -- \\
Direct backprop.\ (R\textsubscript{2}: CLIP+DINOv3+PE+LPIPS) & $50\!\rightarrow\!5$ & 8 & $0.72$ & $16.52$ & $0.4716 \pm 0.0112$ & \best{0.1911 \pm 0.0034} & \second{0.6954 \pm 0.0203} & $0.6312 \pm 0.0071$ & $29.56 \pm 0.57$ & -- \\
\hdashline
ReDiF (R\textsubscript{1}: CLIP) & $50\!\rightarrow\!5$ & 8 & $0.54$ & $17.02$ & $0.4890 \pm 0.0062$ & $0.1864 \pm 0.0047$ & \best{0.7411 \pm 0.0087} & \second{0.6682 \pm 0.0065} & \second{27.42 \pm 0.84} & -- \\
ReDiF (R\textsubscript{2}: CLIP+DINOv3+PE+LPIPS) & $50\!\rightarrow\!5$ & 8 & $0.58$ & $17.04$ & \second{0.5193 \pm 0.0083} & \second{0.1896 \pm 0.0054} & $0.6301 \pm 0.0095$ & \best{0.6812 \pm 0.0056} & \best{26.36 \pm 0.78} & -- \\
\bottomrule
\end{tabular}
}
\end{table}

\textbf{Results.} On ImageNet-64 both ReDiF configurations reach the lowest FID, $18.60$ and $18.84$
against $22.24$ for the strongest baseline, consistency distillation; the gap of $3.64$ exceeds the
across-seed standard deviation of either, and ReDiF's Coverage exceeds the $256$-step teacher's
($0.8223$ against $0.7831$), while DMD2 keeps the highest Precision. On COCO the baselines barely move from the
$5$-step teacher (FID $31.60$--$32.04$ against $31.45$) while ReDiF reaches $26.36$ and $27.42$ with
the highest Coverage and Density. Held-out samples are in Appendix~\ref{app:qualitative_sd}.

\textbf{Policy optimization versus direct backpropagation.} The two direct-backpropagation rows share
the reward, initialization, schedule, conditioning pool, batch and budget of the corresponding ReDiF
rows; only the optimizer differs, its backward pass truncated to fit in 80\,GB. On ImageNet-64 policy
optimization wins on FID for both rewards ($18.84$ against $21.72$, $18.60$ against $19.76$); on COCO
it wins in FID, Precision, Coverage and Density, while direct backpropagation with R\textsubscript{2}
takes the highest Recall. The advantage is consistent in FID, not uniform across diversity metrics.

\textbf{Are the baselines simply under-trained?} Not relative to ReDiF: every row saw the same epochs
and the same pool of teacher samples. They are far from their published numbers because this
regime lies two to four orders of magnitude below those budgets, and because DMD, DMD2, SiD and SiDA
are run outside the one-step regime they were built for. Table~\ref{tab:published} shows
each method at its authors' budget, and Appendix~\ref{app:budget} plots FID against epochs: ReDiF
leads every baseline at every evaluated epoch, by the widest margin in the low-budget regime, and no
baseline reaches it anywhere in the measured range.

\subsection{ReDiF as a Plug-In Objective}
\label{subsec:plugin}

The construction needs only a stochastic policy over denoising transitions and a scalar terminal
reward, so it can be added to existing methods as well as substituted for them.
Table~\ref{tab:plugin_results} tests two integration modes on the EDM teacher. In the
\emph{structural} mode, progressive and consistency distillation keep their sampler or
parameterization, including the intermediate teachers and the EMA target, but their supervised loss
is replaced by the PPO objective; in the \emph{loss} mode, DMD2 keeps its auxiliary networks and
optimizes its loss jointly with the semantic reward. Every pair uses the budget of
Section~\ref{subsec:matched}, so the baseline rows repeat Table~\ref{tab:matched}.

\begin{table}[t]
\centering
\caption{ReDiF's PPO objective combined with existing methods on the EDM teacher, ImageNet-64
($256\rightarrow8$), under the same epoch budget as Section~\ref{subsec:matched}. Each ``+RL'' row keeps
its baseline's architectural modification. PFLOPs is total analytical training compute, teacher
forwards included. Mean $\pm$ sample sd over seeds, last checkpoint. The DMD2 rows keep the
auxiliary score network and are not cost-comparable to the rest; ReDiF alone reaches $18.60$. Within
each baseline/``+RL'' pair the better entry is bold; rows are compared within a pair, not across the
table.}
\label{tab:plugin_results}
\resizebox{\linewidth}{!}{
\begin{tabular}{llccccccc}
\toprule
\textbf{Method} & \textbf{Mode} & \textbf{Aux.\ net} & \textbf{PFLOPs} & \textbf{P} $\uparrow$ & \textbf{R} $\uparrow$ & \textbf{D} $\uparrow$ &
\textbf{C} $\uparrow$ & \textbf{FID} $\downarrow$ \\
\midrule
Progressive Distill.~\citep{salimans2022progressive} & --         & interm.\ teachers & $1137.6$ & $0.5995 \pm 0.0017$ & $\mathbf{0.5591 \pm 0.0073}$ & $0.6171 \pm 0.0010$ & $0.6770 \pm 0.0001$ & $38.02 \pm 0.19$ \\
\quad + RL (ours)                                    & structural & interm.\ teachers & $1342.4$ & $\mathbf{0.7321 \pm 0.0424}$ & $0.5250 \pm 0.1136$ & $\mathbf{0.6472 \pm 0.0601}$ & $\mathbf{0.7950 \pm 0.0451}$ & $\mathbf{34.08 \pm 2.71}$ \\
\hdashline
Consistency Distill.~\citep{song2023consistency}     & --         & EMA target        & $1138.5$ & $0.7110 \pm 0.0028$ & $0.5387 \pm 0.0190$ & $0.4004 \pm 0.0043$ & $0.6219 \pm 0.0051$ & $\mathbf{22.24 \pm 0.68}$ \\
\quad + RL (ours)                                    & structural & EMA target        & $1213.5$ & $\mathbf{0.7620 \pm 0.0221}$ & $\mathbf{0.5932 \pm 0.0520}$ & $\mathbf{0.6490 \pm 0.0680}$ & $\mathbf{0.8160 \pm 0.0300}$ & $22.68 \pm 6.80$ \\
\hdashline
DMD2~\citep{yin2024IDMD}                             & --         & score net + disc. & $1365.9$ & $0.7791 \pm 0.0039$ & $0.6162 \pm 0.0058$ & $0.5349 \pm 0.0016$ & $0.7881 \pm 0.0034$ & $25.67 \pm 0.45$ \\
\quad + RL (ours)                                    & loss       & score net + disc. & $1420.3$ & $\mathbf{0.7982 \pm 0.0126}$ & $\mathbf{0.6915 \pm 0.0109}$ & $\mathbf{0.7672 \pm 0.0171}$ & $\mathbf{0.8742 \pm 0.0051}$ & $\mathbf{21.32 \pm 0.86}$ \\
\bottomrule
\end{tabular}
}
\end{table}

In the loss mode, adding the terminal reward to DMD2 improves all five metrics, lowering FID from
$25.67$ to $21.32$ and raising Coverage from $0.7881$ to $0.8742$: the two signals are not redundant,
since at single-device batch sizes distribution matching alone reaches a worse point than ReDiF. The
combination nonetheless stays above ReDiF alone in FID and keeps its auxiliary-network cost. In the
structural mode, the terminal reward lowers progressive distillation's FID by $3.9$ and raises its
Precision and Coverage, at a larger seed spread; for consistency distillation it raises Precision,
Density and Coverage but leaves FID unchanged within noise, with a seed spread ten times larger
(Section~\ref{subsec:discussion}).

\subsection{Discussion and Limitations}
\label{subsec:discussion}

ReDiF produces high-quality students at low training cost: on a single GPU it reaches quality the
retrained baselines do not, while incorporating preferences through the same terminal objective and
with no extra machinery. Appendix~\ref{app:comparison} compares it with existing methods.

\textbf{Why the terminal reward may not help consistency distillation.} Table~\ref{tab:plugin_results}
shows no FID improvement and increased seed variance for consistency distillation, unlike progressive
distillation and DMD2. One explanation is that it constrains intermediate states to the teacher's
probability-flow trajectory more tightly than the other two, leaving little room for terminal-reward
exploration; ordering the three by that tightness matches the ordering of the gains. We have not
tested the mechanism directly, so we offer it as a hypothesis rather than a result.

\textbf{Diversity.} ReDiF achieves the highest Recall, Density, and Coverage on ImageNet-64 and the
highest Coverage on COCO. Since all samples use $\eta=0$, these metrics measure diversity across
noise and conditioning rather than policy stochasticity; we report the observed effect without
attributing it to a specific mechanism.

\textbf{Limitations.} The central limitation is reward hacking. Policy optimization reaches a useful
local optimum quickly, which is what enables low-cost training, but the reward can keep improving
after sample quality has plateaued: past the fixed budget, samples develop hacking artifacts
(Appendix~\ref{app:overopt}) that limit the attainable FID. The problem is more pronounced at larger
scales and resolutions, where convergence is faster and hacking emerges earlier and is harder to
detect. Multi-encoder rewards mitigate it partly, by requiring samples to satisfy several objectives
at once, but do not eliminate it. This limitation arises from the reward and
optimization process rather than terminal supervision itself; more direct suppression could
potentially raise the quality ceiling, as illustrated by DPOK~\citep{Fan2023DPOK} relative to
DDPO~\citep{Black2023DDPO}.

The advantage of ReDiF is also budget-dependent, and the method is most suitable for the few-step rather than one-step regime, where longer horizons provide greater room for policy exploration (Appendix~\ref{app:step_count}).

\section{Conclusion}
\label{sec:conclusion}

We introduced ReDiF, a reinforcement learning framework for diffusion step distillation that replaces
step-wise supervision with a terminal reward on the final sample. By scoring the endpoint with frozen
encoders it avoids step-wise targets, moving-target supervision and auxiliary-network training, and
trains only a LoRA adapter, on one GPU, from at most a thousand conditioning signals.

In the low-training regime, ReDiF achieves lower FID and higher Coverage than baselines trained on
the same teacher-sample budget, across both teachers, and it improves DMD2 when combined with its
loss. These gains matter most in low-budget settings and remain limited by reward over-optimization.
Since distillation is itself reward optimization, further preferences, including non-differentiable
ones, are added as extra reward terms, whose fidelity cost is left to future work.


\subsection*{AI use statement}

We used generative AI tools throughout this work: as writing assistants for language editing and
clarity of exposition, as a sounding board while developing the analysis, and as a coding assistant
during implementation. The research idea, the method, the experimental design and the claims are our
own. All AI-assisted text was reviewed and verified by the authors; every formal statement and proof
was stated and checked by the authors independently of any tool output; all code was read, executed
and validated by the authors; and all numbers quoted from prior work were verified against the
original publications. We take responsibility for the final content of this work, including any
errors.

\subsection*{Ethics statement}

This work introduces no new dataset and no human-subject data, and uses only publicly available
pretrained models and standard benchmarks. The method distills an existing teacher and inherits its
biases and failure modes; the distillation reward is an imitation objective whose optimum is the
teacher itself, so on its own it neither amplifies nor corrects what the teacher already does. The
preference term of Section~\ref{subsec:preference} changes this: it lets a student depart from its
teacher in a chosen direction. We demonstrate it only with stylistic properties (encoded size and a
sharpness score, Appendix~\ref{app:preference}). Fairness or debiasing criteria that cannot be written as
a reconstruction loss could in principle be optimized the same way, but we have not tested this and
do not claim it works; the same freedom that admits such objectives is what makes the reward
exploitable (Appendix~\ref{app:overopt}), and a debiasing reward would be subject to the same
over-optimization. The same mechanism could equally be used to install undesirable preferences. By
lowering the compute needed to obtain a fast sampler, the method broadens access to accelerated
generative models, which carries the general dual-use risks of cheap high-quality image synthesis.

\subsection*{Reproducibility statement}

The training procedure is given in Algorithm~\ref{alg:redif} (Appendix~\ref{app:algorithm}) and all
hyperparameters and hardware in Appendix~\ref{app:implementation}. The evaluation protocol, including
sample counts and reference sets, is stated in Section~\ref{subsec:setup} and
Appendix~\ref{app:protocol}. Assumptions and proofs for every formal statement are in
Appendix~\ref{app:theory}. The preference rewards are defined exactly, with all thresholds, in
Appendix~\ref{app:preference}. For retrained baselines, Appendix~\ref{app:baseline_lr} reports the
learning rate of every method and where it comes from, and Appendix~\ref{app:published} marks
every quoted number as author-disclosed or a third-party estimate. Source code, configuration files
and the launch scripts for every experiment are available at
\url{https://anonymous.4open.science/r/ReDiF-7BAF/}.


\bibliography{iclr2027_conference}
\bibliographystyle{iclr2027_conference}

\appendix

\section{Method Comparison Table}
\label{app:comparison}

Table~\ref{tab:related_works_comparison} places ReDiF against the families reviewed in
Section~\ref{sec:related_work} on seven axes: whether the method performs step distillation, whether it has been demonstrated at a
single step, whether it uses no real training images, whether it optimizes by reinforcement learning,
whether it trains from few conditioning signals, whether it admits non-differentiable rewards, and
whether it avoids heavy supervised reconstruction and large-scale synthetic datasets.

\begin{table}[t]
\centering
\small
\resizebox{\linewidth}{!}{
\begin{tabular}{lccccccc}
\toprule
Method & Step Distill. & One-Step & Data-Free & RL & Few Pr. & Non-Diff. & Low-Comp. \\
\midrule
\multicolumn{8}{l}{\emph{Trajectory matching}} \\
Progressive Distill.~\citep{salimans2022progressive} & \checkmark & \checkmark & \xmark & \xmark & \xmark & \xmark & \xmark \\
Guided Distill.~\citep{meng2023distillation} & \checkmark & \checkmark & \xmark & \xmark & \xmark & \xmark & \xmark \\
Consistency~\citep{song2023consistency} & \checkmark & \checkmark & \xmark & \xmark & \xmark & \xmark & \xmark \\
LCM / LCM-LoRA~\citep{luo2023lcm,luo2023lcmlora} & \checkmark & \checkmark & \xmark & \xmark & \xmark & \xmark & \xmark \\
CTM~\citep{kim2024ctm} & \checkmark & \checkmark & \xmark & \xmark & \xmark & \xmark & \xmark \\
InstaFlow~\citep{liu2024instaflow} & \checkmark & \checkmark & \xmark & \xmark & \xmark & \xmark & \xmark \\
\hdashline
\multicolumn{8}{l}{\emph{Distribution matching, adversarial and drifting}} \\
DMD / DMD2~\citep{yin2024DMD,yin2024IDMD} & \checkmark & \checkmark & \xmark & \xmark & \xmark & \xmark & \xmark \\
SiD~\citep{zhou2024sid} & \checkmark & \checkmark & \checkmark & \xmark & \xmark & \xmark & \xmark \\
SiDA~\citep{zhou2024sida} & \checkmark & \checkmark & \xmark & \xmark & \xmark & \xmark & \xmark \\
ADD / SD-Turbo~\citep{sauer2023adversarialdiffusiondistillation} & \checkmark & \checkmark & \xmark & \xmark & \xmark & \xmark & \xmark \\
Guided SiD~\citep{zhouguided} & \checkmark & \checkmark & \checkmark & \xmark & \xmark & \xmark & \xmark \\
TFD~\citep{zhang2026tfd} & \checkmark & \checkmark & \xmark & \xmark & \xmark & \xmark & \xmark \\
\hdashline
\multicolumn{8}{l}{\emph{Data-free trajectory methods}} \\
BOOT~\citep{gu2023boot} & \checkmark & \checkmark & \checkmark & \xmark & \xmark & \xmark & \xmark \\
DKDM~\citep{xiang2025dkdm} & \checkmark & -- & \checkmark & \xmark & \xmark & \xmark & \xmark \\
PMDD~\citep{tran2025progressive} & \checkmark & -- & \checkmark & \xmark & \xmark & \xmark & \xmark \\
\hdashline
\multicolumn{8}{l}{\emph{Reward-based}} \\
DDPO~\citep{Black2023DDPO} & \xmark & -- & \checkmark & \checkmark & \checkmark & \checkmark & \checkmark \\
RLCM~\citep{oertell2024rlconsistencymodelsfaster} & \xmark & -- & \checkmark & \checkmark & \checkmark & \checkmark & \checkmark \\
DRaFT~\citep{clark2024draft} & \xmark & -- & \checkmark & \xmark & \checkmark & \xmark & \checkmark \\
RG-LCD~\citep{li2024reward} & \checkmark & -- & \xmark & \xmark & \xmark & \xmark & \xmark \\
\hdashline
ReDiF (ours) & \checkmark & \xmark & \checkmark & \checkmark & \checkmark & \checkmark & \checkmark \\
\bottomrule
\end{tabular}
}
\caption{Comparison of diffusion acceleration and distillation methods. Data-Free indicates that no real images are used for training: SiD and Guided SiD qualify but retain an auxiliary score network; SiDA and ADD do not, because their discriminators are trained against real data; TFD does not, because its drifting field is estimated against a real-data reference set where available; RG-LCD does not, because its consistency loss is computed on real image--text data. The reward-based alignment methods use prompts only. ReDiF uses real images only for evaluation and to fix the training length once (Appendix~\ref{app:implementation}). One-Step indicates that the source paper demonstrates a single-step student; \emph{--} marks methods for which it is not reported or not applicable, and ReDiF is marked \xmark\ because it targets the few-step rather than the one-step regime (Appendix~\ref{app:step_count}). Low-Comp.\ indicates methods that avoid heavy supervised reconstruction and large-scale synthetic datasets.
}
\label{tab:related_works_comparison}
\end{table}

\section{Supporting Analysis}
\label{app:theory}

This appendix collects the formal statements used in Section~\ref{sec:method}. We keep only
statements that are either standard or proved in full here; the remaining arguments of
Section~\ref{subsec:cost} are motivation and are tested empirically in Section~\ref{sec:experiments}.
Throughout, $\{t_k\}_{k=0}^{M}$ is the student schedule, $\Phi_{\theta,k}$ the student map from
$x_{t_k}$ to $x_{t_{k-1}}$, $\Phi^{*}_k$ the teacher run over the same interval with its own finer
steps, and $x_0^{\theta},x_0^{*}$ the terminal samples from a shared $(x_T,c)$.

\subsection{Error Amplification along the Trajectory}
\label{app:amplification}

\begin{lemma}[Error amplification]
\label{lem:amplification}
Suppose each teacher map $\Phi^{*}_k$ is $L_k$-Lipschitz and the student deviates from it by at most
$\varepsilon_k=\sup_x\|\Phi_{\theta,k}(x)-\Phi^{*}_k(x)\|$ over the states it visits. Then
\[
\|x_0^{\theta}-x_0^{*}\|\le\sum_{k=1}^{M}\varepsilon_k\prod_{j<k}L_j\le\Gamma\max_k\varepsilon_k,
\qquad \Gamma:=\sum_{k=1}^{M}\prod_{j<k}L_j .
\]
If $L_j\ge1$ for all $j$ then $\Gamma\ge M$, with equality exactly when $L_j=1$ for $j<M$; if
$L_j\ge L>1$ for all $j$ then $\Gamma\ge(L^{M}-1)/(L-1)$.
\end{lemma}

\begin{proof}
Let $e_k=\|x^{\theta}_{t_k}-x^{*}_{t_k}\|$, with $e_M=0$ since both chains start from the same $x_T$.
Adding and subtracting the teacher map applied to the student's state,
\[
x^{\theta}_{t_{k-1}}-x^{*}_{t_{k-1}}
=\big[\Phi_{\theta,k}(x^{\theta}_{t_k})-\Phi^{*}_k(x^{\theta}_{t_k})\big]
+\big[\Phi^{*}_k(x^{\theta}_{t_k})-\Phi^{*}_k(x^{*}_{t_k})\big],
\]
where the first bracket is bounded by $\varepsilon_k$ and the second by $L_ke_k$. Hence
$e_{k-1}\le L_ke_k+\varepsilon_k$, and unrolling from $e_M=0$ gives
$e_0\le\varepsilon_1+L_1\varepsilon_2+L_1L_2\varepsilon_3+\dots=\sum_k\varepsilon_k\prod_{j<k}L_j$.
Each summand of $\Gamma$ is a product of Lipschitz constants, the $k=1$ term being the empty product
$1$. If all $L_j\ge1$ every summand is at least $1$, so $\Gamma\ge M$, with equality exactly when
every product equals $1$. If $L_j\ge L>1$, the summands are at least $L^{k-1}$ and the geometric sum
gives the last bound.
\end{proof}

The bound is an upper bound, so $\varepsilon_k\le\delta/\Gamma$ is \emph{sufficient} for terminal
accuracy $\delta$, not necessary. Its use in Section~\ref{subsec:cost} is accordingly limited: it
shows what per-step control guarantees, and why guaranteeing terminal accuracy through per-step
losses is demanding when $\Gamma$ is large. If the fitted per-step error decreases as $v/\sqrt{n}$ with
$n$ samples, this sufficient condition is met once $n\gtrsim\Gamma^{2}v^{2}/\delta^{2}$; the rate is
an assumption about the fitting procedure, not a result.

\subsection{Moving Targets}
\label{app:moving_targets}

\begin{remark}
\label{rem:moving}
Suppose the target is available only through an estimate $\hat y_k$ with systematic error $b_k$ that
does not decrease with the number of samples drawn from it. Then the per-step deviation cannot fall
below $b_k$, and the guarantee of Lemma~\ref{lem:amplification} cannot fall below $\Gamma\max_kb_k$
at any sample budget. For reconstruction and progressive targets, which are deterministic functions
of a frozen teacher, $b_k=0$. For an EMA target, which lags the student by construction, and for a
fake-score network fitted to a generator that changes at every update, the systematic component is
refreshed as training proceeds rather than averaged away. This is an informal argument about the
guarantee, not a lower bound on the error of any specific method.
\end{remark}

\subsection{Terminal Credit Assignment}
\label{app:credit}

\begin{proposition}[Policy gradient with a terminal reward]
\label{prop:credit}
For the Gaussian policy of Section~\ref{subsec:mdp}, a bounded terminal reward $R(x_0)$ and any
baseline $b$ that does not depend on the actions of the trajectory,
\[
\nabla_\theta J(\theta)
=\mathbb{E}\!\left[(R-b)\sum_{k=1}^{M}\nabla_\theta\log\pi_\theta\!\left(x_{t_{k-1}}\mid x_{t_k},t_k,c\right)\right].
\]
\end{proposition}

\begin{proof}
This is the likelihood-ratio (REINFORCE) gradient~\citep{williams1992simple} specialized to a
terminal reward. The trajectory density given $(x_T,c)$ is $p_\theta(\tau)=\prod_{k=1}^{M}\pi_\theta(x_{t_{k-1}}\mid
x_{t_k},t_k,c)$, and the dynamics contribute no other $\theta$-dependent factor, so
$\nabla_\theta\log p_\theta(\tau)=\sum_k\nabla_\theta\log\pi_\theta(x_{t_{k-1}}\mid x_{t_k},t_k,c)$. The
score-function identity $\nabla_\theta J=\mathbb{E}[R\,\nabla_\theta\log p_\theta(\tau)]$ follows by
exchanging differentiation and integration, which is justified by dominated convergence since $R$ is
bounded and the Gaussian score is integrable. The reward is terminal and multiplies the whole sum.
A baseline independent of the actions leaves the expectation unchanged because
$\mathbb{E}[\nabla_\theta\log p_\theta(\tau)]=0$. Nothing in the argument uses differentiability of
$R$, which is what permits the piecewise-constant rewards of Appendix~\ref{app:preference}.
\end{proof}

The batch-standardized advantage of Section~\ref{subsec:ppo} satisfies this condition only
approximately: $\mu_{\mathcal{B}}$ is computed from the same rollouts it is subtracted from, so it is
not independent of the actions, division by $\sigma_{\mathcal{B}}$ rescales the gradient by a quantity
that varies from iteration to iteration, and the clip at $A_{\max}$ truncates the largest deviations.
A leave-one-out mean and an unnormalized advantage would remove the first two effects; we follow DDPO
and use the batch statistics directly, accepting a small bias in exchange for variance reduction.

\subsection{Invariance of the Semantic Reward}
\label{app:semantic}

\begin{proposition}[First-order invariance]
\label{prop:semantic}
Let $f:\mathbb{R}^{D}\to\mathbb{R}^{d}$, $d<D$, be continuously differentiable and let the reward be
$R(I)=\mathcal{S}(z_T,f(I))$ for a fixed $z_T$ and a $C^{1}$ similarity $\mathcal{S}$. Then for every
$v\in\ker Df(I)$, $R(I+\epsilon v)=R(I)+o(\epsilon)$, and $\dim\ker Df(I)\ge D-d$. For a sum of
rewards built on encoders $f_1,\dots,f_K$, the same holds for every
$v\in\bigcap_i\ker Df_i(I)$.
\end{proposition}

\begin{proof}
By the chain rule $\nabla_IR=Df(I)^{\top}\nabla_z\mathcal{S}(z_T,f(I))$, which lies in the row space
of $Df(I)$ and is therefore orthogonal to $\ker Df(I)$. The directional derivative of $R$ along any
$v\in\ker Df(I)$ is thus zero, which gives the expansion. Rank--nullity gives
$\dim\ker Df(I)=D-\operatorname{rank}Df(I)\ge D-d$. For a sum of rewards the gradient is the sum of
the individual gradients, each orthogonal to its own kernel, so their sum is orthogonal to the
intersection.
\end{proof}

The statement concerns the reward, not the optimizer. It does not imply that the policy-gradient
estimator has lower variance, which depends on the dimension of the action space rather than on that
of the embedding. It is used in Section~\ref{subsec:reward} and Appendix~\ref{app:overopt} only to
explain which image changes the reward does not see, and why the reward can keep increasing while
image-level metrics do not.

\subsection{Scope}
\label{app:scope}

These statements are either standard (Proposition~\ref{prop:credit}) or elementary
(Lemma~\ref{lem:amplification}, Proposition~\ref{prop:semantic}), and Remark~\ref{rem:moving} is
informal. None is a convergence result for policy optimization, and none establishes that ReDiF
requires fewer updates or smaller batches than other methods, or bounds how far the optimal student
lies from the teacher; those are empirical claims, tested in Section~\ref{sec:experiments} and
Appendices~\ref{app:budget}, \ref{app:reward_curves} and~\ref{app:step_count}.

\section{Cost Model}
\label{app:cost_model}

Every row of Table~\ref{tab:matched} is trained on the same teacher, on the same sampler grid, at the
same student step count, from the same conditioning pool, for the same number of epochs over that
pool, on the same A100. An epoch is $32$ conditioning signals on ImageNet-64 and $16$ on COCO, so the
$100$- and $20$-epoch budgets are $3{,}200$ and $320$ training samples drawn from pools of $1{,}000$
(noise, class) pairs and $160$ prompts. Training compute follows from those choices; this section
states what the PFLOPs column of Table~\ref{tab:matched} counts.

\paragraph{Pool synthesis.} Training without real images begins by turning the conditioning pool into
teacher samples. A trajectory is $N_T$ teacher network evaluations, $N_T=2N-1=511$ on ImageNet-64
(Heun, $N=256$) and $N_T=2N=100$ on COCO ($N=50$, conditional and unconditional branch), and it
carries every grid point it passes: one pass over the pool supplies both the terminal references
ReDiF scores against and the trajectory targets progressive and consistency distillation regress on.
The pool is written once, before training, and read by every row that uses it. Because $N_T$ is two
orders of magnitude larger than the student step count $M$, this term dominates the totals in
Table~\ref{tab:matched}, and it is the same term for every row.

\paragraph{Training cost.} Let $C_S$ and $C_T$ be the forward cost of one student and one teacher
evaluation, $C_{\mathrm{enc}}$ the summed cost of the reward encoders over one image and
$C_{\mathrm{dec}}$ of one VAE decode; FLOPs are counted as $2\times$MACs and backward cost as twice
forward. Over $I$ iterations with $G$ rollouts of $M$ steps, $E$ inner epochs and $K$ encoders,
training costs
\[
C^{\mathrm{ReDiF}}_{\mathrm{train}}
=I\Big[\underbrace{GMC_S}_{\text{rollout}}+\underbrace{3EGMC_S}_{\text{surrogate}}
+\underbrace{\mathbb{1}[\beta>0]\,EGMC_T}_{\text{KL}}
+\underbrace{GKC_{\mathrm{enc}}}_{\text{reward}}
+\underbrace{\mathbb{1}[\text{latent}]\,GC_{\mathrm{dec}}}_{\text{decode}}\Big],
\]
with the student terms doubled under guidance (Section~\ref{subsec:cfg}). The KL term is the only
teacher call ReDiF makes during training and is absent on COCO, where $\beta=0$. The objectives
differ in which of these terms they carry. Progressive and consistency distillation replace the
reward and surrogate terms by a regression loss on stored teacher or EMA targets and add nothing
else. A distribution-matching iteration evaluates the teacher score and trains an auxiliary score
network, and in DMD2 and SiDA a discriminator, on samples drawn from the student during training;
those teacher calls are part of its training cost and are not served by the pool.

\paragraph{What the table reports.} The PFLOPs column of Table~\ref{tab:matched} is the analytical
cost of reproducing a row from scratch: pool synthesis plus that row's training cost. It is
dominated by the pool, which is shared, so the column should be read as the cost of obtaining a
student on one device rather than as a measure of how much optimization each objective performs.
The GPU-h column is the measured counterpart, wall-clock on the same device for the same runs.
\section{Quality as a Function of Training Budget}
\label{app:budget}

A single matched budget invites the objection that it is the one that favors us. This appendix
therefore reports quality along the budget axis rather than at one operating point.

\subsection{FID versus Training Epochs on ImageNet-64}
\label{app:fid_compute}

\begin{figure}[t]
    \centering
    \includegraphics[width=\columnwidth]{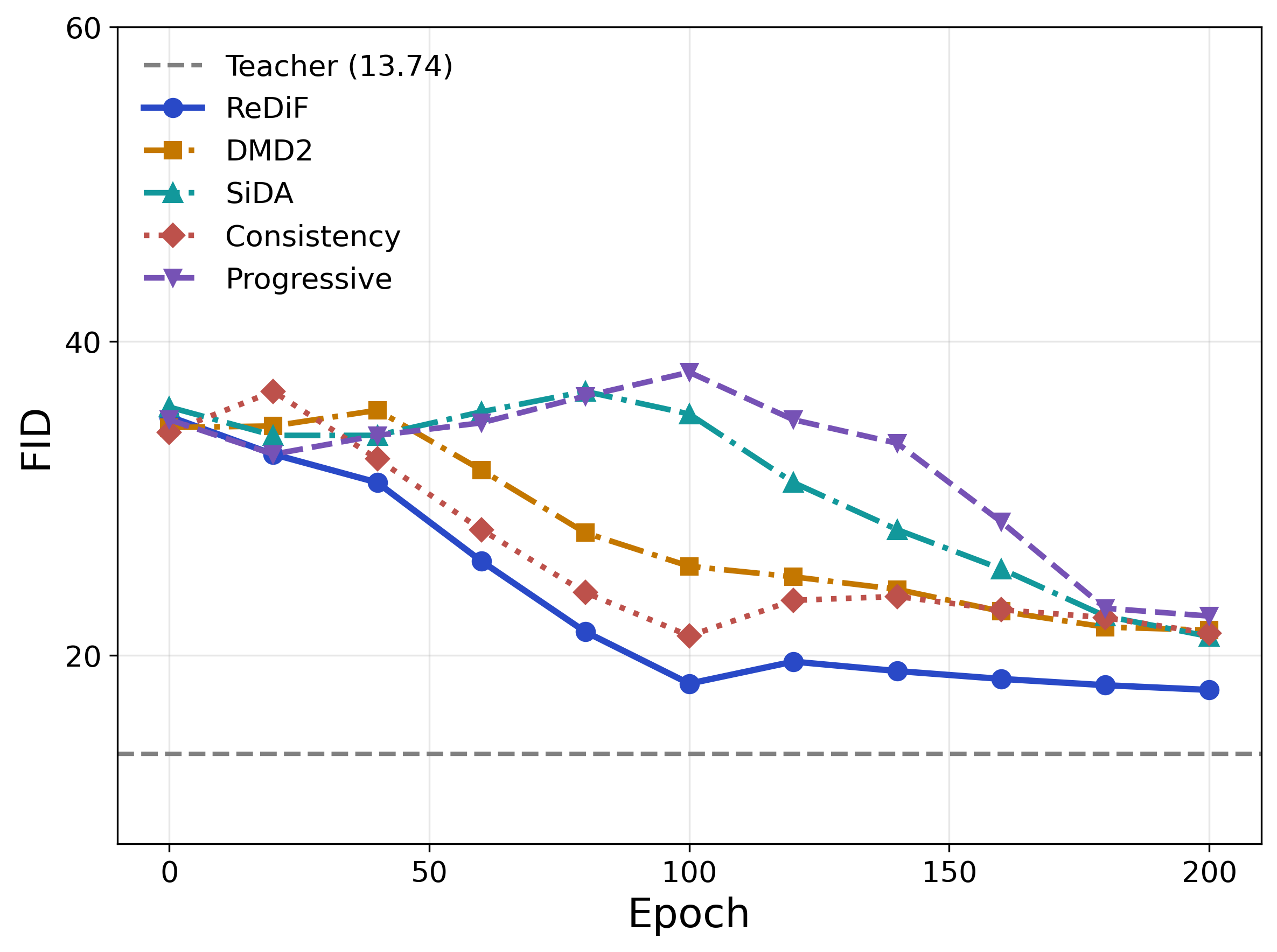}
    \caption{FID versus training epoch on ImageNet-64 ($256\rightarrow8$, EDM teacher). Markers denote
    evaluated checkpoints; the dashed line is the $256$-step teacher ($\mathrm{FID}=13.74$) under the
    same protocol. The horizontal axis counts epochs; because the rest of the protocol is held
    fixed, an epoch is the same number of training samples for every curve
    (Appendix~\ref{app:cost_model}).}
    \label{fig:budget_curves}
\end{figure}

Figure~\ref{fig:budget_curves} plots FID against training epoch for ReDiF and the four strongest
baselines of Table~\ref{tab:matched} on the EDM teacher ($256\rightarrow8$). All five students share
the teacher, the student step count, the checkpoint grid and the evaluation protocol of
Appendix~\ref{app:protocol}. DMD and SiD are omitted for legibility; they track DMD2 and SiDA at a
uniformly higher FID.

\paragraph{ReDiF leads across the whole range, and by the widest margin at low budgets.} ReDiF is
below every baseline at every evaluated epoch. It separates first: by epoch $60$ it is near $27$ FID
while consistency distillation, the best baseline, is near $28.5$ and the other three lie between
$31$ and $35$, and by epoch $100$ it reaches roughly $18.5$ while the best baseline is near $21$.
After that ReDiF is essentially flat, moving by less than $1.5$ FID over the remaining $100$ epochs,
while the baselines keep descending; none of them reaches it. At epoch $200$ ReDiF is near $18$ and
the closest three, SiDA, consistency distillation and DMD2, are between $20.8$ and $21.3$. The margin
of Table~\ref{tab:matched} is therefore widest in the low-training regime, where the baselines have
not yet recovered from their initialization, but it does not close: SiDA and progressive distillation
are still descending at the right edge of the plot, so we report the ordering over the measured range
and do not extrapolate past it.

\paragraph{An epoch is the same amount of data for every curve.} Every curve is trained on the same
teacher, the same conditioning pool and the same number of samples per epoch, so the horizontal axis
carries the same meaning for all of them. The objectives do not cost the same per epoch: the
regression losses of progressive and consistency distillation are the cheapest, and the
distribution-matching rows train an auxiliary network alongside the student
(Appendix~\ref{app:cost_model}). The ordering above is therefore one at equal data, not at equal
compute per epoch.

\paragraph{Two baselines are non-monotone.} SiDA and progressive distillation rise above their
initialization before improving, peaking near epochs $80$ and $100$ respectively at FID above $35$. A
run of either method stopped inside that window returns a checkpoint no better than its
initialization. Their entries in Table~\ref{tab:matched} should therefore be read as what these
objectives deliver at this budget; their results at their own budgets are in
Table~\ref{tab:published}.

\subsection{The SD v1.5 Teacher}
\label{app:budget_sd}

We did not run the equivalent sweep on the SD v1.5 teacher and make no quantitative claim about the
shape of its budget curve. What we did observe is that runs continued past the fixed budget show the
reward-hacking artifacts of Figure~\ref{fig:training_progress_oop}, so we expect the advantage in
Table~\ref{tab:matched}(b) to shrink or reverse at larger budgets. This is the reason we do not
present ReDiF as a drop-in replacement for distribution-matching distillation at large budgets.

\section{ReDiF Algorithm}
\label{app:algorithm}

Algorithm~\ref{alg:redif} summarizes the pipeline. The teacher is used only to draw a terminal
reference image for each rollout and, when $\beta>0$, to evaluate the KL term; it never supervises
intermediate states, and its outputs are never stored. The student receives supervision only after producing its final sample, and the
reward, optionally augmented with the preference term of \eqref{eq:r-combined}, is propagated to
all of its denoising actions by policy optimization.

\begin{algorithm}[t]
\caption{ReDiF: reinforcement learning based diffusion step distillation}
\label{alg:redif}
\begin{algorithmic}[1]
\Require Frozen teacher $G_T$; student $G_S$ initialized from $G_T$; conditioning set
$\mathcal{D}$; encoders $\{f_i\}_{i=1}^{K}$; weights $\{\lambda_i\}_{i=1}^{K}$; optional preference
reward $R_{\mathrm{pref}}$ with weight $\gamma$ ($\gamma=0$ by default); teacher steps $N$; student
steps $M$; rollouts per iteration $B$.
\Ensure Distilled student $G_S$.
\For{each optimization iteration}
    \State Sample a batch $\mathcal{B}$ of $B$ conditionings from $\mathcal{D}$.
    \State For each $(x_T,c)\in\mathcal{B}$, obtain a reference $I_T\leftarrow G_T(x_T,c,N)$ from the teacher's sampler and embed $z_T^{(i)}\leftarrow f_i(I_T)$; no dataset of teacher trajectories is assembled.
    \State Roll out one student trajectory per conditioning, $I_S\leftarrow G_S(x_T,c,M)$, storing the sampled latents and their log-densities.
    \State Compute $z_S^{(i)}\leftarrow f_i(I_S)$ and $R\leftarrow\sum_{i=1}^{K}\lambda_i\,\mathcal{S}_i(z_T^{(i)},z_S^{(i)})+\gamma R_{\mathrm{pref}}(I_S)$.
    \State Standardize over the batch: $\hat A=\operatorname{clip}\big((R-\mu_{\mathcal{B}})/(\sigma_{\mathcal{B}}+\varepsilon),-A_{\max},A_{\max}\big)$.
    \State For $E$ inner epochs, update the student by maximizing $\mathcal{L}=\mathcal{L}_{\mathrm{PPO}}-\beta D_{\mathrm{KL}}(\pi_\theta\|\pi_T)$.
\EndFor
\State \Return $G_S$.
\end{algorithmic}
\end{algorithm}

\section{Qualitative Training Progression}
\label{app:qualitative}

\begin{figure*}[t]
    \centering
    \setlength{\tabcolsep}{2pt}
    \renewcommand{\arraystretch}{1.2}
    \begin{tabular}{c|cccccc|c}
        & \textbf{Epoch 16} & \textbf{Epoch 18} & \textbf{Epoch 20} & \textbf{Epoch 25} & \textbf{Epoch 30} & \textbf{Epoch 35} & \textbf{Teacher Ref.} \\
        \midrule
        \rotatebox[origin=c]{90}{\textbf{ }} &
        \includegraphics[width=0.125\textwidth]{images/cat_progression/prompt3_epoch16.png} &
        \includegraphics[width=0.125\textwidth]{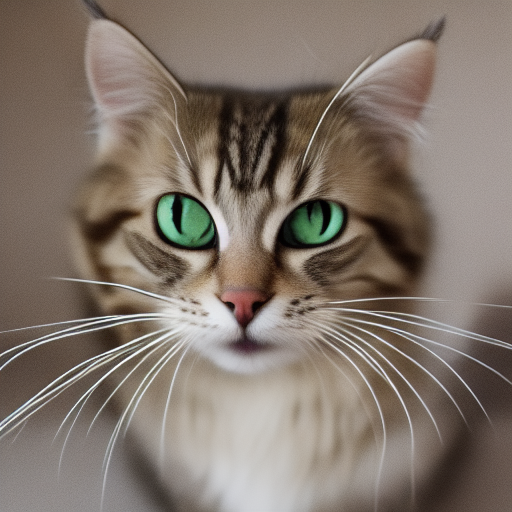} &
        \includegraphics[width=0.125\textwidth]{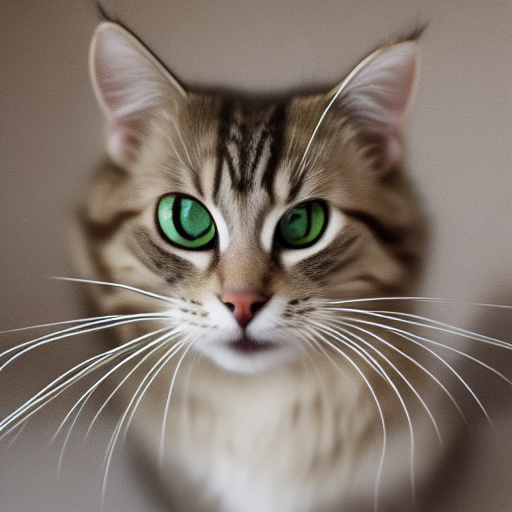} &
        \includegraphics[width=0.125\textwidth]{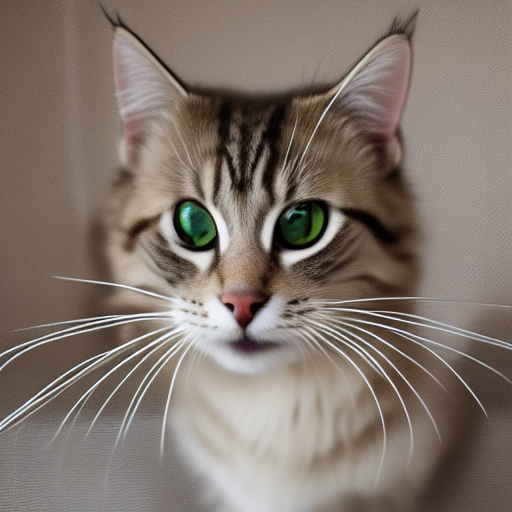} &
        \includegraphics[width=0.125\textwidth]{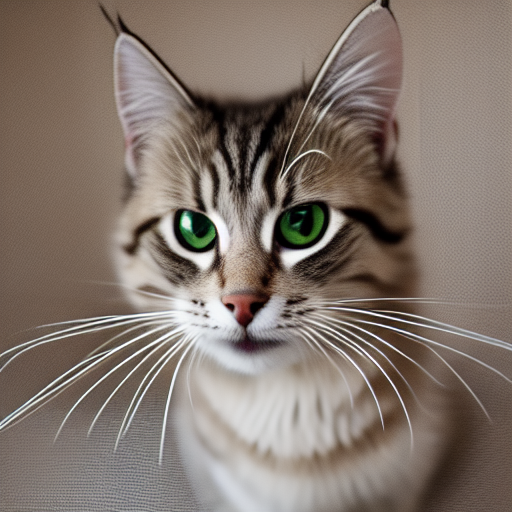} &
        \includegraphics[width=0.125\textwidth]{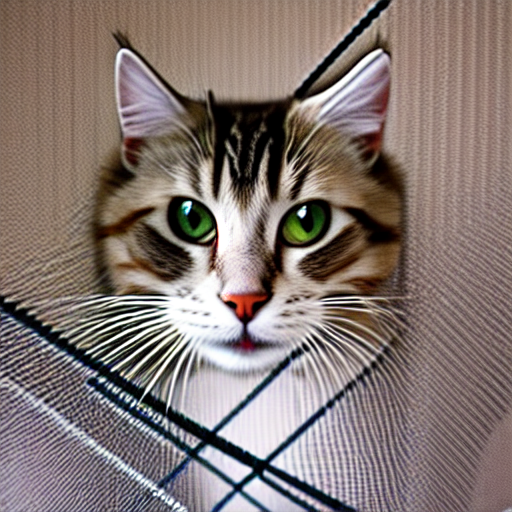} &
        \includegraphics[width=0.125\textwidth]{images/cat_progression/prompt3_teacher.png} \\
    \end{tabular}
    \caption{The SD v1.5 student of Figure~\ref{fig:training_progress} trained for $50$ epochs, well
    past the $20$-epoch budget of the reported COCO runs, for the same conditioning and seed; the
    rightmost column is the teacher's reference. Quality stays stable for a few epochs and then
    degrades as the student exploits directions the reward does not penalize.}
    \label{fig:training_progress_oop}
\end{figure*}

Figures~\ref{fig:training_progress} and~\ref{fig:training_progress_oop} follow one SD v1.5 student's
output for a fixed conditioning and seed across training, with the teacher's reference on the right;
holding the generation conditions fixed isolates the effect of the policy updates.
Figure~\ref{fig:training_progress} covers the $20$-epoch budget of the reported COCO runs: although
the only supervision is a scalar computed on the final image, detail and agreement with the teacher
improve within a few epochs. Figure~\ref{fig:training_progress_oop} continues the same run to $50$
epochs. The sample stays stable for a few epochs past the budget and then degrades, with structured
artifacts appearing by epoch $35$ that the reward does not penalize; this is the over-optimization
analyzed in Appendix~\ref{app:overopt}.

\section{Qualitative Samples, SD v1.5}
\label{app:qualitative_sd}

\begin{figure*}[t]
    \centering
    \setlength{\tabcolsep}{2pt}
    \renewcommand{\arraystretch}{1.1}
    \begin{tabular}{ccc}
        \multicolumn{3}{c}{\textbf{Teacher, 50 steps}} \\
        \includegraphics[width=0.30\textwidth]{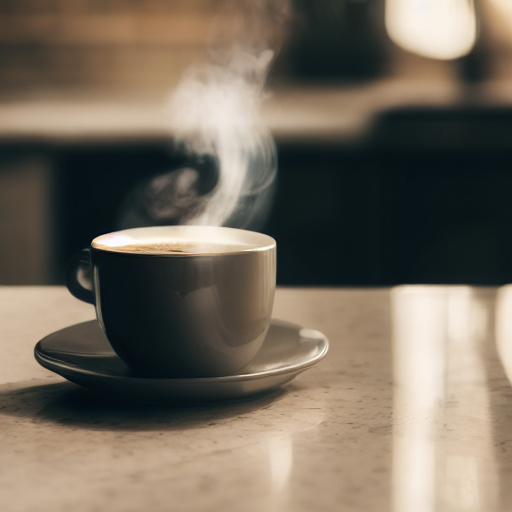} &
        \includegraphics[width=0.30\textwidth]{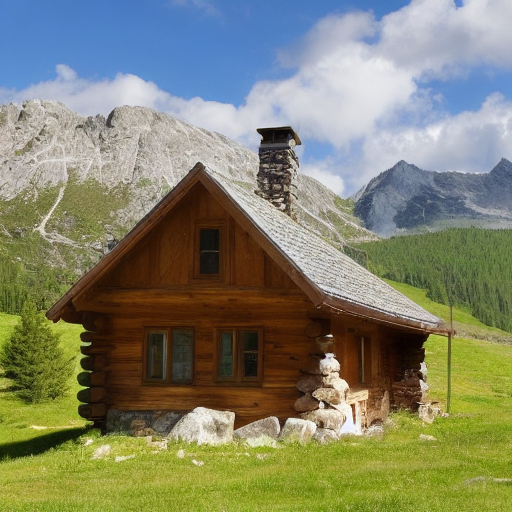} &
        \includegraphics[width=0.30\textwidth]{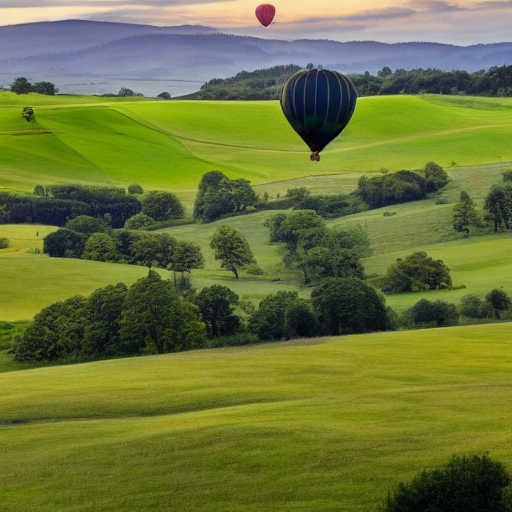} \\[0.6em]
        \multicolumn{3}{c}{\textbf{ReDiF student, 5 steps}} \\
        \includegraphics[width=0.30\textwidth]{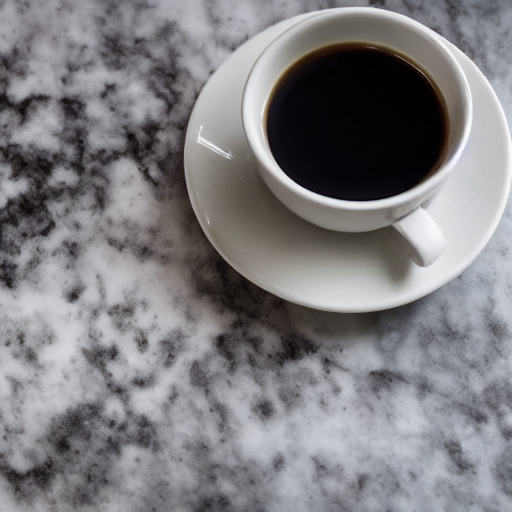} &
        \includegraphics[width=0.30\textwidth]{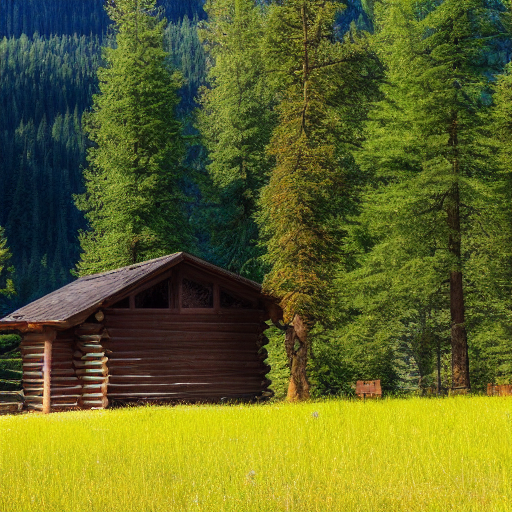} &
        \includegraphics[width=0.30\textwidth]{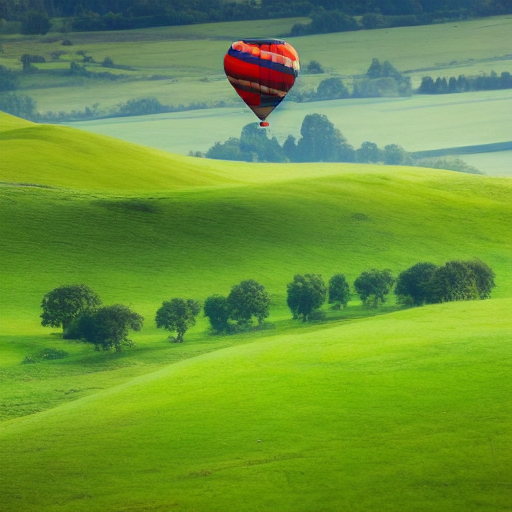} \\
        \small a steaming cup of coffee & \small a wooden cabin with a stone & \small a hot air balloon drifting \\
        \small on a marble countertop & \small chimney in an alpine meadow & \small over rolling green hills \\
    \end{tabular}
    \caption{Teacher and student samples, SD v1.5, first three held-out prompts.}
    \label{fig:sd_samples_a}
\end{figure*}

\begin{figure*}[t]
    \centering
    \setlength{\tabcolsep}{2pt}
    \renewcommand{\arraystretch}{1.1}
    \begin{tabular}{ccc}
        \multicolumn{3}{c}{\textbf{Teacher, 50 steps}} \\
        \includegraphics[width=0.30\textwidth]{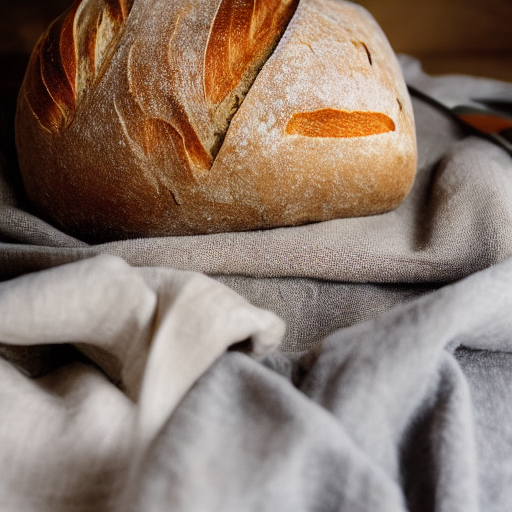} &
        \includegraphics[width=0.30\textwidth]{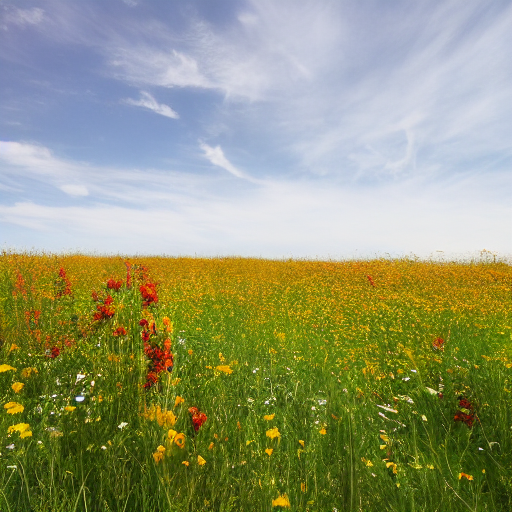} &
        \includegraphics[width=0.30\textwidth]{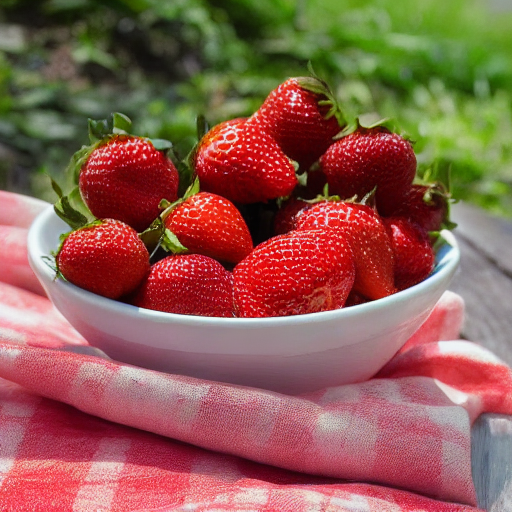} \\[0.6em]
        \multicolumn{3}{c}{\textbf{ReDiF student, 5 steps}} \\
        \includegraphics[width=0.30\textwidth]{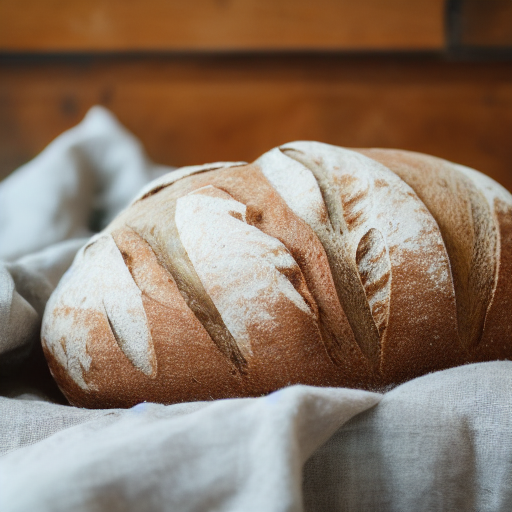} &
        \includegraphics[width=0.30\textwidth]{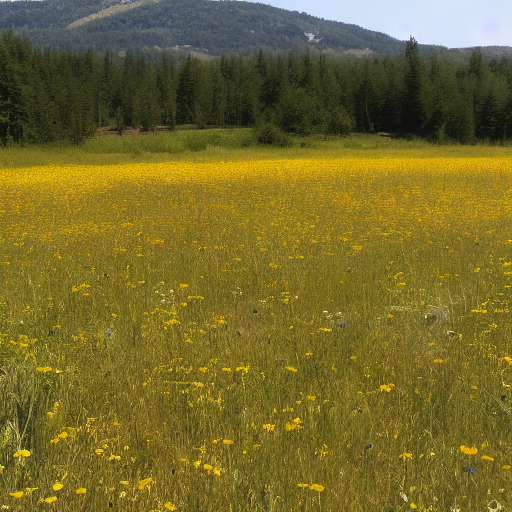} &
        \includegraphics[width=0.30\textwidth]{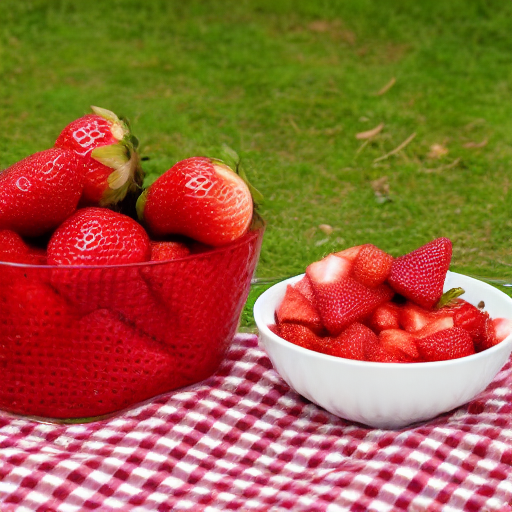} \\
        \small a rustic loaf of sourdough & \small a field of wildflowers & \small a bowl of fresh strawberries \\
        \small bread beside a linen cloth & \small under a bright midday sun & \small and cream on a picnic blanket \\
    \end{tabular}
    \caption{Teacher and student samples, SD v1.5, remaining three held-out prompts.}
    \label{fig:sd_samples_b}
\end{figure*}

Figures~\ref{fig:sd_samples_a} and~\ref{fig:sd_samples_b} place teacher samples above student samples
on the COCO branch. The teacher runs $50$ steps; the student runs $5$, trained with the
R\textsubscript{2} reward and taken at the last checkpoint of the fixed budget. Prompts are held out
from the training pool. Columns are not seed-matched, so the comparison is of output quality rather
than of the same latent rendered twice.

\section{Implementation Details}
\label{app:implementation}

All experiments run on a single NVIDIA A100 (80\,GB). The PPO hyperparameters of
Table~\ref{tab:implementation} follow the Stable Diffusion configuration of
DDPO~\citep{Black2023DDPO}, which optimizes the same clipped surrogate on the same backbone; we did
not search them. The effective batch is $32$ on ImageNet-64 and $8$ on COCO, in both cases without
gradient accumulation. The student is a LoRA adapter of rank~4 on the teacher and is optimized with PPO; the teacher
and the reward encoders stay frozen. Training uses fewer than one thousand conditioning signals,
resampled throughout optimization. No dataset of teacher trajectories is built: references come from
the teacher's own sampler inside the loop, and student rollouts are kept only for the inner epochs of
the iteration that produced them. When several encoders are used the reward is their
equally weighted sum. All reported metrics use the same inference settings and random seeds across
compared methods.

\textbf{Held-out monitoring and training length.} Because the reward keeps improving after sample
quality saturates (Appendix~\ref{app:overopt}), the reward cannot be used to decide when to stop. We
monitor FID on a small held-out set generated from conditioning excluded from training and from
evaluation, and use it once to fix the training length of each branch. All reported numbers, for
ReDiF and for every baseline, are the last checkpoint of that fixed budget.

The remaining hyperparameters are listed in Table~\ref{tab:implementation}.

\begin{table}[t]
\centering
\caption{Hyperparameters used in all reported ReDiF runs. Both branches use AdamW with
$\beta_1{=}0.9$, $\beta_2{=}0.999$, weight decay $10^{-4}$, $\epsilon{=}10^{-8}$ and gradient-norm
clipping at $1.0$. The preference runs of Appendix~\ref{app:preference} use the COCO column with the
preference weight added.}
\label{tab:implementation}
\begin{tabular}{lcc}
\toprule
\textbf{Hyperparameter} & \textbf{EDM / ImageNet-64} & \textbf{SD v1.5 / COCO} \\
\midrule
Optimizer & AdamW & AdamW \\
Trainable parameters & LoRA, rank / alpha 4 / 4 & LoRA, rank / alpha 4 / 4 \\
LoRA target modules & UNet attention ($q,k,v,\text{out}$) & UNet attention ($q,k,v,\text{out}$) \\
Learning rate & $3\times10^{-4}$ & $3\times10^{-4}$ \\
PPO clip range $\epsilon$ & $10^{-4}$ & $10^{-4}$ \\
Inner epochs per rollout & 4 & 4 \\
Rollouts per iteration & 32 & 16 \\
Advantage clip $A_{\max}$ & 5 & 5 \\
Effective batch (no accumulation) & 32 & 8 \\
Training epochs & 100 & 20 \\
Rollouts & 3200 & 320 \\
KL weight $\beta$ & 0.02 & 0 \\
Guidance scale (teacher / student) & -- & 5.0 / 5.0 \\
Preference weight $\gamma$ (Appendix~\ref{app:preference}) & -- & 2 \\
Conditioning pool & $1{,}000$ (noise, class) pairs & $160$ COCO prompts \\
Reported checkpoint & last & last \\
\bottomrule
\end{tabular}
\end{table}

\section{Reward Ablation}
\label{app:reward_ablation}

Section~\ref{subsec:reward} motivates computing the terminal reward in a representation space rather
than in pixels. This appendix tests that choice and selects the two configurations carried into
Section~\ref{sec:experiments} as R\textsubscript{1} and R\textsubscript{2}. Every configuration is
optimized by the identical PPO procedure, from the same initialization, with the same conditioning
pool, batch, budget and seeds, so the reward is the only difference between rows. We group rewards
into single encoders, combinations of encoders, and two non-semantic controls: the negative pixel MSE
between teacher and student images from the same noise, and a divergence-based reward.
Tables~\ref{tab:reward_ablation_imagenet} and~\ref{tab:reward_ablation_coco} report the two branches
separately. The ImageNet branch has no caption to align against, so it reports CLIP image-to-image
similarity to the teacher sample (CLIP-T) in place of CLIPScore.

\begin{table}[t]
\centering
\caption{\textbf{Reward ablation, EDM teacher on ImageNet-64.} Identical PPO settings,
initialization, conditioning pool, budget and seeds across rows; only the reward differs. Last
checkpoint, $M=8$, $\eta=0$, $10{,}000$ generated samples against $10{,}000$ real validation images;
the R\textsubscript{1} and R\textsubscript{2} rows are the ReDiF rows of Table~\ref{tab:matched}. Mean
$\pm$ sample sd over seeds. CLIP-T is the CLIP image-to-image similarity to the teacher sample; for rewards containing a CLIP term it is part of the training objective and not an independent measure. Best
entry in bold, second best bold and dash-underlined; teacher rows are not ranked.}
\label{tab:reward_ablation_imagenet}
\resizebox{\linewidth}{!}{
\begin{tabular}{lcccccc}
\toprule
\textbf{Reward} & \textbf{P} $\uparrow$ & \textbf{R} $\uparrow$ & \textbf{D} $\uparrow$ & \textbf{C} $\uparrow$ &
\textbf{CLIP-T} $\uparrow$ & \textbf{FID} $\downarrow$ \\
\midrule
Teacher (256 steps, reference) & 0.8310 & 0.7103 & 0.8930 & 0.7831 & -- & 13.74 \\
Teacher (8 steps)              & 0.5308 & 0.4780 & 0.3910 & 0.5326 & -- & 35.04 \\
\hdashline
\multicolumn{7}{l}{\emph{Single-encoder semantic rewards}} \\
CLIP (R\textsubscript{1}) & $0.7529 \pm 0.0180$ & \best{0.6265 \pm 0.0213} & $0.6318 \pm 0.0291$ & $0.8223 \pm 0.0169$ & \second{0.8836 \pm 0.0028} & \second{18.84 \pm 2.01} \\
DINOv3     & $0.7198 \pm 0.0161$ & $0.5654 \pm 0.0367$ & $0.5972 \pm 0.0418$ & $0.7828 \pm 0.0191$ & $0.8675 \pm 0.0058$ & $33.11 \pm 9.95$ \\
PE         & $0.6992 \pm 0.0152$ & $0.5685 \pm 0.0182$ & $0.5323 \pm 0.0247$ & $0.7452 \pm 0.0226$ & $0.8522 \pm 0.0054$ & $25.67 \pm 3.62$ \\
LPIPS      & $0.7058 \pm 0.0091$ & $0.5924 \pm 0.0125$ & $0.5409 \pm 0.0232$ & $0.7542 \pm 0.0206$ & $0.8728 \pm 0.0036$ & $21.11 \pm 1.55$ \\
\hdashline
\multicolumn{7}{l}{\emph{Combined semantic rewards}} \\
CLIP + DINOv3        & $0.7456 \pm 0.0051$ & $0.5861 \pm 0.0418$ & $0.5947 \pm 0.0031$ & $0.7792 \pm 0.0177$ & $0.8658 \pm 0.0060$ & $20.12 \pm 1.91$ \\
CLIP + PE            & \best{0.7841 \pm 0.0229} & $0.6152 \pm 0.0141$ & $0.6638 \pm 0.0402$ & $0.8098 \pm 0.0127$ & $0.8789 \pm 0.0045$ & $20.75 \pm 0.12$ \\
CLIP + LPIPS         & $0.6862 \pm 0.0082$ & $0.5364 \pm 0.0144$ & $0.5333 \pm 0.0118$ & $0.7419 \pm 0.0101$ & $0.8385 \pm 0.0080$ & $23.21 \pm 0.79$ \\
DINOv3 + PE          & \second{0.7819 \pm 0.0051} & $0.5271 \pm 0.0003$ & \second{0.7617 \pm 0.0073} & \best{0.8381 \pm 0.0019} & $0.8586 \pm 0.0024$ & $31.92 \pm 1.75$ \\
DINOv3 + LPIPS       & $0.7342 \pm 0.0036$ & $0.5926 \pm 0.0084$ & $0.6202 \pm 0.0139$ & $0.7798 \pm 0.0002$ & $0.8641 \pm 0.0082$ & $20.53 \pm 0.63$ \\
PE + LPIPS           & $0.7499 \pm 0.0021$ & $0.5911 \pm 0.0059$ & $0.6159 \pm 0.0031$ & $0.7937 \pm 0.0043$ & $0.8723 \pm 0.0005$ & $22.14 \pm 0.31$ \\
CLIP + DINOv3 + PE & $0.7281 \pm 0.0149$ & $0.6109 \pm 0.0061$ & $0.5723 \pm 0.0017$ & $0.7837 \pm 0.0063$ & $0.8738 \pm 0.0067$ & $19.42 \pm 2.05$ \\
CLIP + DINOv3 + PE + LPIPS (R\textsubscript{2}) & $0.7432 \pm 0.0084$ & \second{0.6221 \pm 0.0158} & \best{0.7991 \pm 0.0101} & $0.8022 \pm 0.0129$ & $0.8793 \pm 0.0030$ & \best{18.60 \pm 1.63} \\
\hdashline
\multicolumn{7}{l}{\emph{Non-semantic controls}} \\
Negative MSE (pixel)     & $0.7618 \pm 0.0091$ & $0.6018 \pm 0.0131$ & $0.6702 \pm 0.0090$ & \second{0.8268 \pm 0.0063} & $0.8801 \pm 0.0015$ & $20.55 \pm 3.05$ \\
Negative KL (divergence) & $0.7439 \pm 0.0111$ & $0.6151 \pm 0.0334$ & $0.6628 \pm 0.0042$ & $0.8191 \pm 0.0127$ & \best{0.8863 \pm 0.0032} & $23.01 \pm 2.35$ \\
\bottomrule
\end{tabular}
}
\end{table}

\begin{table}[t]
\centering
\caption{\textbf{Reward ablation, Stable Diffusion v1.5 on COCO ($50\rightarrow5$).} Same design as
Table~\ref{tab:reward_ablation_imagenet}: identical optimization, only the reward differs. Last
checkpoint, $M=5$, $5{,}000$ COCO validation prompts; the R\textsubscript{1} and R\textsubscript{2}
rows are the ReDiF rows of Table~\ref{tab:matched}. Mean $\pm$ sample sd over seeds. Best entry bold
in bold, second best bold and dash-underlined; teacher rows are not ranked.}
\label{tab:reward_ablation_coco}
\resizebox{\linewidth}{!}{
\begin{tabular}{lcccccc}
\toprule
\textbf{Reward} & \textbf{P} $\uparrow$ & \textbf{R} $\uparrow$ & \textbf{D} $\uparrow$ & \textbf{C} $\uparrow$ & \textbf{FID} $\downarrow$ & \textbf{CLIPScore} $\uparrow$ \\
\midrule
Teacher (50 steps, reference) & 0.6103 & 0.2029 & 0.8878 & 0.6832 & 24.59 & 0.3137 \\
Teacher (5 steps)             & 0.4420 & 0.1762 & 0.6291 & 0.5813 & 31.45 & 0.3041 \\
\hdashline
\multicolumn{7}{l}{\emph{Single-encoder semantic rewards}} \\
CLIP (R\textsubscript{1}) & \second{0.4890 \pm 0.0062} & $0.1864 \pm 0.0047$ & \best{0.7411 \pm 0.0087} & \second{0.6682 \pm 0.0065} & $27.42 \pm 0.84$ & $0.3038 \pm 0.0007$ \\
DINOv3     & $0.4310 \pm 0.0084$ & $0.1809 \pm 0.0061$ & $0.6406 \pm 0.0112$ & $0.5903 \pm 0.0075$ & $27.40 \pm 0.91$ & $0.3046 \pm 0.0004$ \\
PE         & $0.4498 \pm 0.0072$ & $0.1770 \pm 0.0057$ & $0.6733 \pm 0.0104$ & $0.5927 \pm 0.0069$ & $27.97 \pm 0.76$ & $0.3044 \pm 0.0001$ \\
LPIPS      & $0.4616 \pm 0.0059$ & $0.1770 \pm 0.0049$ & $0.6842 \pm 0.0091$ & $0.5813 \pm 0.0062$ & $27.64 \pm 0.69$ & \second{0.3053 \pm 0.0008} \\
\hdashline
\multicolumn{7}{l}{\emph{Combined semantic rewards}} \\
CLIP + DINOv3        & $0.4550 \pm 0.0068$ & $0.1813 \pm 0.0053$ & $0.6512 \pm 0.0101$ & $0.5965 \pm 0.0071$ & $27.61 \pm 0.82$ & $0.3042 \pm 0.0002$ \\
CLIP + PE            & $0.4497 \pm 0.0075$ & $0.1820 \pm 0.0058$ & $0.6420 \pm 0.0098$ & $0.6043 \pm 0.0065$ & $27.72 \pm 0.73$ & $0.3043 \pm 0.0005$ \\
CLIP + LPIPS         & $0.4410 \pm 0.0062$ & $0.1714 \pm 0.0051$ & $0.6402 \pm 0.0107$ & $0.5939 \pm 0.0070$ & $27.48 \pm 0.68$ & $0.3037 \pm 0.0007$ \\
DINOv3 + PE          & $0.4150 \pm 0.0071$ & \second{0.1865 \pm 0.0055} & $0.6368 \pm 0.0115$ & $0.5997 \pm 0.0068$ & $27.99 \pm 0.85$ & $0.3041 \pm 0.0009$ \\
DINOv3 + LPIPS       & $0.4503 \pm 0.0065$ & $0.1719 \pm 0.0052$ & $0.6451 \pm 0.0103$ & $0.6126 \pm 0.0074$ & $27.44 \pm 0.71$ & $0.3043 \pm 0.0003$ \\
PE + LPIPS           & $0.4743 \pm 0.0069$ & $0.1732 \pm 0.0059$ & \second{0.7406 \pm 0.0108} & $0.5182 \pm 0.0077$ & $27.97 \pm 0.79$ & $0.3051 \pm 0.0001$ \\
CLIP + DINOv3 + PE & $0.4490 \pm 0.0076$ & $0.1770 \pm 0.0056$ & $0.7340 \pm 0.0119$ & $0.5910 \pm 0.0068$ & \second{27.01 \pm 0.90} & $0.3035 \pm 0.0002$ \\
CLIP + DINOv3 + PE + LPIPS (R\textsubscript{2}) & \best{0.5193 \pm 0.0083} & \best{0.1896 \pm 0.0054} & $0.6301 \pm 0.0095$ & \best{0.6812 \pm 0.0056} & \best{26.36 \pm 0.78} & $0.3043 \pm 0.0003$ \\
\hdashline
\multicolumn{7}{l}{\emph{Non-semantic controls}} \\
Negative MSE (pixel)     & $0.4714 \pm 0.0067$ & $0.1690 \pm 0.0052$ & $0.6728 \pm 0.0102$ & $0.5838 \pm 0.0064$ & $29.80 \pm 1.12$ & \best{0.3058 \pm 0.0003} \\
Negative KL (divergence) & $0.4329 \pm 0.0069$ & $0.1593 \pm 0.0053$ & $0.6632 \pm 0.0117$ & $0.5954 \pm 0.0042$ & $30.13 \pm 1.07$ & $0.3050 \pm 0.0002$ \\
\bottomrule
\end{tabular}
}
\end{table}

\paragraph{Findings.} The four-encoder reward R\textsubscript{2} has the lowest FID on both branches,
and CLIP alone (R\textsubscript{1}) is second on ImageNet-64 and within about one FID of the best on
COCO at the lowest evaluation cost; these are the two configurations carried forward. The pixel
control behaves differently on the two branches. On COCO both non-semantic controls are worse in FID
than every semantic configuration ($29.80$ and $30.13$ against $26.36$--$27.99$). On ImageNet-64 the
pixel MSE is competitive ($20.55$): it is beaten by R\textsubscript{1}, R\textsubscript{2},
CLIP + DINOv3 + PE, CLIP + DINOv3 and, marginally, DINOv3 + LPIPS, but not by the other seven
semantic configurations, and single DINOv3 or DINOv3 + PE rewards are much worse. On this
branch the gain of a representation-space reward is therefore specific to the encoder, and it is
concentrated in configurations that contain CLIP. Combining encoders does not help monotonically:
CLIP + LPIPS is worse than CLIP alone on ImageNet-64, while the full combination R\textsubscript{2} is
the best. CLIPScore on COCO is essentially unchanged across rewards ($0.3035$--$0.3058$), so none of
the rewards trades text alignment for image quality. Each additional encoder costs one more forward
pass per trajectory (Appendix~\ref{app:cost_model}).

\subsection{Reward Optimization Curves}
\label{app:reward_curves}

\begin{figure*}[t]
    \centering
    \begin{subfigure}[b]{0.245\textwidth}
        \includegraphics[width=\textwidth]{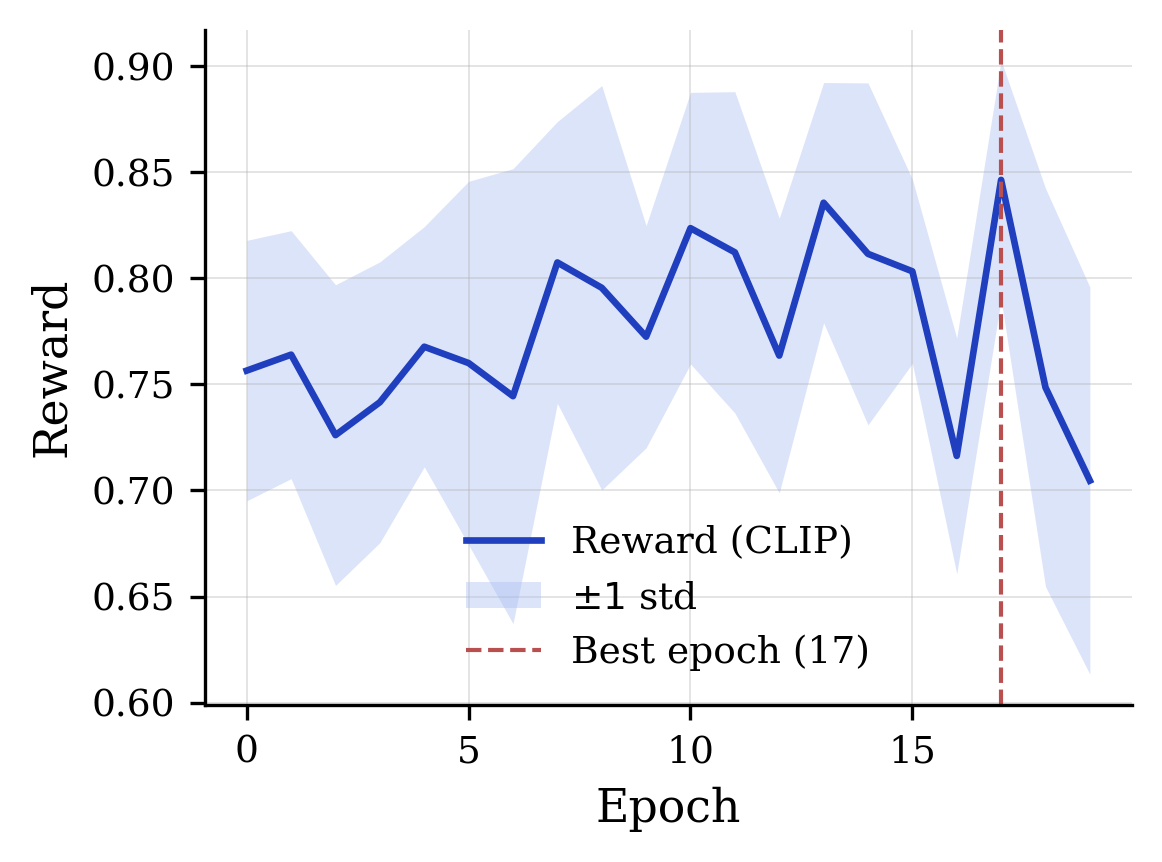}
        \caption{CLIP}\label{fig:rc_clip}
    \end{subfigure}\hfill
    \begin{subfigure}[b]{0.245\textwidth}
        \includegraphics[width=\textwidth]{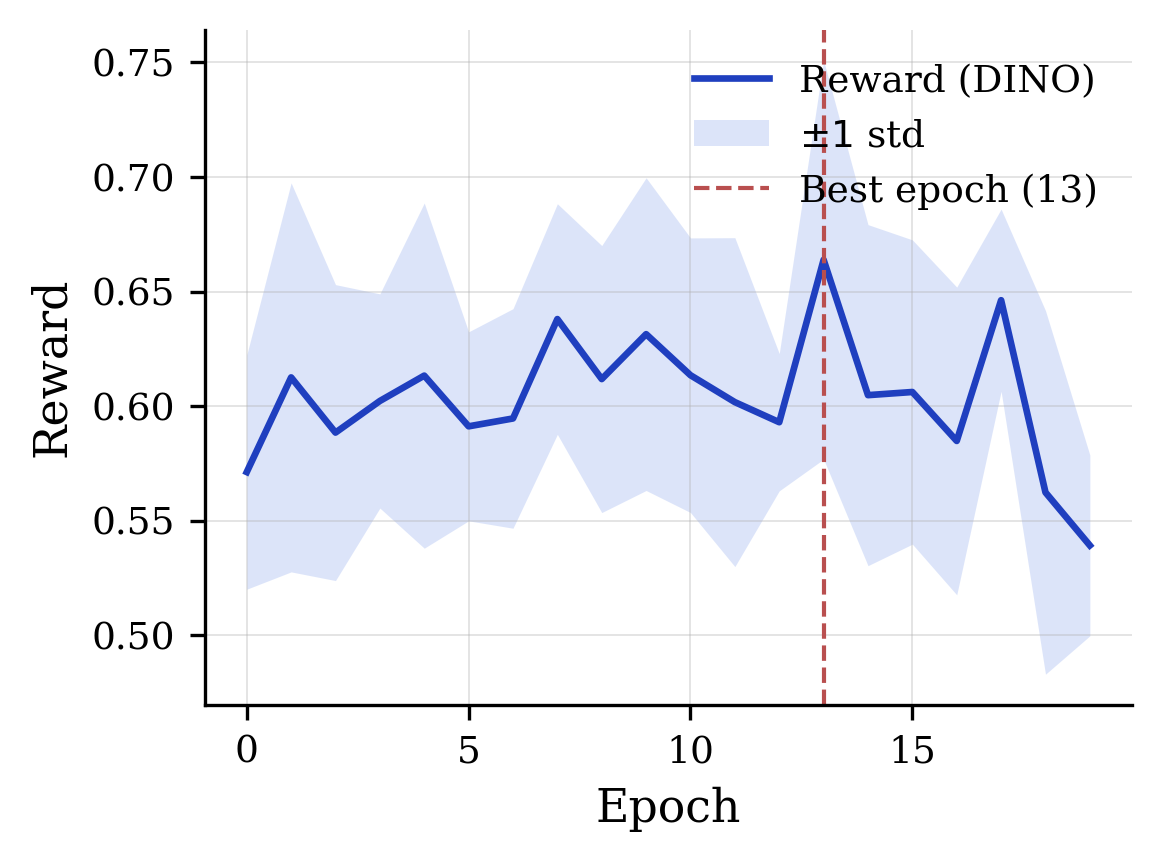}
        \caption{DINO}\label{fig:rc_dino}
    \end{subfigure}\hfill
    \begin{subfigure}[b]{0.245\textwidth}
        \includegraphics[width=\textwidth]{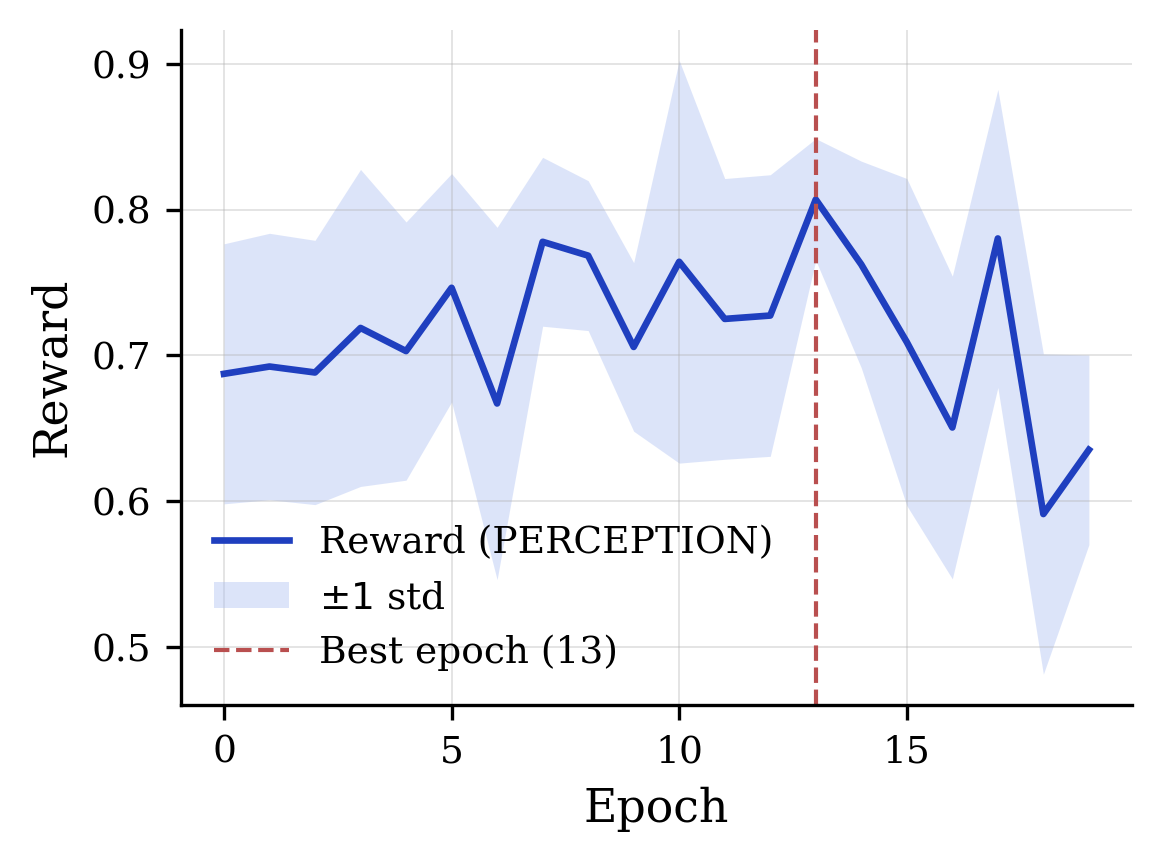}
        \caption{PE}\label{fig:rc_perc}
    \end{subfigure}\hfill
    \begin{subfigure}[b]{0.245\textwidth}
        \includegraphics[width=\textwidth]{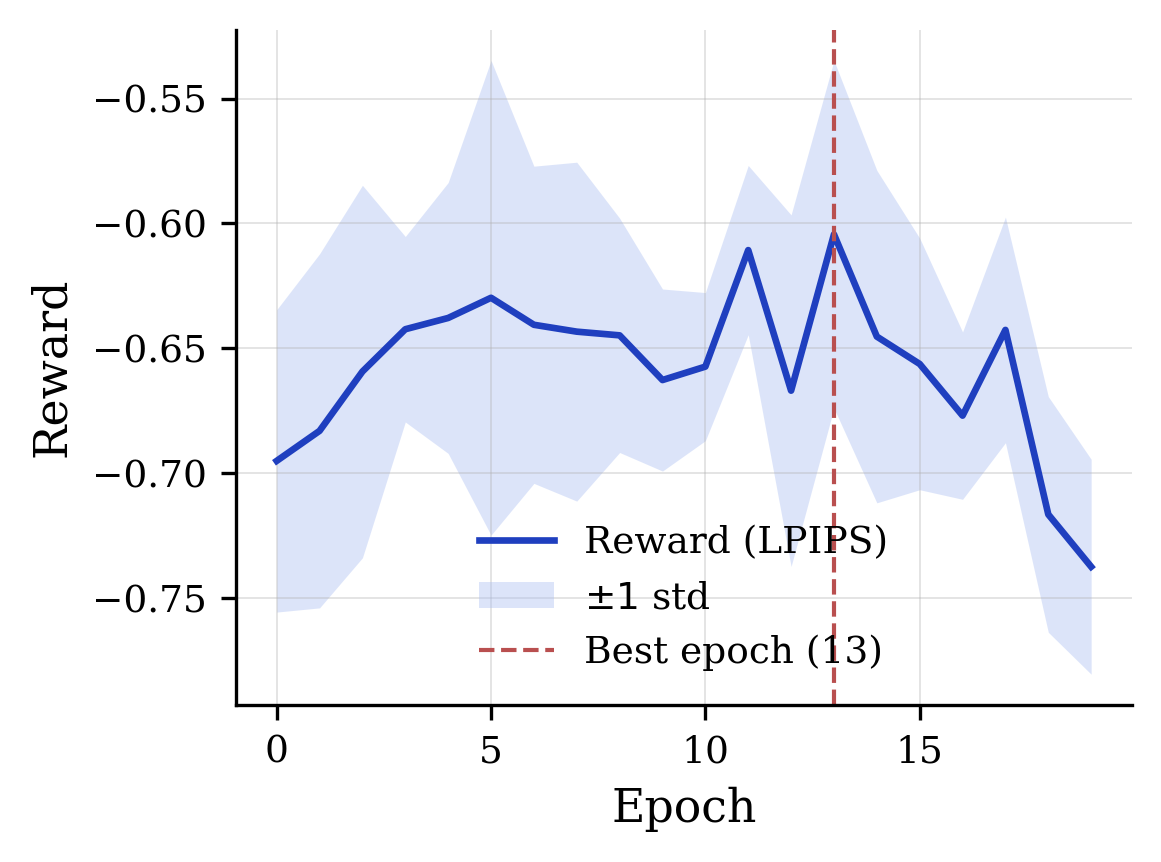}
        \caption{LPIPS}\label{fig:rc_lpips}
    \end{subfigure}

    \vspace{0.6em}

    \begin{subfigure}[b]{0.245\textwidth}
        \includegraphics[width=\textwidth]{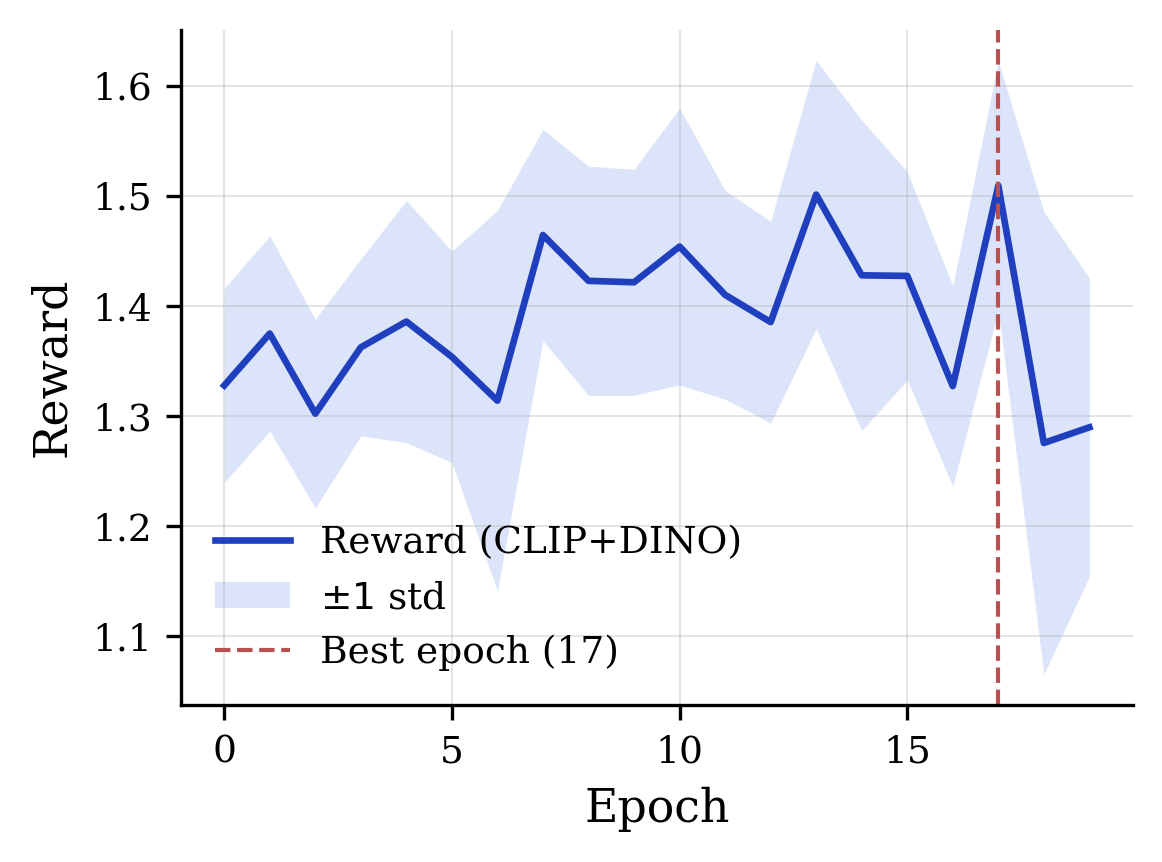}
        \caption{CLIP + DINO}\label{fig:rc_clip_dino}
    \end{subfigure}\hfill
    \begin{subfigure}[b]{0.245\textwidth}
        \includegraphics[width=\textwidth]{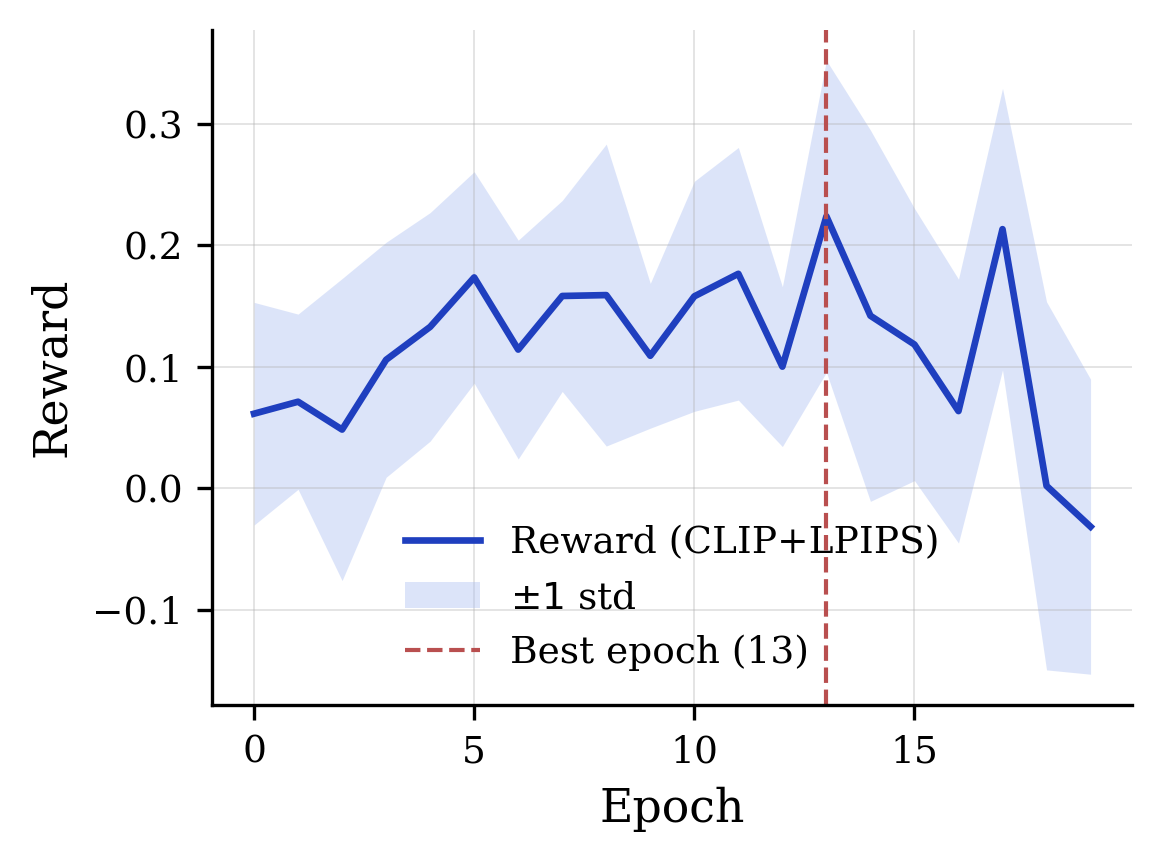}
        \caption{CLIP + LPIPS}\label{fig:rc_clip_lpips}
    \end{subfigure}\hfill
    \begin{subfigure}[b]{0.245\textwidth}
        \includegraphics[width=\textwidth]{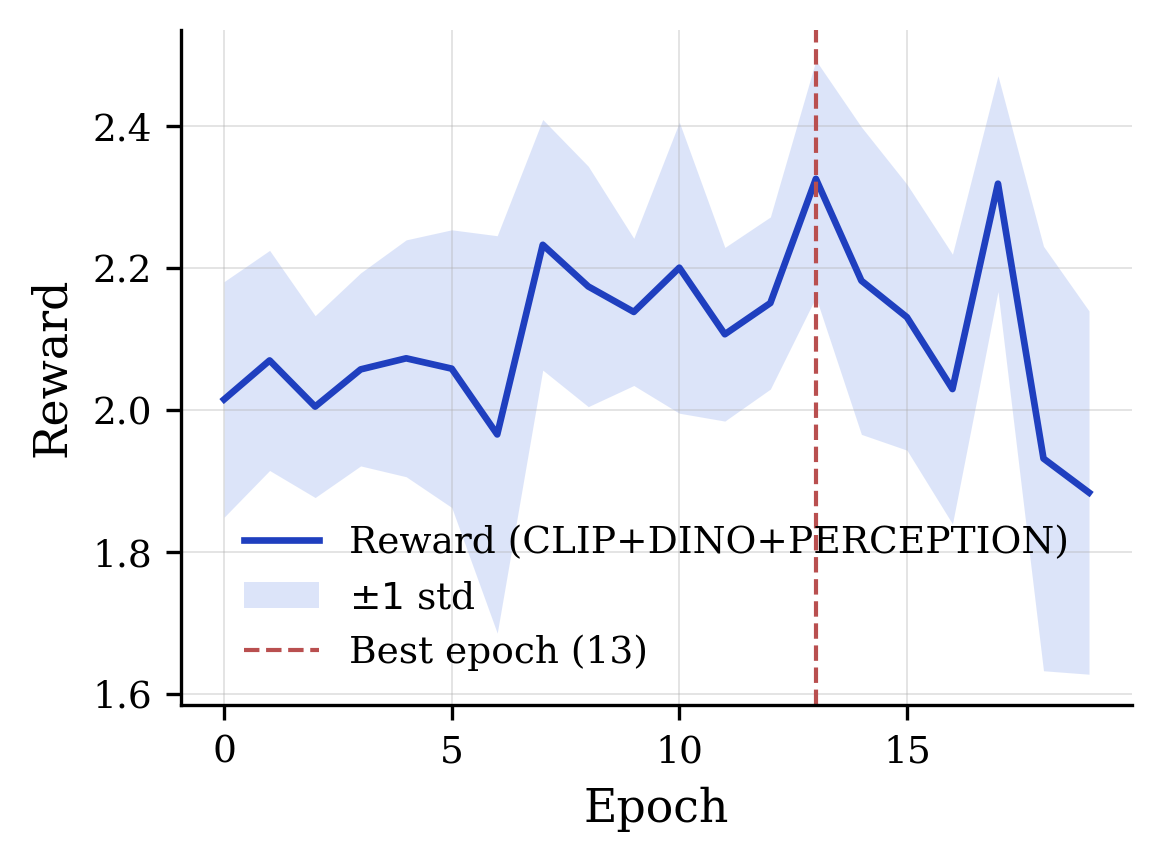}
        \caption{CLIP + DINO + PE}\label{fig:rc_cdp}
    \end{subfigure}\hfill
    \begin{subfigure}[b]{0.245\textwidth}
        \includegraphics[width=\textwidth]{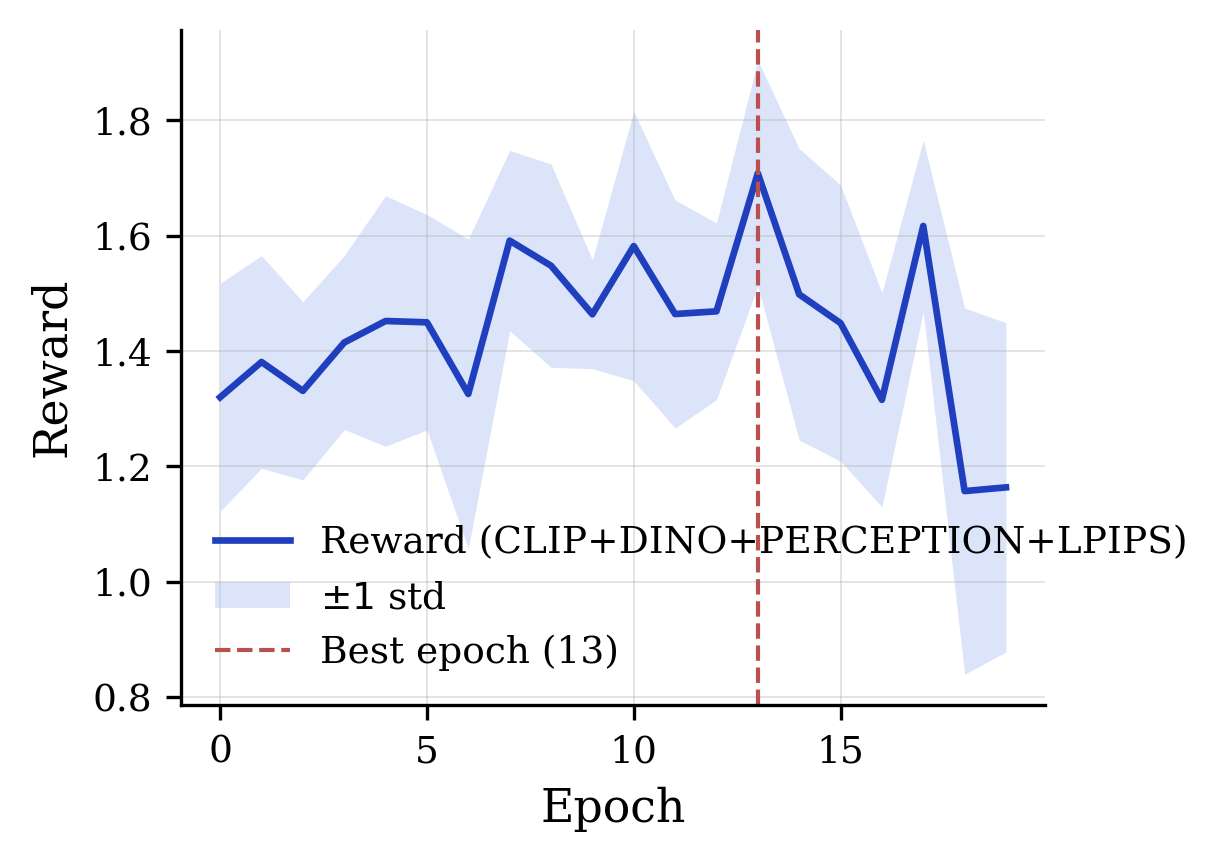}
        \caption{CLIP + DINO + PE + LPIPS}\label{fig:rc_cdpl}
    \end{subfigure}

    \vspace{0.6em}

    \begin{subfigure}[b]{0.245\textwidth}
        \includegraphics[width=\textwidth]{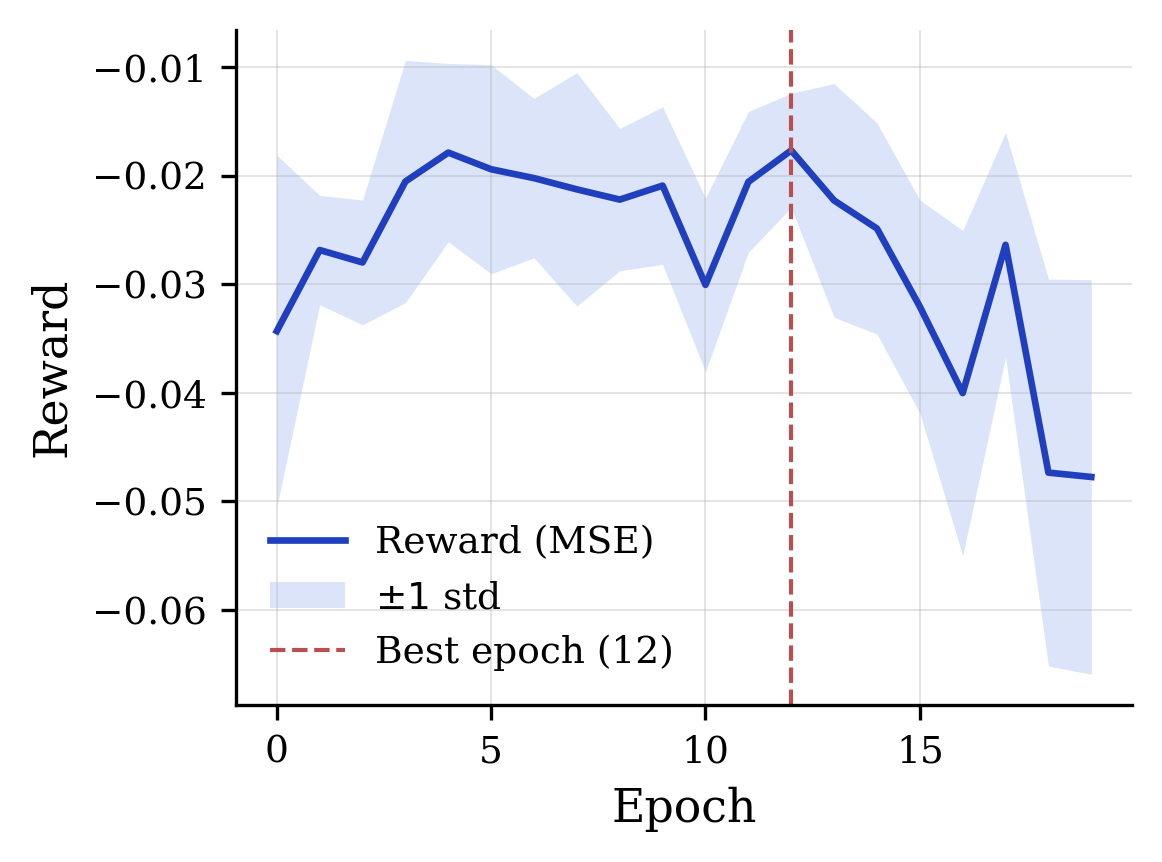}
        \caption{Negative MSE}\label{fig:rc_mse}
    \end{subfigure}\hfill
    \begin{subfigure}[b]{0.245\textwidth}
        \includegraphics[width=\textwidth]{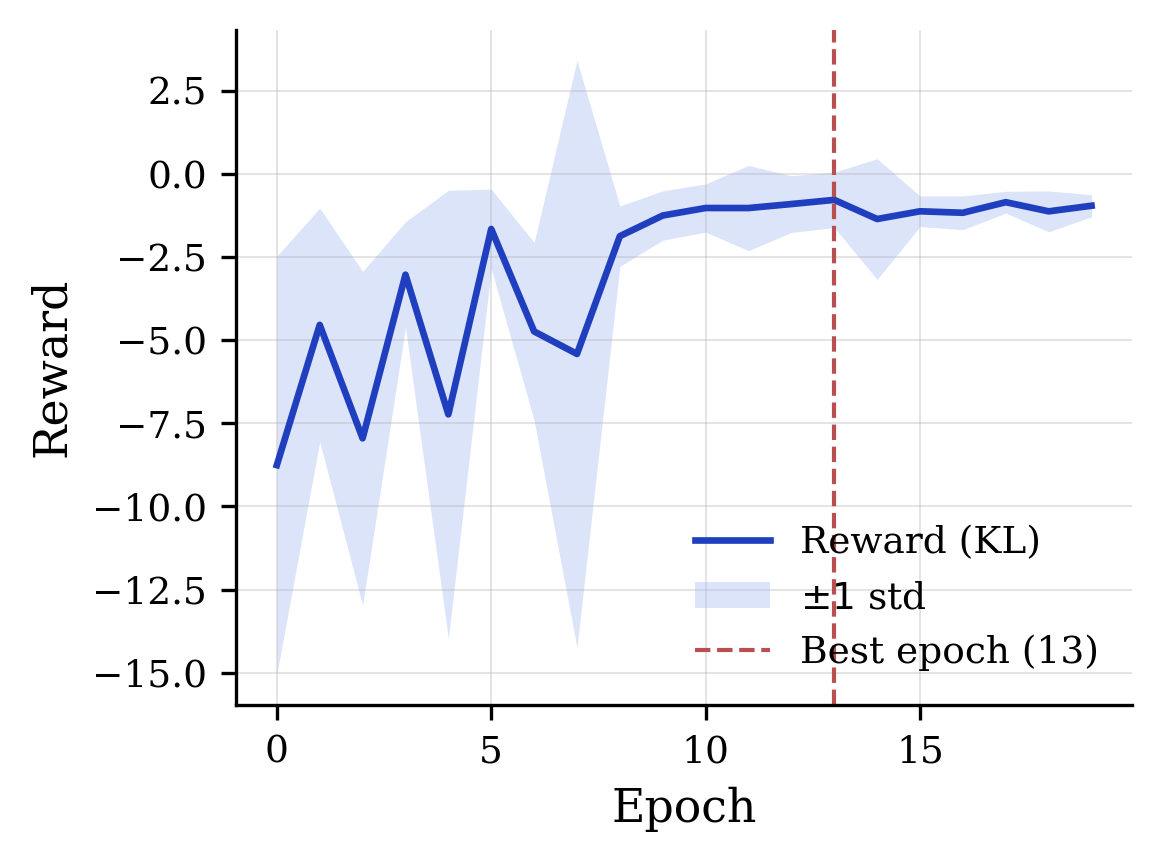}
        \caption{Negative KL}\label{fig:rc_kl}
    \end{subfigure}\hfill
    \begin{subfigure}[b]{0.245\textwidth}
        \phantom{\includegraphics[width=\textwidth]{images/reward_curves/KL.png}}
    \end{subfigure}\hfill
    \begin{subfigure}[b]{0.245\textwidth}
        \phantom{\includegraphics[width=\textwidth]{images/reward_curves/KL.png}}
    \end{subfigure}

    \caption{Training reward curves for ten of the fourteen reward configurations of
    Table~\ref{tab:reward_ablation_imagenet}, under identical PPO optimization, conditioning pool and
    seeds. All panels are from the Stable Diffusion v1.5 branch ($50\rightarrow5$ steps); the
    corresponding curves on ImageNet-64 differ, and the text below says how. These are $20$-epoch
    runs, shorter than the runs reported in Table~\ref{tab:reward_ablation_imagenet}, and are shown
    for the shape of the curves rather than for their end points. Solid lines show the mean reward per
    epoch, shaded regions $\pm1$ standard deviation across conditioning signals, and the dashed
    vertical line the reward-maximizing epoch. Panels (a)--(d) are single-encoder rewards, (e)--(h)
    combinations, and (i)--(j) the pixel and divergence controls.}
    \label{fig:reward_curves_all}
\end{figure*}

The tables summarize the end point of training. Figure~\ref{fig:reward_curves_all} reports the
training dynamics of every configuration under the same procedure, conditioning pool and seeds: the
mean reward per epoch, a $\pm1$ standard-deviation band across the conditioning signals of that epoch,
and the reward-maximizing epoch. Reward scales are not comparable across panels, since single-encoder
cosine rewards and their equally weighted combinations both lie in $[-1,1]$, while the MSE and KL
controls are unbounded below. The quantity of interest is the shape of each curve.

\paragraph{Observations.} Most of the improvement occurs within the first ten epochs for every
configuration, consistent with the fine-tuning view of Section~\ref{subsec:cost}. The per-epoch
reward is noisy, and the reward-maximizing epoch of every semantic configuration falls between $12$
and $17$ of these $20$ epochs. That is where the semantic rewards are optimized in the sense that
matters: past it the reward stays high or keeps rising while sample quality no longer improves, which
is over-optimization rather than further progress. Figure~\ref{fig:training_progress_oop} shows what
that looks like on the sample of Figure~\ref{fig:training_progress}, continued past the budget into
the same phase. None of the curves is monotone, and the behavior is analyzed in
Appendix~\ref{app:overopt}.

\paragraph{The pixel and divergence controls behave differently.} The MSE and KL panels are not the
semantic curves on a different scale. They rise slowly and are still rising at the end of the run,
and the divergence control fluctuates strongly over the first epochs before settling. Neither reaches
a comparable operating point within this budget: matching an image pixel by pixel, or matching a
distribution, is a much larger fitting problem than agreeing with a frozen encoder, and it grows with
the image, so on larger images and correspondingly larger search spaces these rewards converge slowly
and are poorly suited to the low-training regime. The compensation is on the other side: because they
constrain every direction rather than only the directions an encoder sees, their exploitable
directions are far fewer, and over-optimization arrives later than it does for a semantic reward.
Two properties follow from the same fact and should be kept in mind when reading these panels. They
are unbounded below, so their scale is not comparable with the cosine rewards, which sit in $[-1,1]$;
and Proposition~\ref{prop:semantic} does not apply to them at all, since it is a statement about the
null space of an encoder's Jacobian and these rewards have no encoder.

\paragraph{The size of the search space sets the timing.} The epoch at which over-optimization begins
is not a property of the reward alone. The smaller the space the policy searches, the later it
arrives: on ImageNet-64, where the images are $64\times64$ and the trajectory space is correspondingly
smaller than on the $512\times512$ Stable Diffusion branch shown here, the same rewards stay useful
for longer before the student starts finding directions the encoders cannot see. The numbers quoted
above are therefore specific to the branch these panels come from, and the $12$--$17$ epoch window
should be read as an instance of the procedure in Appendix~\ref{app:implementation}, which fixes the
training length from held-out FID on each branch, rather than as a constant of the method.

\section{Over-Optimization}
\label{app:overopt}

The semantic reward is, to first order, blind to every image change in the null space of its
encoders (Proposition~\ref{prop:semantic}). Once the student has matched the teacher in the directions
the encoders see, further optimization can therefore raise the reward, or keep it constant, while
changing the image in directions that image-level metrics do penalize. We observe this in two ways.
The reward curves of Figure~\ref{fig:reward_curves_all} keep rising or stay at their maximum late in
training, after samples have stopped improving visually. When a run is continued past the fixed
budget, samples degrade and acquire structured artifacts that the reward does not penalize
(Figure~\ref{fig:training_progress_oop}).

Two consequences follow for the protocol. First, the training reward is not a valid stopping signal,
which is why the training length is fixed once from held-out FID (Appendix~\ref{app:implementation}).
Second, the reward configuration matters: summing encoders shrinks the first-order invariant set to
the intersection of their null spaces, which is consistent with the four-encoder reward
R\textsubscript{2} being the best configuration on both branches (Appendix~\ref{app:reward_ablation}).
The KL regularizer of Section~\ref{subsec:ppo} bounds the drift of each update relative to the teacher
policy, but we have not isolated its effect, and the COCO runs use $\beta=0$. Preference rewards
(Appendix~\ref{app:preference}) are more exposed still, because their degenerate optimum lies along the
same direction as the intended effect.

\section{Non-Differentiable Preference Rewards}
\label{app:preference}

Section~\ref{subsec:preference} introduces a preference term into the distillation reward and summarizes the corresponding result. This appendix describes the two preference rewards and the evaluation protocol. The rewards were designed to encourage opposing behaviors: the compressibility reward favors removing high-frequency content, whereas the sharpness reward encourages concentrating it into narrow transitions. Our goal is to demonstrate that preference objectives can be incorporated into the distillation process without modifying the underlying procedure. We do not claim novelty in optimizing non-differentiable rewards using policy gradients, as DDPO~\citep{Black2023DDPO} already demonstrates this capability through JPEG compressibility optimization. Instead, we show that our framework is compatible with such objectives, enabling preferences to be incorporated into the distillation process.

\subsection{Reward Definitions}
\label{app:preference:rewards}

Let $I \in [0,1]^{3 \times H \times W}$ be a decoded sample, $g(I) \in [0,1]^{H \times W}$ its
luminance ($0.299r + 0.587g + 0.114b$), and $p$ a pixel. All quantities below are computed on the
decoded image.

\paragraph{Compressibility.}
Let $\mathrm{JPEG}_q(I)$ be the byte string produced by a baseline JPEG encoder at quality $q$, and
$|\cdot|$ its length in kilobytes:
\begin{equation}
  R_{\mathrm{jpeg}}(I) \;=\; -\,\frac{\bigl|\mathrm{JPEG}_q(I)\bigr|}{\kappa},
  \qquad q = 95,\quad \kappa = 100\,\mathrm{kB} .
  \label{eq:r-jpeg}
\end{equation}
The normalizer $\kappa$ only changes the value of $\gamma$ in \eqref{eq:r-combined} that gives a
desired trade-off. The encoder is a discrete pipeline (block DCT, integer quantization, run-length and
Huffman coding) whose output is a byte count, so $R_{\mathrm{jpeg}}$ is piecewise constant in its
input and its gradient is zero wherever it exists. Differentiable approximations of JPEG exist, but
they optimize a surrogate rather than the encoded size itself.

\paragraph{Sharpness.}
An edge is sharp when its transition completes in few pixels, independently of how much contrast it
carries. Let $d(p) = \max_{p' \in \mathcal{N}_4(p)} |g(p) - g(p')|$ be the largest luminance step to a
$4$-neighbor and $r(p)$ the range of $g$ over the $5\times5$ window at $p$, and define
\begin{equation}
  \mathcal{E} = \bigl\{\, p : d(p) \ge 0.08 \ \wedge\ r(p) \ge 0.15 \,\bigr\},
  \qquad
  \mathcal{E}_{\mathrm{sharp}} = \bigl\{\, p \in \mathcal{E} : d(p) \ge \tau\, r(p) \,\bigr\},
  \label{eq:edge-sets}
\end{equation}
with $\tau = 0.5$, so that $C(I) = |\mathcal{E}_{\mathrm{sharp}}| / \max(|\mathcal{E}|,1)$ is the
fraction of edge pixels whose transition completes within about $1/\tau = 2$ pixels. Because the test
$d(p)\ge\tau r(p)$ compares a step with the local range, it measures edge \emph{width} and is invariant
to rescaling the contrast of an edge that is already in $\mathcal{E}$; contrast still matters through
the absolute thresholds that define $\mathcal{E}$. With the edge-presence factor
$\omega(I) = \min\bigl(1,\, |\mathcal{E}| / (0.01\,HW)\bigr)$,
\begin{equation}
  R_{\mathrm{sharp}}(I) \;=\; \omega(I)\,C(I) \;\in\; [0,1] .
  \label{eq:r-sharp}
\end{equation}
The factor $\omega$ stops an image with almost no edges from scoring highly on the strength of a
handful of crisp ones. It does not remove the more general weakness of a ratio: $C$ is also raised by
suppressing soft edges rather than by sharpening them, and $\omega$ catches that only once fewer than
$1\%$ of pixels remain in $\mathcal{E}$. Appendix~\ref{app:preference:results} returns to this. Both
$C$ and $\omega$ are counts of pixels passing fixed thresholds; the thresholds were set by hand and
not tuned against the outcome.

\paragraph{Why these are not differentiable.}
Each reward is a composition of indicator functions, integer counts, set cardinalities or an entropy
coder. Each is constant on an open neighborhood of almost every input, so
$\partial R / \partial I = 0$ wherever it is defined. A pathwise method such as
DRaFT~\citep{clark2024draft} receives an exactly zero gradient, and a supervised distillation loss
cannot use them, since none defines a per-step target. The score-function estimator of
Proposition~\ref{prop:credit} is unaffected: it weights $\nabla_\theta \log \pi_\theta$ by the scalar
$R - b$ and never differentiates $R$.

\subsection{Protocol}
\label{app:preference:protocol}

We train the $8$-step COCO student exactly as in Section~\ref{sec:experiments}, with the distillation
reward R\textsubscript{1} and one preference term added as in \eqref{eq:r-combined}.
All other settings are those of the COCO column of Table~\ref{tab:implementation}: LoRA rank~4 on the
UNet attention projections~\citep{hu2021lora}, learning rate $3\times10^{-4}$, clip range $10^{-4}$,
effective batch $8$, guidance $5.0$, $\beta = 0$, and $20$ epochs; the preference weight $\gamma$ is
listed in the same table.
Evaluation images are generated at every epoch from a fixed prompt set, each prompt with its own fixed
seed, with the sampler settings held constant, so the only quantity that differs between two columns
of Figure~\ref{fig:preference-progression} is the policy's weights, as in
Appendix~\ref{app:qualitative}.

\subsection{Results}
\label{app:preference:results}

\begin{figure*}[t]
    \centering
    \setlength{\tabcolsep}{2pt}
    \renewcommand{\arraystretch}{1.2}
    \begin{tabular}{c|cccccc}
        & \textbf{Init} & \textbf{Epoch 5} & \textbf{Epoch 10} & \textbf{Epoch 13} & \textbf{Epoch 15} & \textbf{Epoch 18} \\
        \midrule
        \rotatebox[origin=c]{90}{\textbf{Compressibility}} &
        \includegraphics[width=0.145\textwidth]{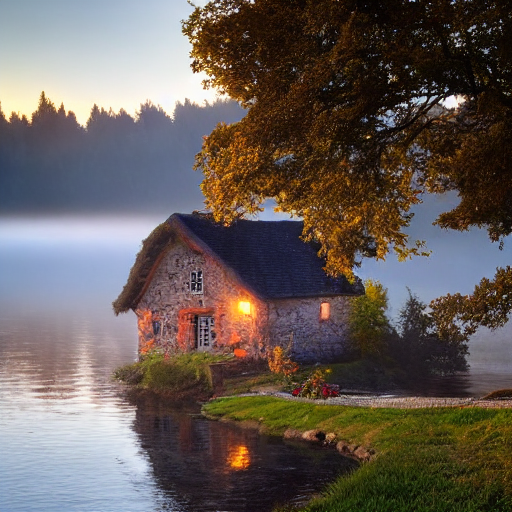} &
        \includegraphics[width=0.145\textwidth]{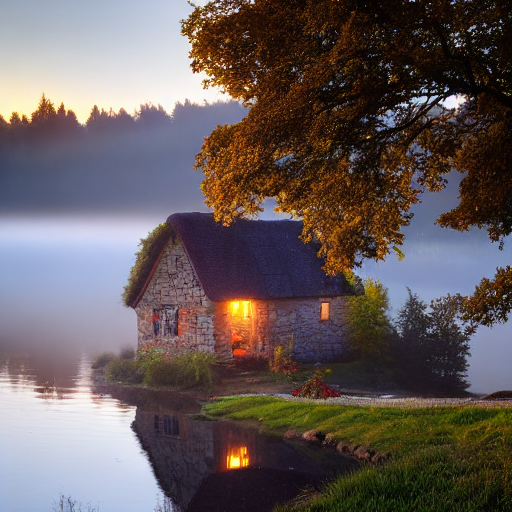} &
        \includegraphics[width=0.145\textwidth]{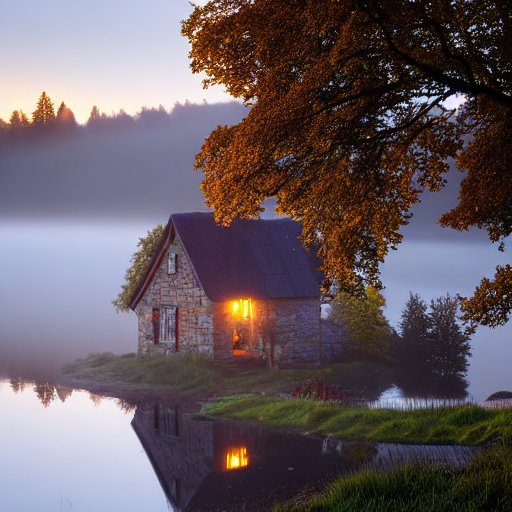} &
        \includegraphics[width=0.145\textwidth]{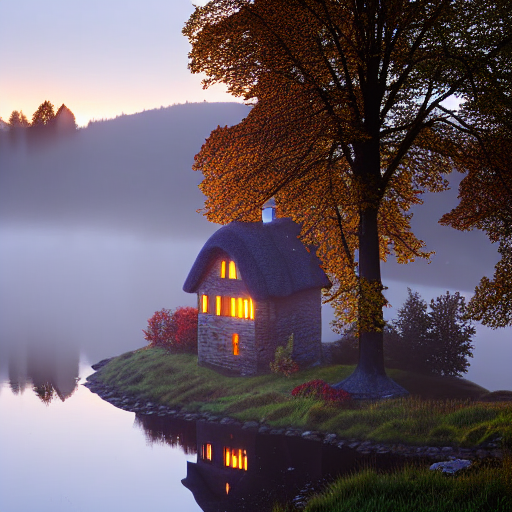} &
        \includegraphics[width=0.145\textwidth]{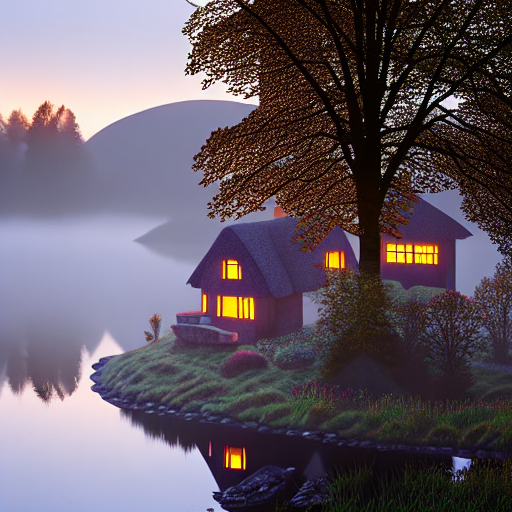} &
        \includegraphics[width=0.145\textwidth]{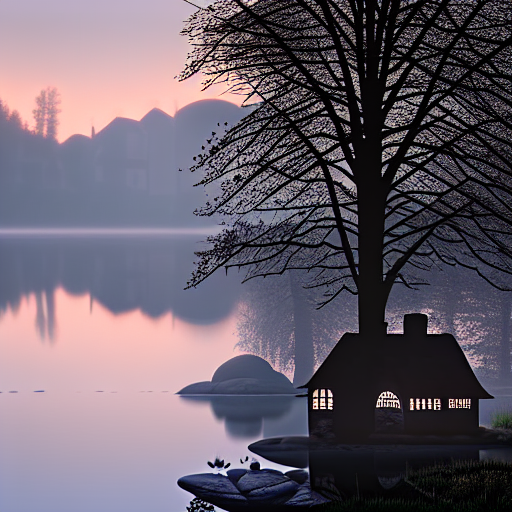} \\
        \rotatebox[origin=c]{90}{\textbf{Sharpness}} &
        \includegraphics[width=0.145\textwidth]{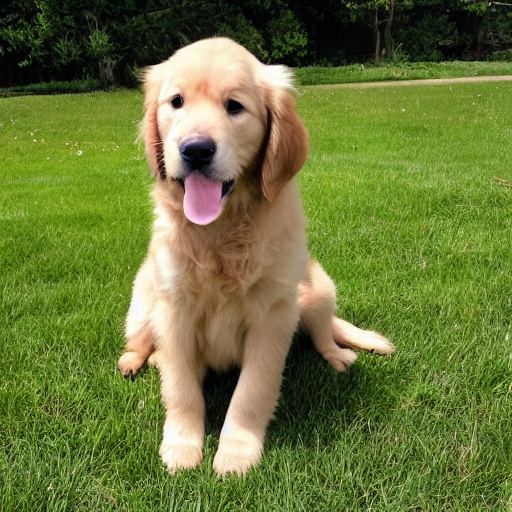} &
        \includegraphics[width=0.145\textwidth]{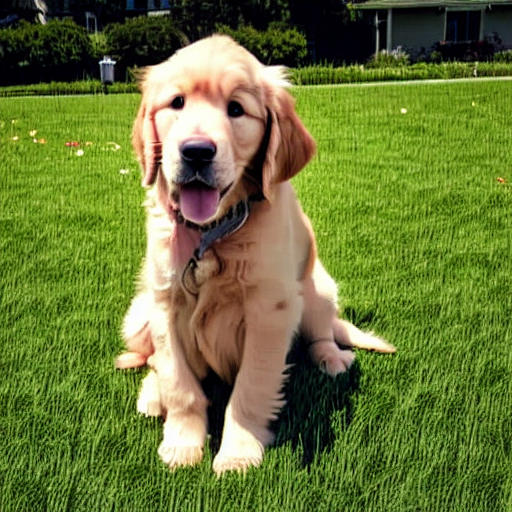} &
        \includegraphics[width=0.145\textwidth]{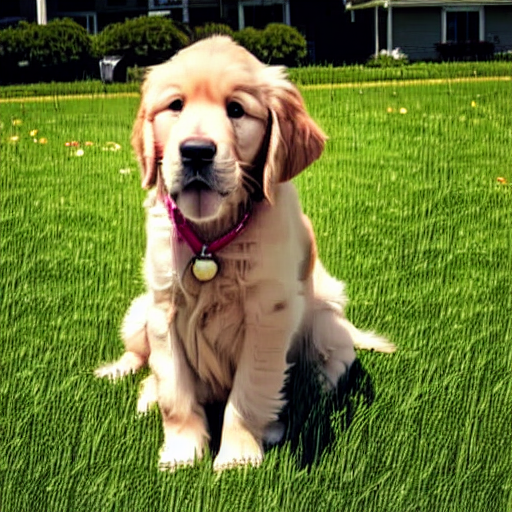} &
        \includegraphics[width=0.145\textwidth]{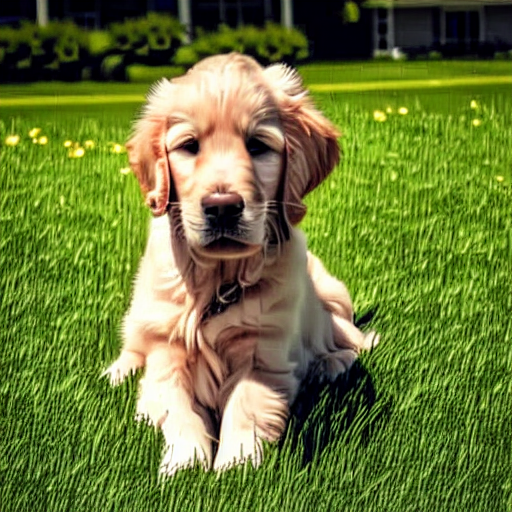} &
        \includegraphics[width=0.145\textwidth]{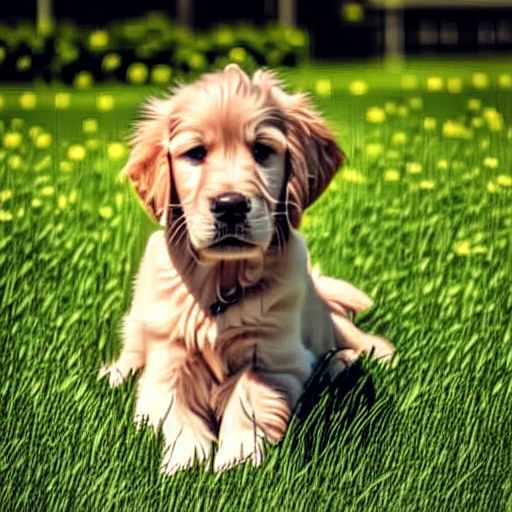} &
        \includegraphics[width=0.145\textwidth]{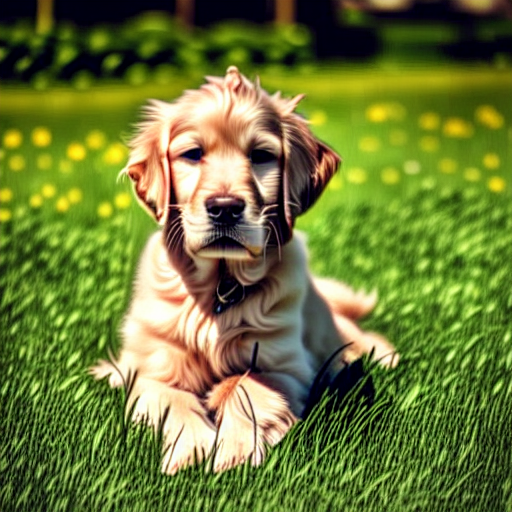} \\
    \end{tabular}
    \caption{The COCO student trained with the distillation reward plus one non-differentiable preference, for two fixed prompts and seeds: compressibility $R_{\mathrm{jpeg}}$ \eqref{eq:r-jpeg} above, sharpness $R_{\mathrm{sharp}}$ \eqref{eq:r-sharp} below.
    \emph{Init} is the student before any update. Only the policy's weights differ between columns.}
    \label{fig:preference-progression}
\end{figure*}

Figure~\ref{fig:preference-progression} follows two fixed prompts and one seed through training under each preference. The leftmost column is the student before any update. Under $R_{\mathrm{jpeg}}$ the sample loses high-frequency detail and moves toward large regions of near-constant and dark color, the configuration that minimizes encoded size. Under $R_{\mathrm{sharp}}$ the transitions tighten and fine detail is retained, the opposite direction in the same image statistic. In both cases the preference was added by adding one term to the reward: no other component of the procedure was modified.

The evidence is qualitative. A single prompt and seed cannot distinguish a systematic effect from a favorable example, and without image-quality metrics the figure does not show how much fidelity each
preference costs; quantifying that trade-off requires the reward, FID and CLIPScore on held-out prompts across $\gamma$. The degenerate optimum of each preference also lies along the same direction as the
intended effect: a flat color field for $R_{\mathrm{jpeg}}$, and for $R_{\mathrm{sharp}}$ either haloing and grain or, as noted above, the suppression of soft edges, which raises the ratio $C$ without sharpening anything. A preference metric should therefore never be reported on its own, and the training length should be fixed from held-out quality rather than from the reward, as in Appendix~\ref{app:implementation}.

\section{LoRA and the Clip Range}
\label{app:lora}

PPO keeps each update small through the clip range $\epsilon$ on the likelihood ratio, and the KL term
of Section~\ref{subsec:ppo} keeps the policy near the teacher. A LoRA student adds a third restriction
of a different kind: the adapter starts at zero and confines every update to a rank-$r$ subspace of the
attention weights, so the policy can move only along few directions. We test whether this matters for
optimization by crossing the parameterization with the clip range on the EDM/ImageNet-64 branch.

\paragraph{Setup.} All runs use the R\textsubscript{2} reward, $M=8$, effective batch $32$, four inner
epochs, $\beta=0.02$ and the $100$-epoch budget of Table~\ref{tab:matched}, and differ only in the
student parameterization and in $\epsilon\in\{10^{-3},10^{-4}\}$. The LoRA student is the rank-4 adapter
of Table~\ref{tab:implementation} at learning rate $3\times10^{-4}$; the full-parameter student trains
every weight of the UNet at $1\times10^{-5}$, a rate thirty times smaller, so the full-parameter rows
are not disadvantaged by a larger step. Each configuration is run with two seeds and evaluated at the
last checkpoint under the protocol of Appendix~\ref{app:protocol}. The LoRA, $\epsilon=10^{-4}$ row is
the ReDiF (R\textsubscript{2}) configuration of Table~\ref{tab:matched}, run separately.

\begin{table}[t]
\centering
\caption{Student parameterization crossed with the PPO clip range, EDM teacher, ImageNet-64
($256\rightarrow8$), R\textsubscript{2} reward, $100$ epochs. Mean $\pm$ sample sd over two seeds, last
checkpoint. The full-parameter student uses a learning rate thirty times smaller than the LoRA
student. Best in bold, second best bold and dash-underlined.}
\label{tab:lora_clip}
\resizebox{\linewidth}{!}{
\begin{tabular}{lccccccc}
\toprule
\textbf{Student} & $\boldsymbol{\epsilon}$ & \textbf{LR} & \textbf{P} $\uparrow$ & \textbf{R} $\uparrow$ &
\textbf{D} $\uparrow$ & \textbf{C} $\uparrow$ & \textbf{FID} $\downarrow$ \\
\midrule
LoRA ($r=4$) & $10^{-3}$ & $3\times10^{-4}$ & $0.7257 \pm 0.0013$ & \second{0.6003 \pm 0.0075} & $0.5751 \pm 0.0102$ & $0.7857 \pm 0.0010$ & \second{20.01 \pm 0.46} \\
LoRA ($r=4$) & $10^{-4}$ & $3\times10^{-4}$ & $0.7391 \pm 0.0016$ & \best{0.6145 \pm 0.0171} & \best{0.7829 \pm 0.0084} & \second{0.8043 \pm 0.0001} & \best{18.96 \pm 1.66} \\
\hdashline
Full         & $10^{-3}$ & $1\times10^{-5}$ & \best{0.7574 \pm 0.0535} & $0.4652 \pm 0.1089$ & $0.6131 \pm 0.1521$ & $0.6229 \pm 0.2581$ & $29.78 \pm 0.85$ \\
Full         & $10^{-4}$ & $1\times10^{-5}$ & \second{0.7572 \pm 0.0147} & $0.5677 \pm 0.0250$ & \second{0.6624 \pm 0.0137} & \best{0.8129 \pm 0.0199} & $22.22 \pm 4.81$ \\
\bottomrule
\end{tabular}
}
\end{table}

\paragraph{Findings.} Table~\ref{tab:lora_clip} reports the four configurations. Under LoRA the clip
range leaves the operating point intact: FID moves by $1.05$ across the two settings ($20.01$ against
$18.96$), inside the seed spread of either row, and Recall is unchanged within the same spread. The
diversity metrics do move, Density from $0.5751$ to $0.7829$ and Coverage from $0.7857$ to $0.8043$,
so the adapter does not make the clip range irrelevant; both settings remain usable, and neither
collapses. The full-parameter student behaves differently: it depends strongly on $\epsilon$. At $\epsilon=10^{-3}$
it is the worst configuration in FID ($29.78$), and one of its two seeds loses most of its diversity
(Recall $0.39$, Coverage $0.44$, against $0.54$ and $0.81$ for the other), which is the source of the
large spread in that row. Tightening the clip to $10^{-4}$ restores diversity and brings the mean FID
to $22.22$, but the two seeds still differ by almost seven FID points ($18.81$ and $25.62$). The best
single full-parameter run thus matches the LoRA runs, but the full-parameter student reaches that
point only with the tighter clip and then unreliably, while the LoRA student reaches it at either
setting. The full-parameter student does keep a higher Precision at $\epsilon=10^{-4}$ ($0.7572$ against
$0.7391$), so its advantage, where it has one, is in fidelity rather than in diversity.

We read this as the adapter supplying part of the stabilization that the clip range otherwise has to
provide alone: with LoRA neither setting of $\epsilon$ produces a degenerate run, and without it one
setting does. We
do not measure the per-update policy change directly, and with two seeds per configuration the
comparison of means is indicative; the contrast in sensitivity, between rows that agree across
settings and seeds and rows that do not, is the more robust observation. The ablation concerns ReDiF
only and says nothing about whether LoRA helps or hurts the retrained baselines of
Table~\ref{tab:matched}.

\section{Effect of the Student Step Count}
\label{app:step_count}

Section~\ref{subsec:setup} states that the student step count $M$ is bounded below by the need for a horizon over which the policy can explore, and above by the growth of estimator variance and per-iteration cost with the horizon. This appendix reports the measurements behind that choice on the EDM/ImageNet-64 branch, where the absence of guidance removes the confound of
Section~\ref{subsec:cfg}.

\begin{figure}[t]
    \centering
    \includegraphics[width=0.49\linewidth]{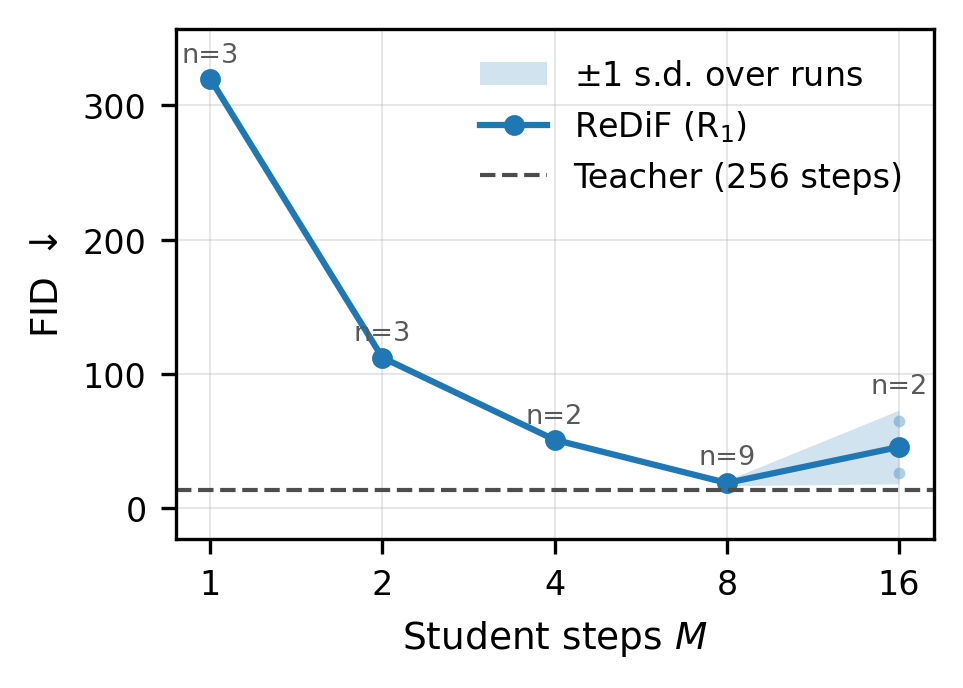}\hfill
    \includegraphics[width=0.49\linewidth]{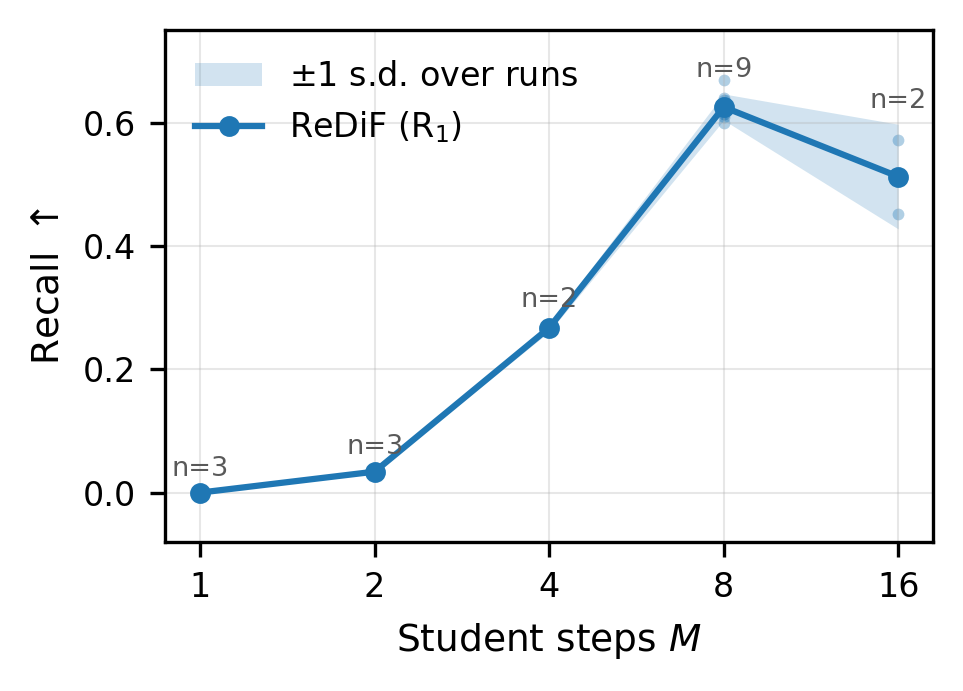}\\[2pt]
    \includegraphics[width=0.49\linewidth]{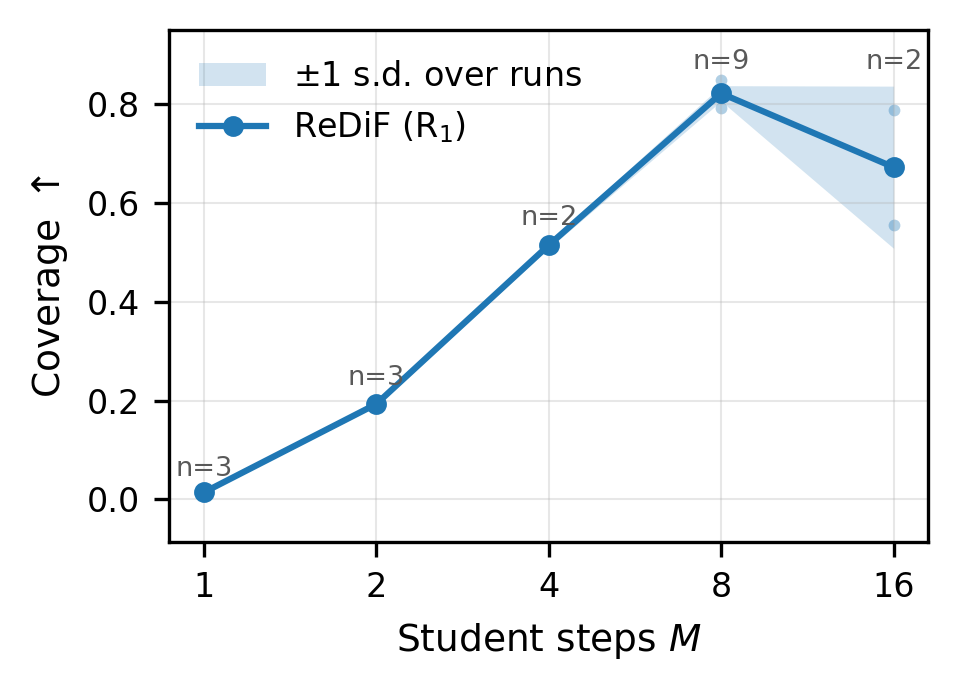}\hfill
    \includegraphics[width=0.49\linewidth]{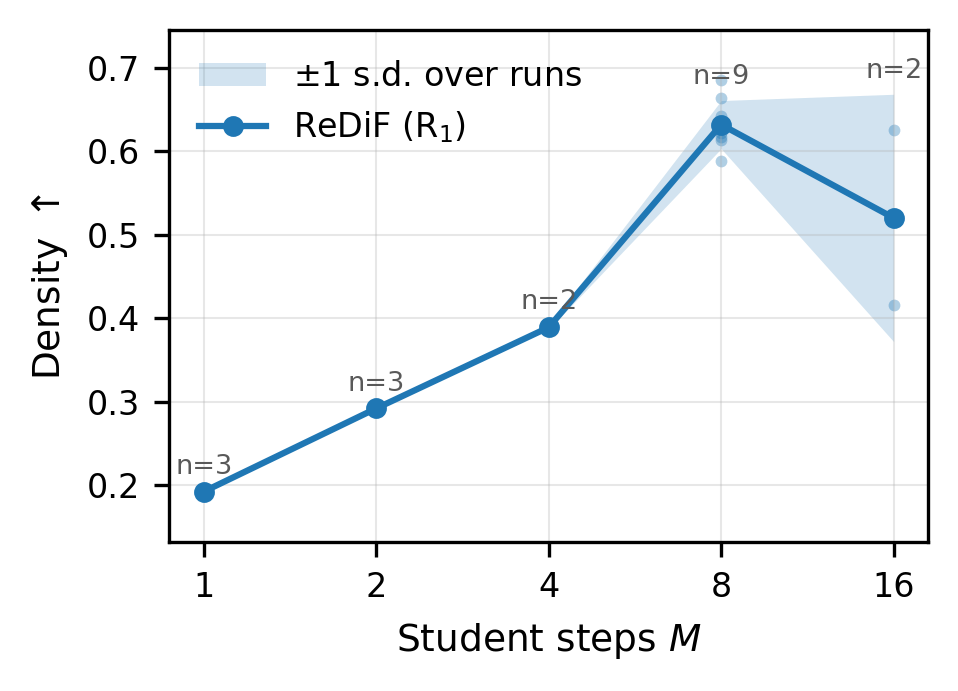}
    \caption{Effect of the student step count $M$ on the EDM/ImageNet-64 branch under a fixed training
    budget. Top: FID (left) and Recall (right). Bottom: Coverage (left) and Density (right). Dashed
    lines mark the $256$-step teacher and the dotted line the operating point $M=8$. $n$ is the number
    of runs at each $M$ and the band is $\pm1$ s.d.\ over runs.}
    \label{fig:step_count}
\end{figure}

We repeat the matched-budget run of Section~\ref{subsec:matched} for $M\in\{1,2,4,8,16\}$, holding the reward (R\textsubscript{1}), conditioning pool, LoRA configuration, batch and epoch budget fixed, so every step count receives the same number of policy updates. Figure~\ref{fig:step_count} reports FID, Recall, Coverage and Density as functions of $M$ against the $256$-step teacher. Each point is the mean over independent runs, between two and nine per step count as annotated, and the band is one standard deviation. The $M=8$ point corresponds to the ReDiF (R\textsubscript{1}) row of Table~\ref{tab:matched}.

Three observations follow. First, very short horizons fail, and they fail most visibly in diversity.
At $M\le2$ Recall and Coverage are close to zero, and at $M=4$ they remain well below their best
values even though FID has already recovered most of the way from its $M=1$ value. The reason is
specific to how the method works. ReDiF improves a student by exploring: the policy perturbs its own
transitions, keeps the perturbations that score well and pushes probability mass toward them. A
one- or two-step student has almost nothing to explore over. Its entire trajectory is one or two
decisions, each of which fixes a large part of the image at once, so the policy has very little
freedom: the perturbations it can make are either too small to change the sample or large enough to
destroy it, and there are no later transitions left to correct an early move. What limits quality
there is not how well the policy explores but how little there is to explore over. The task also
changes character: a one- or two-step student has to memorize a direct map from noise to image rather
than learn a policy that generalizes across trajectories, and diversity and generalization matter
less to it than raw fitting capacity. Fitting that map is bound by iterations and compute, which is
what the one-step literature spends thousands of GPU-hours on, so dense per-step targets have more to
offer there than a terminal reward does and the regime is not a low-training one in the first place.
This is also why we treat Recall and Coverage as primary metrics: the collapse at $M\le2$ shows up in
diversity long before it shows up in FID.

Second, $M=8$ is the best step count on all four metrics. At this point the student closes most of the gap between the $8$-step teacher (FID $35.04$) and the $256$-step teacher ($13.74$), reaching $18.84$. Its Coverage exceeds the teacher's ($0.8223$ against $0.7831$), while Recall ($0.6265$ against $0.7103$) and Density ($0.6318$ against $0.8930$) remain below it.

Third, $M=16$ regresses relative to $M=8$ on every metric, and the spread across runs grows sharply.
The failure at the long end is the opposite of the one at the short end. What the policy searches
over is trajectories, and a trajectory is a choice at each of $M$ transitions: the policy factorizes
as $\prod_{t}\pi_\theta(a_t\mid s_t)$, so the space it must cover is a product over transitions and
compounds with the horizon rather than growing in proportion to it. Coarsely, if one counts
trajectories that differ at a fixed resolution on the state space, doubling the horizon squares that
count; for continuous actions this is a counting heuristic rather than a statement about the measure,
and the exact version is only that the trajectory distribution lives on a space of $M$ times the
per-state dimension. Either way the supervision does not grow with it: a single terminal scalar is
still all the estimator has, and it is now spread over twice as many score terms
(Proposition~\ref{prop:credit}). Reaching a good policy over the larger space therefore takes more
updates, and at a fixed batch that means more epochs, which is exactly what the low-training regime
does not allow. With the epoch budget held at the value used everywhere else, the $M=16$ runs have
not converged: which part of the space each run settles in still varies between seeds, and that is
what the widening band reports. With two runs at $M=16$ we do not separate these effects
quantitatively. We read the result as showing that, at this budget, the longer horizon does not pay
for itself, not as evidence that longer students are worse at convergence; $M=8$ is where the horizon
is long enough to explore and short enough to cover within the budget.

\section{Published Results at Convergence}
\label{app:published}

Section~\ref{subsec:matched} compares algorithms in the low-training regime, which invites the
objection that the baselines did not receive the compute their authors gave them. Table~\ref{tab:published}
represents each method by the best result its own authors report, quoted without re-implementation,
together with the effective batch and hardware they disclose. The point is position on the
quality--cost plane rather than state-of-the-art FID, so the cost columns are first-class. These
methods are far ahead of where we retrained them, and the distance is the budget.

We report cost as disclosed and mark \emph{n/r} where a paper does not state it, with one exception:
for progressive distillation, consistency distillation and CTM on ImageNet-64 we quote the
third-party A100-hour estimates of~\citet{Zhou2024SFD}, marked $\dagger$. Batch is listed separately
from accelerator-hours because the two are not interchangeable at fixed hardware.

\begin{table}[t]
\centering
\caption{\textbf{Published results at convergence.} FID as reported by the original authors, except
where marked: ImageNet-64 with $50$k balanced samples, COCO-2014 zero-shot with $30$k samples. Batch
is the effective training batch and Devices the reported hardware, both as disclosed; $\dagger$ marks
a third-party A100-hour estimate from~\citet{Zhou2024SFD}; $\ddagger$ marks an FID we quote
from~\citet{song2023consistency}, who evaluate progressive and consistency distillation on this
teacher, rather than from the method's own paper; $\ast$ marks the batch of the default ablation
setting reported by~\citet{sauer2023adversarialdiffusiondistillation}, which the paper does not state
to be the setting of its final run; \emph{n/r} not reported; -- no number under this protocol. ReDiF has not been evaluated under these protocols, so its FID is omitted; its numbers under
the single-device protocol are in Table~\ref{tab:matched}. Results published only under other COCO
protocols are excluded rather than converted. Best published FID in bold, second best bold
and dash-underlined; teacher rows are not ranked.}
\label{tab:published}
\resizebox{\linewidth}{!}{
\begin{tabular}{llcccccc}
\toprule
\textbf{Method} & \textbf{Family} & \textbf{Steps} & \textbf{Aux.\ net} & \textbf{Trainable} & \textbf{Batch} & \textbf{Devices / training cost (reported)} & \textbf{FID} $\downarrow$ \\
\midrule
\multicolumn{8}{l}{\emph{(a) ImageNet-64}} \\
EDM teacher~\citep{karras2022elucidating} & -- & 256 & -- & -- & -- & -- & 1.36 \\
\hdashline
Progressive Distill.~\citep{salimans2022progressive} & trajectory & $N\!\rightarrow\!1$ & interm.\ teachers & full & 2048 & $\sim$64 GPUs; $\sim$5{,}533 A100-h$^\dagger$ & 15.39$^\ddagger$ \\
Consistency Distill.~\citep{song2023consistency} & trajectory & $N\!\rightarrow\!1$ & target copy & full & 2048 & $\sim$64 GPUs; $\sim$7{,}867 A100-h$^\dagger$ & 6.20$^\ddagger$ \\
CTM~\citep{kim2024ctm} & trajectory & $N\!\rightarrow\!1$ & target copy + disc. & full & 2048 & $\sim$64 GPUs; $\sim$902 A100-h$^\dagger$ & 1.92 \\
DMD~\citep{yin2024DMD} & distribution & $N\!\rightarrow\!1$ & score net & full & n/r & n/r & 2.62 \\
DMD2~\citep{yin2024IDMD} & distribution & $N\!\rightarrow\!1$ & score net + disc. & full & 280 & 7$\times$A100, $\sim$2 d ($\sim$336 A100-h) & \second{1.28} \\
SiD~\citep{zhou2024sid} & distribution & $N\!\rightarrow\!1$ & score net & full & 8192 & 16$\times$A100-40GB (accum.\ 32) & 1.524 \\
SiDA~\citep{zhou2024sida} & distribution & $N\!\rightarrow\!1$ & score net + disc. & full & 8192 & 8$\times$H100 (accum.\ 32) & \best{1.110} \\
TFD~\citep{zhang2026tfd} & drifting & $N\!\rightarrow\!1$ & none & full & n/r & 200k iters & 1.58 \\
\hdashline
ReDiF (ours) & RL & $256\!\rightarrow\!8$ & none & LoRA & 32 & 1$\times$A100-80GB, 3.7--3.8 h & -- \\
\midrule
\multicolumn{8}{l}{\emph{(b) COCO, Stable Diffusion v1.5 teacher}} \\
SD v1.5 teacher~\citep{rombach2022ldm} & -- & 50 & -- & -- & -- & -- & 8.74 \\
\hdashline
Guided Distill.~\citep{meng2023distillation} & trajectory & $50\!\rightarrow\!4$ & interm.\ teachers & full & n/r & n/r & -- \\
LCM~\citep{luo2023lcm} & trajectory & $50\!\rightarrow\!4$ & target copy & full & 72 & $\sim$32 A100-h & -- \\
InstaFlow~\citep{liu2024instaflow} & rectified flow & $50\!\rightarrow\!1$ & none & full & 1024 & 199 A100-days ($\sim$4{,}780 A100-h) & 13.10 \\
ADD / SD-Turbo~\citep{sauer2023adversarialdiffusiondistillation} & adversarial & $50\!\rightarrow\!1$ & disc.\ (+ real data) & full & 128$^{\ast}$ & n/r & -- \\
DMD~\citep{yin2024DMD} & distribution & $50\!\rightarrow\!1$ & score net & full & n/r & 72$\times$A100, $\sim$36 h ($\sim$2{,}600 A100-h) & 11.49 \\
DMD2~\citep{yin2024IDMD} & distribution & $50\!\rightarrow\!1$ & score net + disc. & full & 128 & 64$\times$A100, $\sim$60 h ($\sim$3{,}840 A100-h) & \second{8.35} \\
Guided SiD (GDD)~\citep{zhouguided} & distribution & $50\!\rightarrow\!1$ & score net & full & n/r & 20M fake images seen & \best{8.15} \\
\hdashline
ReDiF (ours) & RL & $50\!\rightarrow\!5$ & none & LoRA & 8 & 1$\times$A100-80GB, 0.54--0.58 h & -- \\
\bottomrule
\end{tabular}
}
\end{table}

These methods are not subsumed by our results and we do not claim otherwise. GDD~\citep{zhouguided}
reaches the strongest COCO FID without real images but retains an auxiliary score network;
TFD~\citep{zhang2026tfd} removes that network by working in the teacher's hidden states but estimates
its drifting field against real data where available; DMD2 and the adversarial variants require real
images outright. All keep a full-parameter student and a budget far above ours, and our position is
the LoRA-scale, single-device corner of the same design space. Taken with Table~\ref{tab:matched}, the
ordering at convergence is substantially a function of the budget.

\section{Baseline Learning Rates}
\label{app:baseline_lr}

Every baseline retrained in Section~\ref{subsec:matched} is run at a single learning rate, chosen per
method rather than shared, because these objectives do not tolerate the same step size: losses that
regress onto a fixed teacher target accept a larger rate than those whose target moves with an
auxiliary network. We take each value from the method's own paper where it reports a setting for this
teacher, and otherwise from the work that reports the strongest published result for that method. No
value was selected by us. ReDiF's own hyperparameters are likewise adopted rather than searched: they
follow the Stable Diffusion configuration of DDPO~\citep{Black2023DDPO}, which optimizes the same
clipped surrogate on the same backbone. Table~\ref{tab:baseline_lr} records the rate used for every
method.

\begin{table}[t]
\centering
\caption{Learning rate used for each retrained baseline in Table~\ref{tab:matched}. Values are taken
from each method's own paper where it reports a setting for this teacher, and otherwise from the work
reporting its strongest published result; none was selected by us. Objectives that regress onto a
fixed teacher target take a larger rate than those whose target moves with an auxiliary network.
Direct reward backpropagation on COCO uses the ReDiF rate because it optimizes the same reward.
``Aux.'' is the rate of the fake-score network where one exists. All students are LoRA adapters
trained with AdamW.}
\label{tab:baseline_lr}
\resizebox{\linewidth}{!}{
\begin{tabular}{llccl}
\toprule
\textbf{Branch} & \textbf{Method} & \textbf{LR} & \textbf{Aux.\ LR} & \textbf{Target type} \\
\midrule
ImageNet-64 & Progressive Distillation & $3\times10^{-4}$ & --               & fixed teacher target \\
            & Consistency Distillation & $3\times10^{-4}$ & --               & fixed teacher target \\
            & DMD                      & $1\times10^{-4}$ & $3\times10^{-4}$ & moving target (aux.\ network) \\
            & DMD2                     & $1\times10^{-4}$ & $3\times10^{-4}$ & moving target (aux.\ network) \\
            & SiD                      & $1\times10^{-4}$ & $3\times10^{-4}$ & moving target (aux.\ network) \\
            & SiDA                     & $1\times10^{-4}$ & $3\times10^{-4}$ & moving target (aux.\ network) \\
            & Direct reward backprop.  & $1\times10^{-4}$ & --               & reward gradient \\
\midrule
COCO        & Progressive Distillation & $1\times10^{-4}$ & --               & fixed teacher target \\
            & Consistency Distillation & $1\times10^{-4}$ & --               & EMA target \\
            & DMD                      & $5\times10^{-5}$ & $1\times10^{-4}$ & moving target (aux.\ network) \\
            & DMD2  & $5\times10^{-5}$ & $1\times10^{-4}$ & moving target (aux.\ network) \\
            & Direct reward backprop.  & $3\times10^{-4}$ & --               & matched to ReDiF \\
\bottomrule
\end{tabular}
}
\end{table}

\section{Evaluation and Baseline Protocol}
\label{app:protocol}

This section states the protocol behind Tables~\ref{tab:matched}, \ref{tab:plugin_results},
\ref{tab:reward_ablation_imagenet} and~\ref{tab:reward_ablation_coco} in full, and how it relates to
the published numbers of Table~\ref{tab:published}.

\textbf{Evaluation.} On ImageNet-64 we generate $10{,}000$ class-conditional samples with the
deterministic sampler ($\eta=0$) and compute FID and PRDC against $10{,}000$ real ImageNet-64
validation images. On COCO we generate one image for each of $5{,}000$ MS-COCO validation prompts and
compare against the validation images. The same prompts, seeds and inference settings are used for
every method. This protocol is chosen to be affordable on one device and to be identical across all
retrained rows; it is not the protocol of the published numbers, which use $50$k samples on
ImageNet-64 and the $30$k COCO-2014 zero-shot protocol of~\citet{yin2024DMD}. Absolute values are
therefore higher than published ones, and numbers from the two protocols are never placed in the same
column. Every table reports the distillation ratio as teacher steps $\rightarrow$ student steps.

\textbf{Baseline selection.} A method is retrained on a branch only if its authors report results
with that teacher, so that its published number in Table~\ref{tab:published} refers to the same
teacher. Direct reward backpropagation in the style of DRaFT~\citep{clark2024draft} appears only in
the matched-budget comparison: it is not a step-distillation method and has no published
step-distillation number, but it is the control that isolates the contribution of policy
optimization. CTM~\citep{kim2024ctm} and TFD~\citep{zhang2026tfd} appear only in
Table~\ref{tab:published} and not in Table~\ref{tab:matched}. Both were difficult to bring into the
protocol: each carries a larger set of interacting hyperparameters than the other baselines
(CTM through its jump distribution, adversarial term and target schedule, TFD through the
hidden-state objective and the schedule of its drifting field), and their retrained quality was very
sensitive to that setting. Within the single-device budget and the uniform learning-rate policy of
Appendix~\ref{app:baseline_lr} we did not find a configuration that reached a reasonable operating
point, and reporting a number we could not tune would understate them rather than compare them. We
therefore quote both at the budget their own authors used and make no matched-budget claim about
them.

\textbf{Budget procedure.} All rows are trained for the same number of epochs on the same teacher,
sampler grid, student step count, conditioning pool and hardware, so the budget is controlled by the
epoch count rather than set per method. Two configurations depart from an otherwise uniform setup:
DMD and DMD2 on ImageNet-64 use a larger batch, and the ImageNet-64 distribution-matching students
use a higher LoRA rank than the rank-$4$ adapter used elsewhere, in both cases because the smaller
setting left them at a degenerate point. The realized totals are in the PFLOPs column of
Table~\ref{tab:matched}, which Appendix~\ref{app:cost_model} defines.
All methods are reported at the last checkpoint of the budget. Learning rates are given in Appendix~\ref{app:baseline_lr}. Gradient checkpointing and mixed precision are enabled for all methods. Because the direct-backpropagation control differentiates through the sampling chain, it cannot be memory-matched by default; its backward pass is truncated to the trailing sampler steps that fit in 80\,GB, and this truncation is part of the comparison.

\textbf{Training FLOPs.} We compute analytical training FLOPs as the sum, over all trained and frozen networks, of (number of forward passes)$\times$(forward FLOPs) plus (number of backward passes)$\times$(backward FLOPs), taking backward cost as twice forward cost. This is independent of hardware, kernel efficiency and implementation quality. Every row trains a LoRA adapter rather than the full backbone, so the trainable parameter count is a small fraction of the model and the backward term is an upper bound on the realized cost; the convention is the same for all rows. Pool synthesis is included in the totals rather than counted separately (Appendix~\ref{app:cost_model}) and is identical across rows. Inference FLOPs are never combined with training FLOPs.

\end{document}